\documentclass{article}

\usepackage[nonatbib,final]{neurips_2024}  %

\usepackage[utf8]{inputenc} %
\usepackage[T1]{fontenc}    %
\usepackage[hidelinks]{hyperref}       %
\usepackage{url}            %
\usepackage{booktabs}       %
\usepackage{amsfonts}       %
\usepackage{nicefrac}       %
\usepackage{microtype}      %
\usepackage{xcolor}         %
\usepackage{subcaption}
\usepackage{pgfplots,pgfplotstable}
\usepgfplotslibrary{fillbetween}
\usetikzlibrary{arrows.meta}

\usepackage{graphicx}
\usepackage{color}
\usepackage{float}
\usepackage{comment}
\usepackage{amsmath,amssymb} %
\usepackage{caption}
\usepackage{tabu}
\usepackage{multirow}
\usepackage{calc}
\usepackage{placeins}

\title{Applying Guidance in a Limited Interval Improves Sample and Distribution Quality in Diffusion Models}

\author{%
   \makebox[40mm]{Tuomas Kynk\"a\"anniemi} \\
   Aalto University \\
   \And
   \makebox[25mm]{Miika Aittala} \\
   NVIDIA \\
   \And
   \makebox[40mm]{Tero Karras} \\
   NVIDIA \\
   \AND
   \makebox[40mm]{Samuli Laine} \\
   NVIDIA \\
   \And
   \makebox[25mm]{Timo Aila} \\
   NVIDIA \\
   \And
   \makebox[40mm]{Jaakko Lehtinen} \\
   Aalto University, NVIDIA \\
}

\graphicspath{{./figures/}}
\usepackage[normalem]{ulem}
\def\clap#1{\hbox to 0pt{\hss #1\hss}}%

\definecolor{lightgreen}{rgb}{0, 0.6, 0.3}
\definecolor{olive}{rgb}{0.5, 0.5, 0.0}
\definecolor{maroon}{rgb}{0.69, 0.19, 0.38}
\definecolor{celestialblue}{rgb}{0.29, 0.59, 0.82}
\definecolor{darkgreen}{rgb}{0.0, 0.5, 0.0}
\definecolor{darkblue}{rgb}{0.19, 0.19, 0.62}
\definecolor{grey}{rgb}{0.5,0.5,0.5}

\newcommand{\tstrut}{\vphantom{$\hat{A}$}}   %
\newcommand{\hs}[1]{\hspace{#1mm}}

\newcommand{\FID}{$\text{FID}$}
\newcommand{\FDDINO}{$\text{FD}_{\text{DINOv2}}$}

\newcommand{\xx}{\mathbf{x}}
\newcommand{\yy}{\mathbf{y}}
\newcommand{\nn}{\mathbf{n}}
\newcommand{\cc}{\mathbf{c}}
\newcommand{\boldzero}{\boldsymbol{0}}
\newcommand{\boldI}{\mathbf{I}}
\newcommand{\EE}{\mathbb{E}}
\newcommand{\pdata}{p_\text{data}}
\newcommand{\gdots}{\cdots\hspace*{-.08em}}
\newcommand{\smax}{\sigma_\text{max}}

\newcommand{\gweight}{w}
\newcommand{\sigmalo}{\sigma_\text{lo}}
\newcommand{\sigmahi}{\sigma_\text{hi}}
\newcommand{\sigmamax}{\sigma_{\text{max}}}
\newcommand{\sigmamin}{\sigma_{\text{min}}}

\pgfplotsset{compat=1.18}
\pgfplotsset{tick label style={font=\scriptsize}}
\pgfplotsset{legend style={font=\tiny, row sep=-0.5mm}}
\pgfplotsset{tick style={draw=none}}
\pgfplotsset{major grid style={gray!40}}
\pgfplotsset{every axis plot/.append style={line width=0.5pt, mark size=0.6pt}}
\pgfplotsset{dashed/.style={dash pattern=on 1.2pt off 1.2pt}}
\pgfplotsset{legend image code/.code={\draw[mark repeat=2, mark phase=2] plot coordinates {(0mm, 0.2mm) (1.5mm, 0.2mm) (3mm, 0.2mm)};}} %
\pgfplotsset{plot graphics/includegraphics={interpolate=true}}
\pgfplotsset{filter/.style 2 args={x filter/.code={\edef\tempa{\thisrow{#1}}\edef\tempb{#2}\ifx\tempa\tempb\else\fi}}}

\definecolor{C0}{rgb}{0.121569, 0.466667, 0.705882}
\definecolor{C1}{rgb}{1.000000, 0.498039, 0.054902}
\definecolor{C2}{rgb}{0.172549, 0.627451, 0.172549}
\definecolor{C3}{rgb}{0.839216, 0.152941, 0.156863}
\definecolor{C4}{rgb}{0.580392, 0.403922, 0.741176}
\definecolor{C5}{rgb}{0.549020, 0.337255, 0.294118}
\definecolor{C6}{rgb}{0.890196, 0.466667, 0.760784}
\definecolor{C7}{rgb}{0.498039, 0.498039, 0.498039}
\definecolor{C8}{rgb}{0.737255, 0.741176, 0.133333}
\definecolor{C9}{rgb}{0.090196, 0.745098, 0.811765}

\definecolor{C10}{rgb}{0.429020, 0.653334, 0.808823}
\definecolor{C11}{rgb}{0.648628, 0.786667, 0.882353}
\definecolor{C12}{rgb}{0.462157, 0.757843, 0.462157}
\definecolor{C13}{rgb}{0.669020, 0.850980, 0.669020}

\newcommand{\h}{0mm}
\newcommand{\hh}{0mm}
\newcommand{\hhh}{0mm}
\newcommand{\hhhh}{0mm}
\newcommand{\s}{\hphantom{0}}

\newcommand{\tabResultsFiveTwelve}{%
\begin{table}[t]%
\centering\footnotesize%
\resizebox{1.0\linewidth}{!}{%
\begin{tabu}{|l@{\hs{1.5}}l|r@{\hs{3.5}}c|r@{\hs{2.5}}r|c@{\hs{1.5}}c|c@{\hs{0}}c|}
\tabucline{-}
\multicolumn{2}{|l|}{\multirow{2}{*}{\bf ImageNet-512}} & \multicolumn{2}{c|}{\bf Quality metric} & \multicolumn{2}{c|}{\bf Model size} & \multicolumn{2}{c|}{\bf Guidance interval} & \multicolumn{2}{c|}{\bf Guidance weight}\\
&& \multicolumn{1}{c}{FID $\downarrow$} & \multicolumn{1}{@{}c|}{FD\textsubscript{DINOv2} $\downarrow$} & \multicolumn{1}{c}{Mparams} & \multicolumn{1}{@{}c|}{Gflops} & \multicolumn{1}{c}{FID} & \multicolumn{1}{@{}c|}{FD\textsubscript{DINOv2}} &  \multicolumn{1}{@{\hs{3}}r@{\hs{0}}}{\makebox[1mm][]{}FID} & \multicolumn{1}{@{\hs{0}}c@{\hs{0}}|}{FD\textsubscript{DINOv2}}\\
\tabucline{-}\tstrut%
{EDM2-S}    & {\cite{Karras2023edm2} w/ CFG \cite{Ho2021classifierfree}}  & {2.23} & {52.32} & {280\s\s}  & {102\s} & Full & Full  & {\s1.4}& {1.9}\\
{EDM2-S}    & {\cite{Karras2023edm2} w/ guidance interval}  & {1.68} & {46.25} & {280\s\s}  & {102\s} & {$\left(0.28, 2.90\right]$} & {$\left(0.60, 5.00\right]$}  & {\s2.1} & {3.2}\\
{EDM2-XXL}  & {\cite{Karras2023edm2} w/ CFG \cite{Ho2021classifierfree}}  & {1.81} & {33.09} & {1523\s\s} & {552\s} & Full & Full  & {\s1.2}& {1.7}\\
{EDM2-XXL}  & {\cite{Karras2023edm2} w/ guidance interval}  & {\bf 1.40} & {\bf 29.16} & {1523\s\s} & {552\s}  & {$\left(0.19, 1.61\right]$} & {$\left(0.60, 5.00\right]$}  & {\s2.0} & {2.9}\\
{DiT-XL/2}    & {\cite{Peebles2022} w/ CFG \cite{Ho2021classifierfree}} & {3.04} & {51.97}  & {675\s\s}  & {525\s}  & Full & {Full} & {\s1.5} & {2.0}\\
{DiT-XL/2}    & {\cite{Peebles2022} w/ guidance interval}    & {2.40} & {43.94}  & {675\s\s}  & {525\s}  & {$\left(0.34, 1.02\right]$} & {$\left(0.45,1.23\right]$} & {\s2.5}& {4.0}\\
\tabucline{-}
\end{tabu}
}%
\vspace*{\baselineskip}
\caption{\label{tabResultsFiveTwelve}%
Quantitative results on ImageNet-512. Limiting the classifier-free guidance (CFG) to an interval improves both FID and FD\textsubscript{DINOv2} significantly, without altering the model complexity. 
The sampling cost is a bit lower due to fewer guidance evaluations.
This holds for a small (S) and large (XXL) variants of the state-of-the-art EDM2 model~\cite{Karras2023edm2}, as well as diffusion transformers \cite{Peebles2022}.
The model complexity numbers are copied from the EDM2 paper.
}%
\end{table}
}%

\newcommand{\figImageNetGuidanceFIDCurvesA}{%
\centering\footnotesize%
\gdef\datafile{figures/imagenet512/metrics/fd_inception_v3_tf.txt}%
\begin{tikzpicture}%
\begin{axis}[
  width={0.529\linewidth}, height={50mm}, grid={major},
  xmin={1.0}, xmax={4.0}, xtick={1.0, 1.5, 2.0, 2.5, 3.0, 3.5, 4.0}, xticklabels={$1.0$, $1.5$, $2.0$, $2.5$, $3.0$, $3.5$, $4.0$},
  ymin={0}, ymax={13}, ytick={2, 4, 6, 8, 10, 12}, yticklabels={$2$, $4$, $6$, $8$, $10$, $12$},
  xlabel={Guidance weight $\gweight$}, x label style={at={(axis description cs:0.5,-0.09)}, anchor=north},
  ylabel={\FID}, y label style={at={(axis description cs:-0.06,0.5)}, anchor=south},
  legend pos={north west}, legend cell align={left}, legend columns={2},
]
\addplot[C0, opacity=0, line width=0pt, name path=S_ours_hi, forget plot] table[x=g, y=fid_max, col sep=comma, filter={label}{edm2s-icfg_0.42_2.90}]{\datafile};
\addplot[C0, name path=S_ours_lo] table[x=g, y=fid_min, col sep=comma, filter={label}{edm2s-icfg_0.42_2.90}] {\datafile};
\addplot[C0, opacity=0.4, forget plot] fill between[of=S_ours_lo and S_ours_hi];
\addplot[C0, mark=*, forget plot, nodes near coords align={south}, nodes near coords={\tiny$1.68$}] coordinates {(2.1, 1.68)};
\addlegendentry{S, Ours, $\sigma \in (0.28, 2.90]$}
\addplot[C1, opacity=0, line width=0pt, name path=S_cfg_hi, forget plot] table[x=g, y=fid_max, col sep=comma, filter={label}{edm2s-cfg_0_inf}]{\datafile};
\addplot[C1, name path=S_cfg_lo] table[x=g, y=fid_min, col sep=comma, filter={label}{edm2s-cfg_0_inf}] {\datafile};
\addplot[C1, opacity=0.4, forget plot] fill between[of=S_cfg_lo and S_cfg_hi];
\addplot[C1, mark=*, forget plot, nodes near coords align={south}, nodes near coords={\tiny\hspace{-1mm}$2.23$}] coordinates {(1.4, 2.27)}; %
\addlegendentry{S, CFG}
\addplot[C2, opacity=0, line width=0pt, name path=XXL_ours_hi, forget plot] table[x=g, y=fid_max, col sep=comma, filter={label}{edm2xxl-icfg_0.28_1.61}]{\datafile};
\addplot[C2, name path=XXL_ours_lo] table[x=g, y=fid_min, col sep=comma, filter={label}{edm2xxl-icfg_0.28_1.61}] {\datafile};
\addplot[C2, opacity=0.4, forget plot] fill between[of=XXL_ours_lo and XXL_ours_hi];
\addplot[C2, mark=*, forget plot, nodes near coords align={north}, nodes near coords={\tiny$1.40$}] coordinates {(2.0, 1.40)};
\addlegendentry{XXL, Ours, $\sigma \in (0.19, 1.61]$}
\addplot[C3, opacity=0, line width=0pt, name path=XXL_cfg_hi, forget plot] table[x=g, y=fid_max, col sep=comma, filter={label}{edm2xxl-cfg_0_inf}]{\datafile};
\addplot[C3, name path=XXL_cfg_lo] table[x=g, y=fid_min, col sep=comma, filter={label}{edm2xxl-cfg_0_inf}] {\datafile};
\addplot[C3, opacity=0.4, forget plot] fill between[of=XXL_cfg_lo and XXL_cfg_hi];
\addplot[C3, mark=*, forget plot, nodes near coords align={north}, nodes near coords={\tiny$1.81$}] coordinates {(1.2, 1.81)};
\addlegendentry{XXL, CFG}
\end{axis}
\end{tikzpicture}%
}%

\newcommand{\figImageNetGuidanceFIDCurvesB}{%
\centering\footnotesize%
\gdef\datafile{figures/imagenet512/metrics/fd_dinov2.txt}%
\begin{tikzpicture}%
\begin{axis}[
  width={0.529\linewidth}, height={50mm}, grid={major},
  xmin={1.0}, xmax={4.0}, xtick={1.0, 1.5, 2.0, 2.5, 3.0, 3.5, 4.0}, xticklabels={$1.0$, $1.5$, $2.0$, $2.5$, $3.0$, $3.5$, $4.0$},
  ymin={20}, ymax={130}, ytick={40, 60, 80, 100, 120}, yticklabels={$40$, $60$, $80$, $100$, $120$},
  xlabel={Guidance weight $\gweight$}, x label style={at={(axis description cs:0.5,-0.09)}, anchor=north},
  ylabel={\FDDINO}, y label style={at={(axis description cs:-0.09,0.5)}, anchor=south},
  legend pos={north east}, legend cell align={left}, legend columns={2},
]
\addplot[C0, opacity=0, line width=0pt, name path=S_ours_hi, forget plot] table[x=g, y=fid_max, col sep=comma, filter={label}{edm2s-icfg_0.85_5.00}]{\datafile};
\addplot[C0, name path=S_ours_lo] table[x=g, y=fid_min, col sep=comma, filter={label}{edm2s-icfg_0.85_5.00}] {\datafile};
\addplot[C0, opacity=0.4, forget plot] fill between[of=S_ours_lo and S_ours_hi];
\addplot[C0, mark=*, forget plot, nodes near coords align={north}, nodes near coords={\tiny$46.25$}] coordinates {(3.2, 46.25)};
\addlegendentry{S, Ours, $\sigma \in (0.60, 5.00]$}
\addplot[C1, opacity=0, line width=0pt, name path=S_cfg_hi, forget plot] table[x=g, y=fid_max, col sep=comma, filter={label}{edm2s-cfg_0_inf}]{\datafile};
\addplot[C1, name path=S_cfg_lo] table[x=g, y=fid_min, col sep=comma, filter={label}{edm2s-cfg_0_inf}] {\datafile};
\addplot[C1, opacity=0.4, forget plot] fill between[of=S_cfg_lo and S_cfg_hi];
\addplot[C1, mark=*, forget plot, nodes near coords align={north}, nodes near coords={\tiny$52.32$}] coordinates {(1.9, 52.17)}; %
\addlegendentry{S, CFG}
\addplot[C2, opacity=0, line width=0pt, name path=XXL_ours_hi, forget plot] table[x=g, y=fid_max, col sep=comma, filter={label}{edm2xxl-icfg_0.85_5.00}]{\datafile};
\addplot[C2, name path=XXL_ours_lo] table[x=g, y=fid_min, col sep=comma, filter={label}{edm2xxl-icfg_0.85_5.00}] {\datafile};
\addplot[C2, opacity=0.4, forget plot] fill between[of=XXL_ours_lo and XXL_ours_hi];
\addplot[C2, mark=*, forget plot, nodes near coords align={north}, nodes near coords={\tiny$29.16$}] coordinates {(2.9, 29.16)};
\addlegendentry{XXL, Ours, $\sigma \in (0.60, 5.00]$}
\addplot[C3, opacity=0, line width=0pt, name path=XXL_cfg_hi, forget plot] table[x=g, y=fid_max, col sep=comma, filter={label}{edm2xxl-cfg_0_inf}]{\datafile};
\addplot[C3, name path=XXL_cfg_lo] table[x=g, y=fid_min, col sep=comma, filter={label}{edm2xxl-cfg_0_inf}] {\datafile};
\addplot[C3, opacity=0.4, forget plot] fill between[of=XXL_cfg_lo and XXL_cfg_hi];
\addplot[C3, mark=*, forget plot, nodes near coords align={north}, nodes near coords={\tiny$33.09$}] coordinates {(1.7, 32.70)}; %
\addlegendentry{XXL, CFG}
\end{axis}
\end{tikzpicture}%
}%

\newcommand{\figImageNetGuidanceFIDCurves}{%
\begin{figure}[t]%
\centering\footnotesize%
\figImageNetGuidanceFIDCurvesA\hfill\figImageNetGuidanceFIDCurvesB%
\caption{%
\FID{} and \FDDINO{} as a function of guidance weight for classifier-free guidance (orange, red) and our method where the guidance has been limited to the stated interval (blue, green). The shaded regions indicate the min/max over three evaluations.
}\vspace*{-2mm}%
\label{fig:imagenet512_guidance_vs_fid}%
\end{figure}
}%

\newcommand{\figAblationGraphsA}{%
\centering\footnotesize%
\gdef\datafile{figures/imagenet512/metrics/imagenet512-ablation_graphs_A.txt}%
\begin{tikzpicture}%
\begin{axis}[
  width={0.53\linewidth}, height={50mm}, grid={major},
  xmin={22}, xmax={0}, x coord trafo/.code=\pgfmathparse{-##1}, xtick={0, 5, 10, 15, 20, 25, 31}, xticklabels={$80$, $31.22$, $10.52$, $2.90$, $0.60$, $0.08$, $0.002$},
  ymin={0.5}, ymax={8}, ytick={1, 2, 3, 4, 5, 6, 7, 8}, yticklabels={$1$, $2$, $3$, $4$, $5$, $6$, $7$, $8$},
  xlabel={Highest noise level with guidance, $\sigmahi$}, x label style={at={(axis description cs:0.5,-0.09)}, anchor=north},
  ylabel={\FID}, y label style={at={(axis description cs:-0.05,0.5)}, anchor=south},
  legend pos={north west}, legend cell align={left}, %
]
\addplot[C0, opacity=0, line width=0pt, name path=S_ours_hi, forget plot] table[x=idx_start, y=fid_max, col sep=comma, filter={label}{edm2s-a}]{\datafile};
\addplot[C0, name path=S_ours_lo] table[x=idx_start, y=fid_min, col sep=comma, filter={label}{edm2s-a}] {\datafile}; \addlegendentry{S, Ours, $\gweight = 2.1$, $\sigmalo=0.28$}
\addplot[C0, opacity=0.4, forget plot] fill between[of=S_ours_lo and S_ours_hi];
\addplot[C0, mark=*, forget plot, nodes near coords align={north}, nodes near coords={\hspace{-1mm}\tiny$1.68$}] coordinates {(15, 1.68)};
\addplot[C2, opacity=0, line width=0pt, name path=XXL_ours_hi, forget plot] table[x=idx_start, y=fid_max, col sep=comma, filter={label}{edm2xxl-a}]{\datafile};
\addplot[C2, name path=XXL_ours_lo] table[x=idx_start, y=fid_min, col sep=comma, filter={label}{edm2xxl-a}] {\datafile}; \addlegendentry{XXL, Ours, $\gweight = 2.0$, $\sigmalo=0.19$}
\addplot[C2, opacity=0.4, forget plot] fill between[of=XXL_ours_lo and XXL_ours_hi];
\addplot[C2, mark=*, forget plot, nodes near coords align={north}, nodes near coords={\hspace{1mm}\tiny$1.40$}] coordinates {(17, 1.40)};
\end{axis}
\end{tikzpicture}%
}%

\newcommand{\figAblationGraphsB}{%
\centering\footnotesize%
\gdef\datafile{figures/imagenet512/metrics/imagenet512-ablation_graphs_B.txt}%
\begin{tikzpicture}%
\begin{axis}[
  width={0.53\linewidth}, height={50mm}, grid={major},
  xmin={31}, xmax={15}, x coord trafo/.code=\pgfmathparse{-##1}, xtick={15, 19, 23, 27, 31}, xticklabels={$2.90$, $0.85$, $0.19$, $0.028$, $0.002$},
  ymin={0.5}, ymax={8}, ytick={1, 2, 3, 4, 5, 6, 7, 8}, yticklabels={$1$, $2$, $3$, $4$, $5$, $6$, $7$, $8$},
  xlabel={Lowest noise level with guidance, $\sigmalo$}, x label style={at={(axis description cs:0.5,-0.09)}, anchor=north},
  ylabel={\FID}, y label style={at={(axis description cs:-0.05,0.5)}, anchor=south},
  legend pos={north west}, legend cell align={left}, %
]
\addplot[C0, opacity=0, line width=0pt, name path=S_ours_hi, forget plot] table[x=idx_stop, y=fid_max, col sep=comma, filter={label}{edm2s-a}]{\datafile};
\addplot[C0, name path=S_ours_lo] table[x=idx_stop, y=fid_min, col sep=comma, filter={label}{edm2s-a}] {\datafile}; \addlegendentry{S, Ours, $\gweight = 2.1$, $\sigmahi=2.90$}
\addplot[C0, opacity=0.4, forget plot] fill between[of=S_ours_lo and S_ours_hi];
\addplot[C0, mark=*, forget plot, nodes near coords align={south}, nodes near coords={\tiny$1.68$}] coordinates {(21, 1.68)};
\addplot[C2, opacity=0, line width=0pt, name path=XXL_ours_hi, forget plot] table[x=idx_stop, y=fid_max, col sep=comma, filter={label}{edm2xxl-a}]{\datafile};
\addplot[C2, name path=XXL_ours_lo] table[x=idx_stop, y=fid_min, col sep=comma, filter={label}{edm2xxl-a}] {\datafile}; \addlegendentry{XXL, Ours, $\gweight = 2.0$, $\sigmahi=1.61$}
\addplot[C2, opacity=0.4, forget plot] fill between[of=XXL_ours_lo and XXL_ours_hi];
\addplot[C2, mark=*, forget plot, nodes near coords align={north}, nodes near coords={\tiny$1.40$}] coordinates {(22, 1.40)};
\end{axis}
\end{tikzpicture}%
}%

\newcommand{\figAblationGraphs}{%
\begin{figure}[t]%
\centering\footnotesize%
\figAblationGraphsA\hfill\figAblationGraphsB%
\caption{
Sensitivity of FID to the chosen guidance interval.
\textbf{Left:} Sweep over $\sigmahi$ with optimal $\sigmalo$ and $\gweight$.
\textbf{Right:} Sweep over $\sigmalo$ with optimal $\sigmahi$ and $\gweight$. The shaded regions indicate the min/max over three evaluations.
}\vspace*{2mm}%
\label{fig:ablation_graphs}%
\end{figure}%
}%

\newcommand{\figQualitativeResultsA}{
    \renewcommand{\h}{0.313\linewidth}
    \renewcommand{\hh}{4mm}
    \newcommand{\trim}{1}
    \begin{figure}[p]%
        \makebox[\hh][l]{}%
        \makebox[\h]{CFG with low guidance}\hfill%
        \makebox[\h]{CFG with high guidance}\hfill%
        \makebox[\h]{Ours with high guidance}\\%
        \makebox[\hh][l]{}%
        \makebox[\h]{\small $\gweight=2$,\,\,\,$\sigma\in\left(0,\infty\right)$}\hfill%
        \makebox[\h]{\small $\gweight=16$,\,\,\,$\sigma\in\left(0,\infty\right)$}\hfill%
        \makebox[\h]{\small $\gweight=16$,\,\,\,$\sigma\in\left(0.28,5.42\right]$}\\%
        \makebox[\hh]{}%
        \makebox[\h]{\small{\color{red}fuzzy details}, {\color{darkgreen}high diversity}}\hfill%
        \makebox[\h]{\small{\color{darkgreen}crisp details}, {\color{red}low diversity}}\hfill%
        \makebox[\h]{\small{\color{darkgreen}crisp details}, {\color{darkgreen}high diversity}}\\%
        \makebox[\hh][l]{\rotatebox[origin=l]{90}{\makebox[0mm][c]{\hspace*{21mm}\small\emph{Rembrandt painting of a raccoon.}}}}%
        \includegraphics[width=\h, trim={0 7 0 7}, clip]{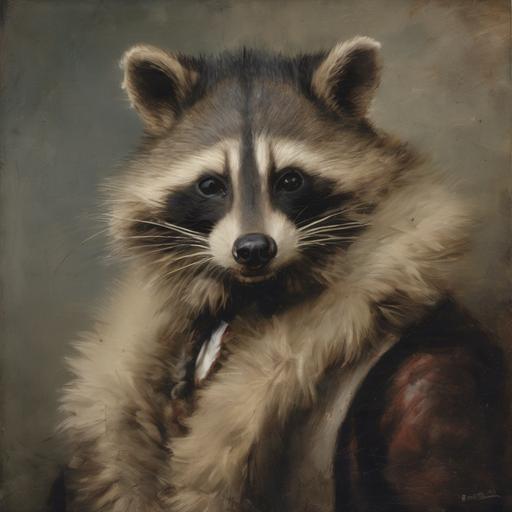}\hfill%
        \includegraphics[width=\h, trim={0 7 0 7}, clip]{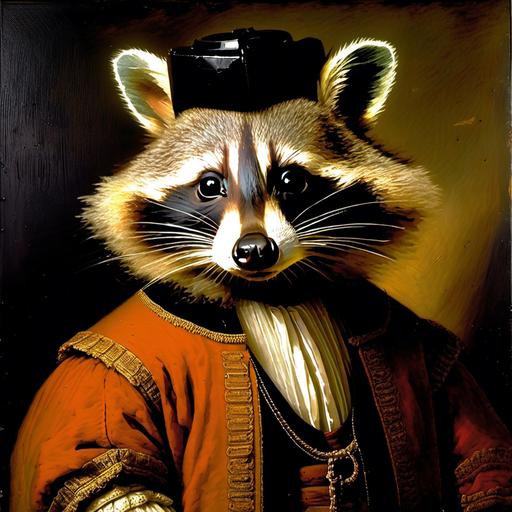}\hfill%
        \includegraphics[width=\h, trim={0 7 0 7}, clip]{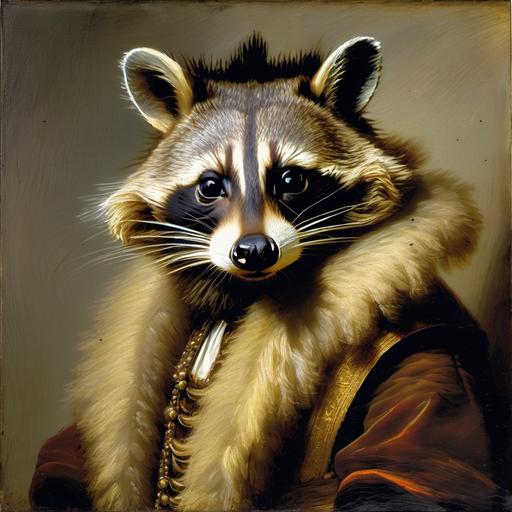}\\[-0.5mm]%
        \makebox[\hh][l]{}%
        \includegraphics[width=\h, trim={0 7 0 7}, clip]{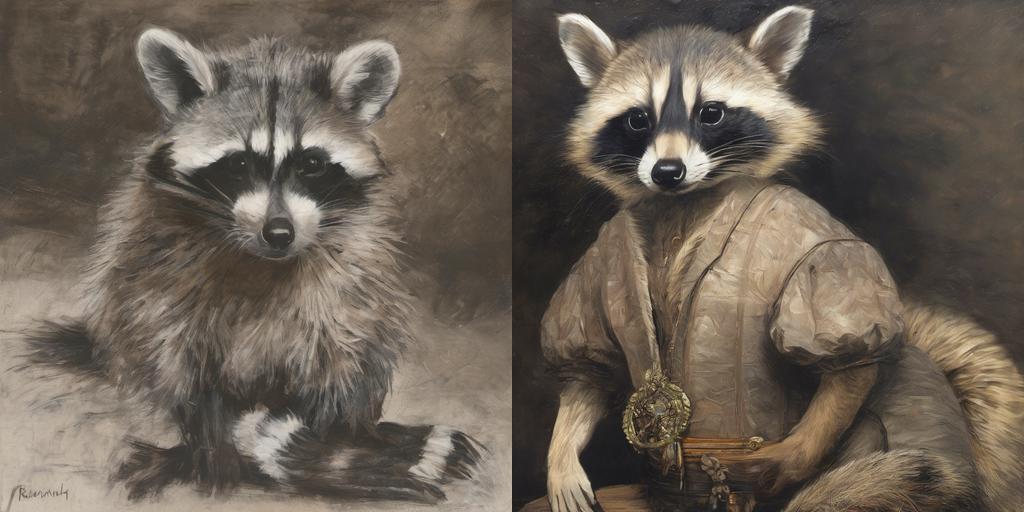}\hfill%
        \includegraphics[width=\h, trim={0 7 0 7}, clip]{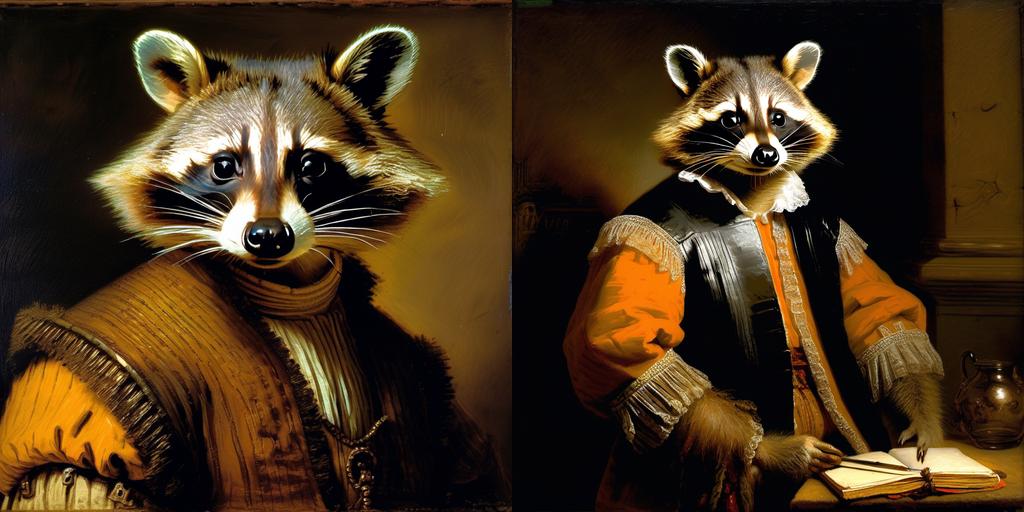}\hfill%
        \includegraphics[width=\h, trim={0 7 0 7}, clip]{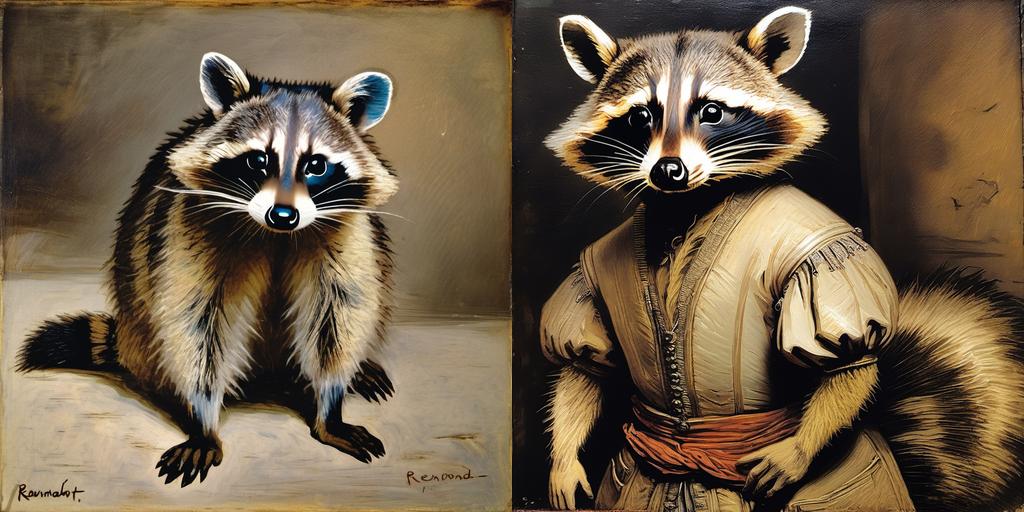}\\[2mm]%
        \makebox[\hh][l]{\rotatebox[origin=l]{90}{\makebox[0mm][c]{\hspace*{42mm}\small\emph{An adorable painting of a Dachshund.}}}}%
        \includegraphics[width=\h, trim={0 7 0 7}, clip]{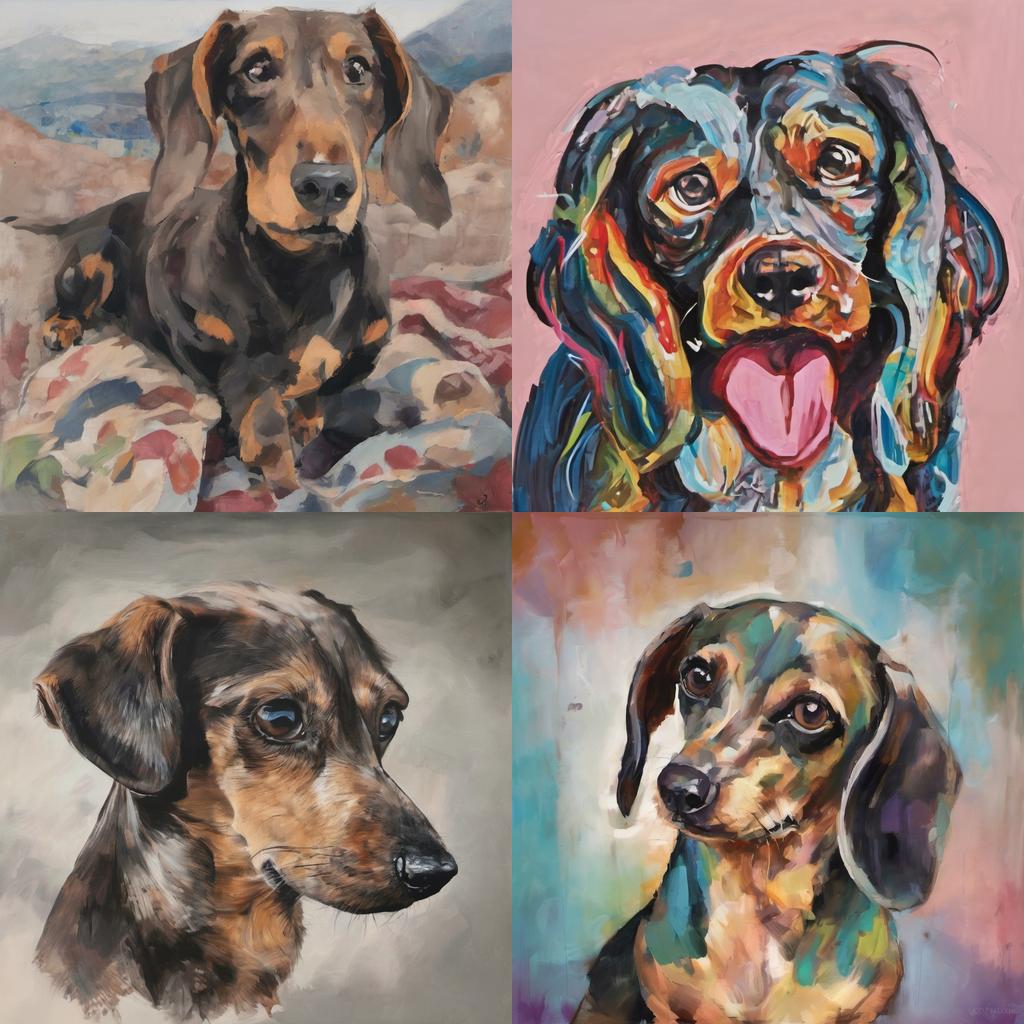}\hfill%
        \includegraphics[width=\h, trim={0 7 0 7}, clip]{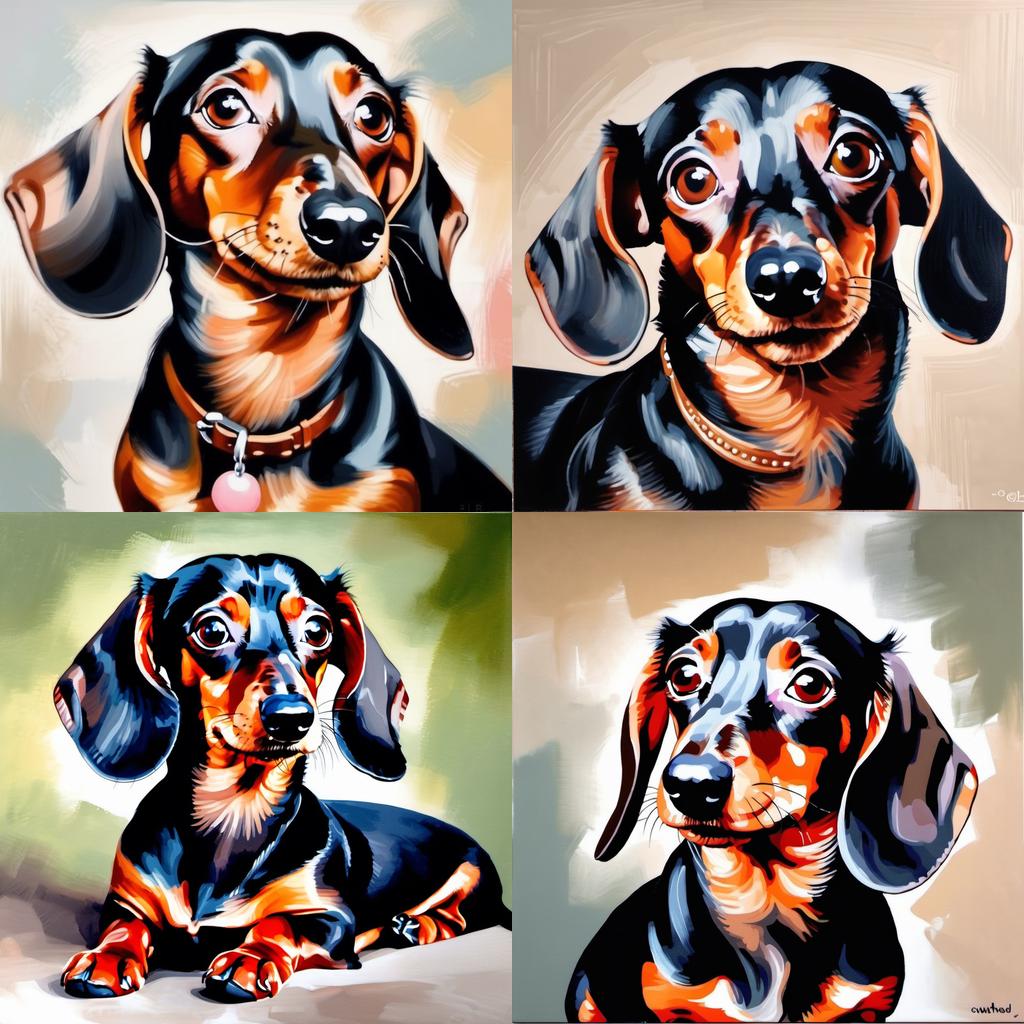}\hfill%
        \includegraphics[width=\h, trim={0 7 0 7}, clip]{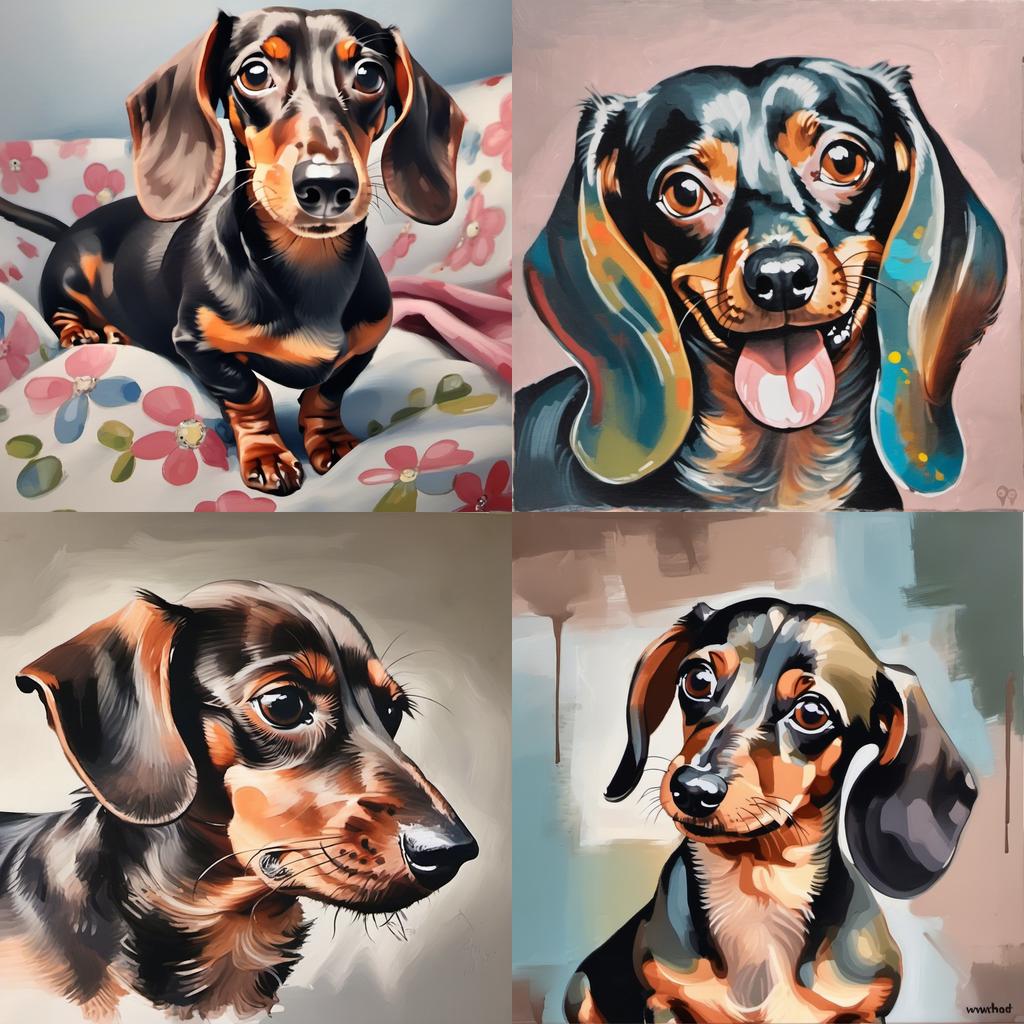}\\[1.0mm]%
        \makebox[\hh][l]{}%
        \makebox[\h]{\small $\gweight=1$\,\,\,(no guidance)}\hfill%
        \makebox[\h]{\small $\gweight=5$,\,\,\,$\sigma\in\left(0,\infty\right)$}\hfill%
        \makebox[\h]{\small $\gweight=5$,\,\,\,$\sigma\in\left(0.19,1.61\right]$}\\%
        \makebox[\hh]{}%
        \makebox[\h]{\small{\color{red}fuzzy details}, {\color{darkgreen}high diversity}}\hfill%
        \makebox[\h]{\small{\color{darkgreen}crisp details}, {\color{red}low diversity}}\hfill%
        \makebox[\h]{\small{\color{darkgreen}crisp details}, {\color{darkgreen}high diversity}}\\%
        \makebox[\hh][l]{\rotatebox[origin=l]{90}{\makebox[0mm][c]{\hspace*{43mm}\small ImageNet class 64: \emph{green mamba}}}}%
        \includegraphics[width=\h, trim={0 7 0 7}, clip]{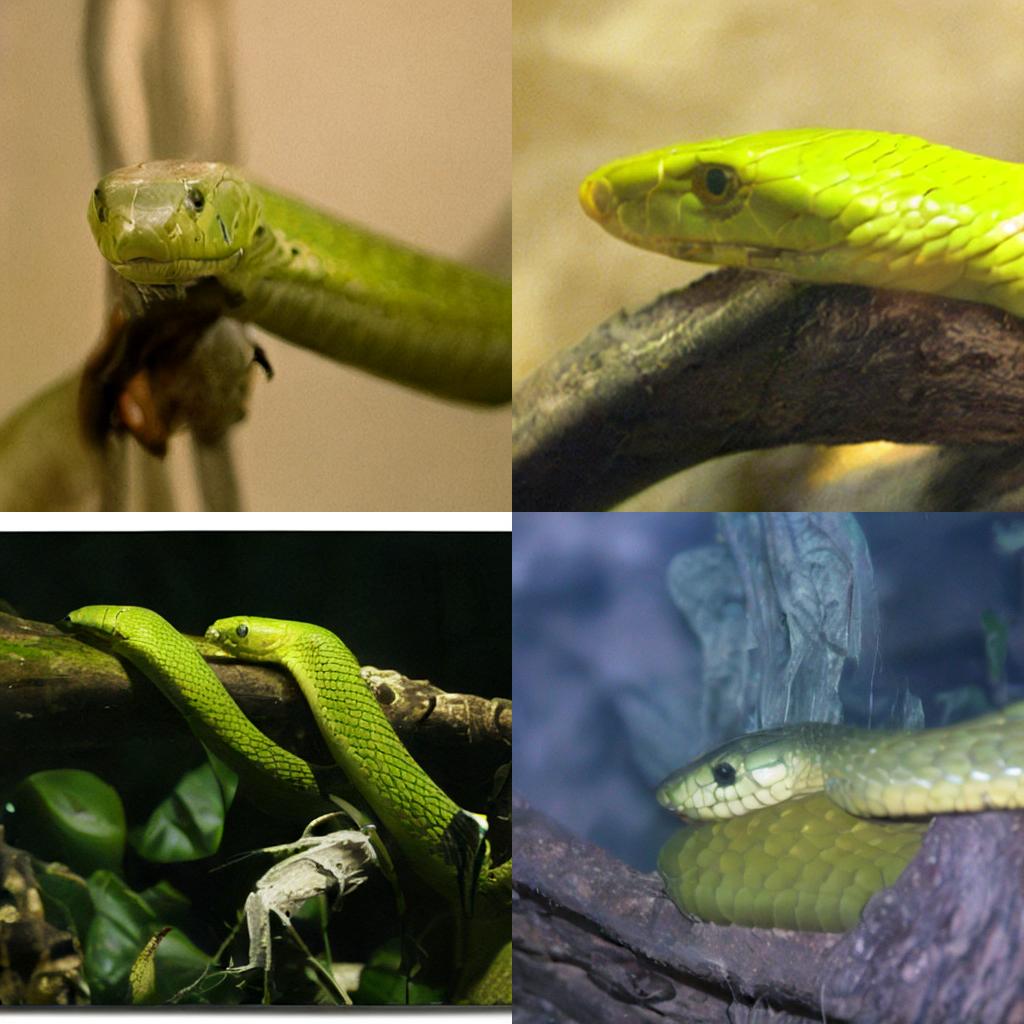}\hfill%
        \includegraphics[width=\h, trim={0 7 0 7}, clip]{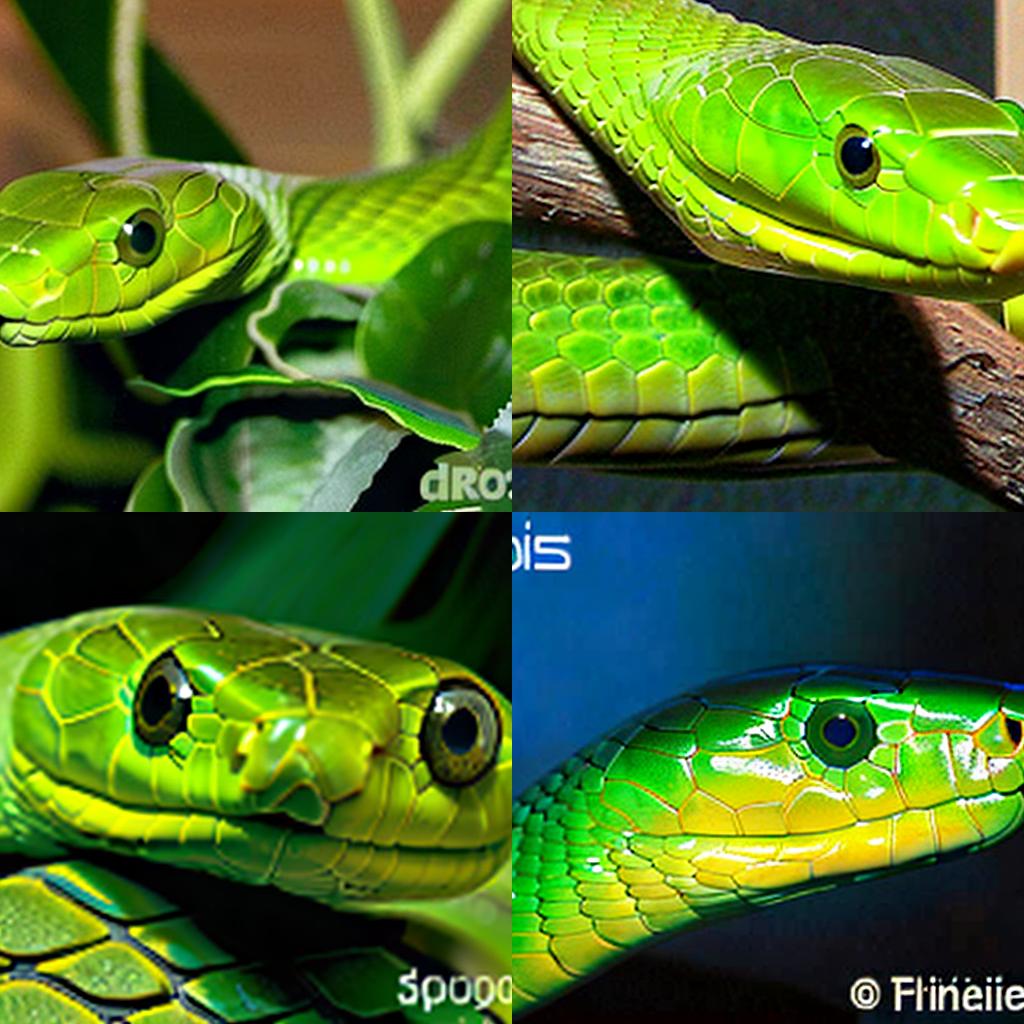}\hfill%
        \includegraphics[width=\h, trim={0 7 0 7}, clip]{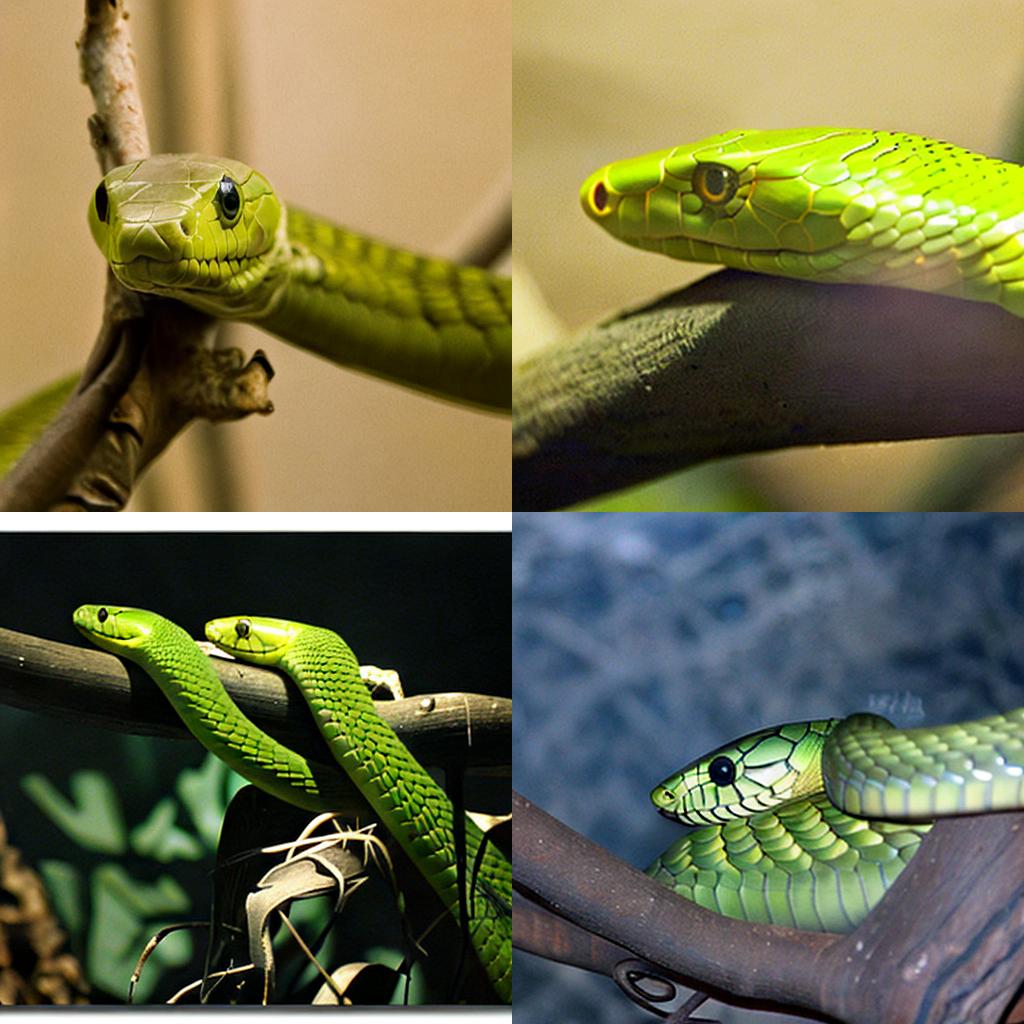}\\[2mm]%
        \makebox[\hh][l]{\rotatebox[origin=l]{90}{\makebox[0mm][c]{\hspace*{42mm}\small ImageNet class 959: \emph{carbonara}}}}%
        \includegraphics[width=\h, trim={0 7 0 7}, clip]{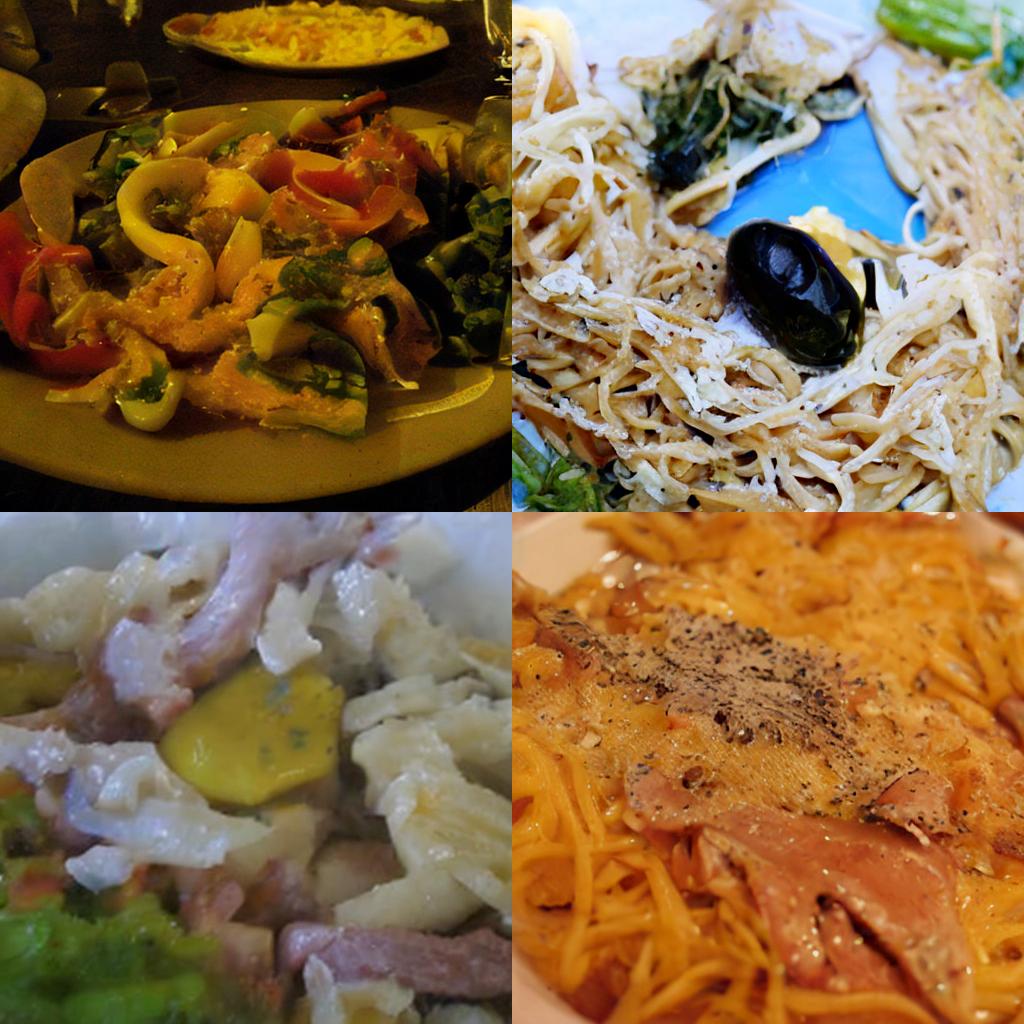}\hfill%
        \includegraphics[width=\h, trim={0 7 0 7}, clip]{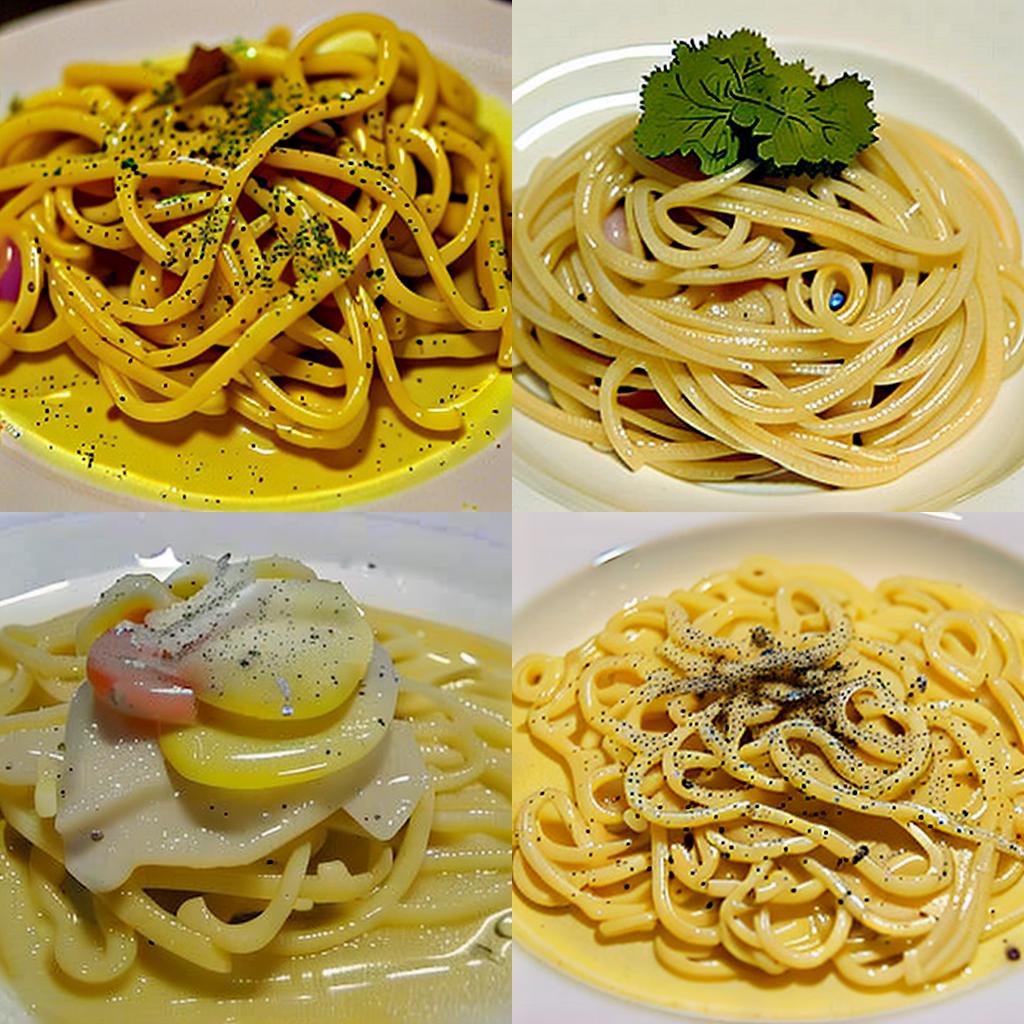}\hfill%
        \includegraphics[width=\h, trim={0 7 0 7}, clip]{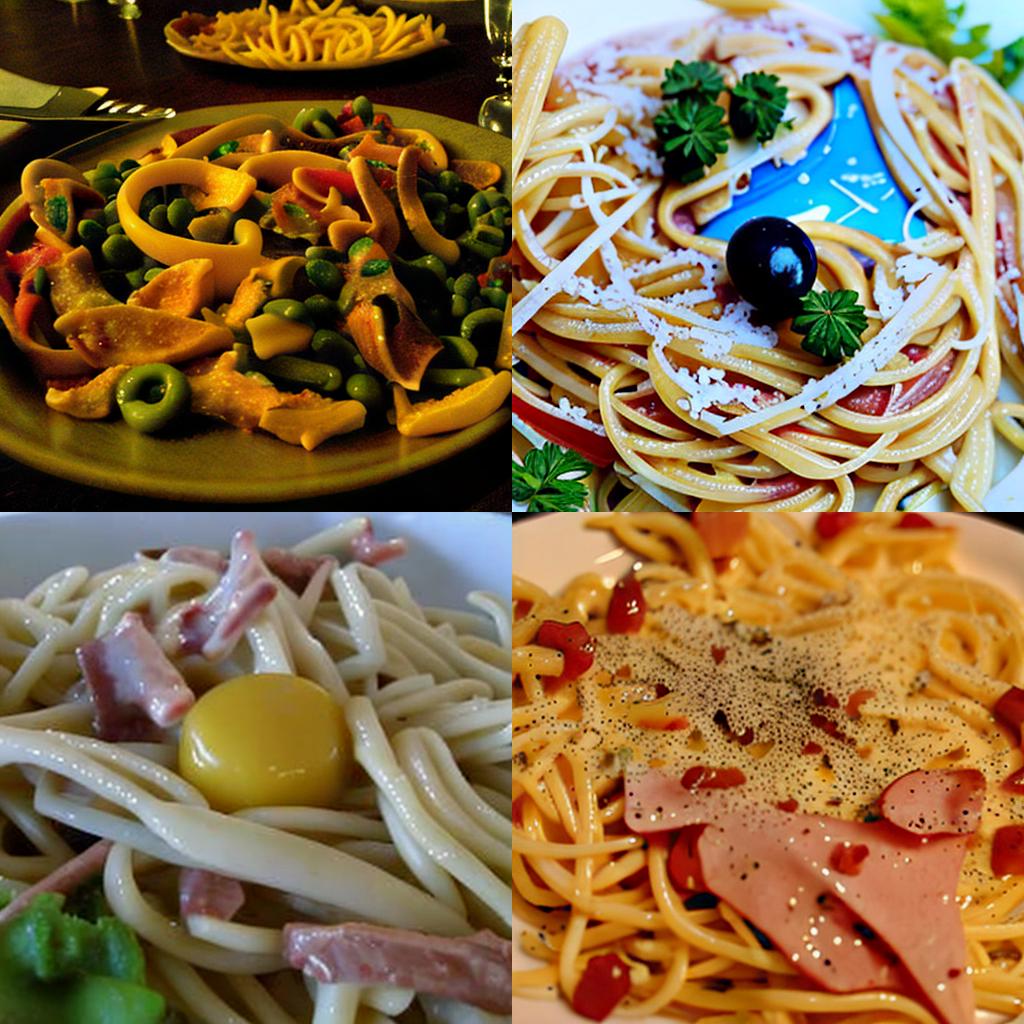}%
        \caption{%
            Traditional CFG vs. our method.
            \textbf{Left:} Low $\gweight$ yields diverse but fuzzy images that lack detail.
            \textbf{Middle:} Increasing $\gweight$ adds crispness, but reduces diversity and oversaturates the colors.
            \textbf{Right:} Our method reduces these effects while retaining the crisp look.
        }%
        \label{fig:sdxl_imagenet_qualitative_results_A}
    \end{figure}
}

\newcommand{\figQualitativeResultsB}{
    \renewcommand{\h}{0.2\linewidth}
    \renewcommand{\hh}{\linewidth}
    \renewcommand{\hhh}{0.495\linewidth}
    \renewcommand{\hhhh}{0.167\linewidth}
    \begin{figure}[t]%
        \makebox[\h][c]{$\gweight = 2$}%
        \makebox[\h][c]{$\gweight = 4$}%
        \makebox[\h][c]{$\gweight = 8$}%
        \makebox[\h][c]{$\gweight = 12$}%
        \makebox[\h][c]{$\gweight = 16$}\\%
        \includegraphics[width=\hh]{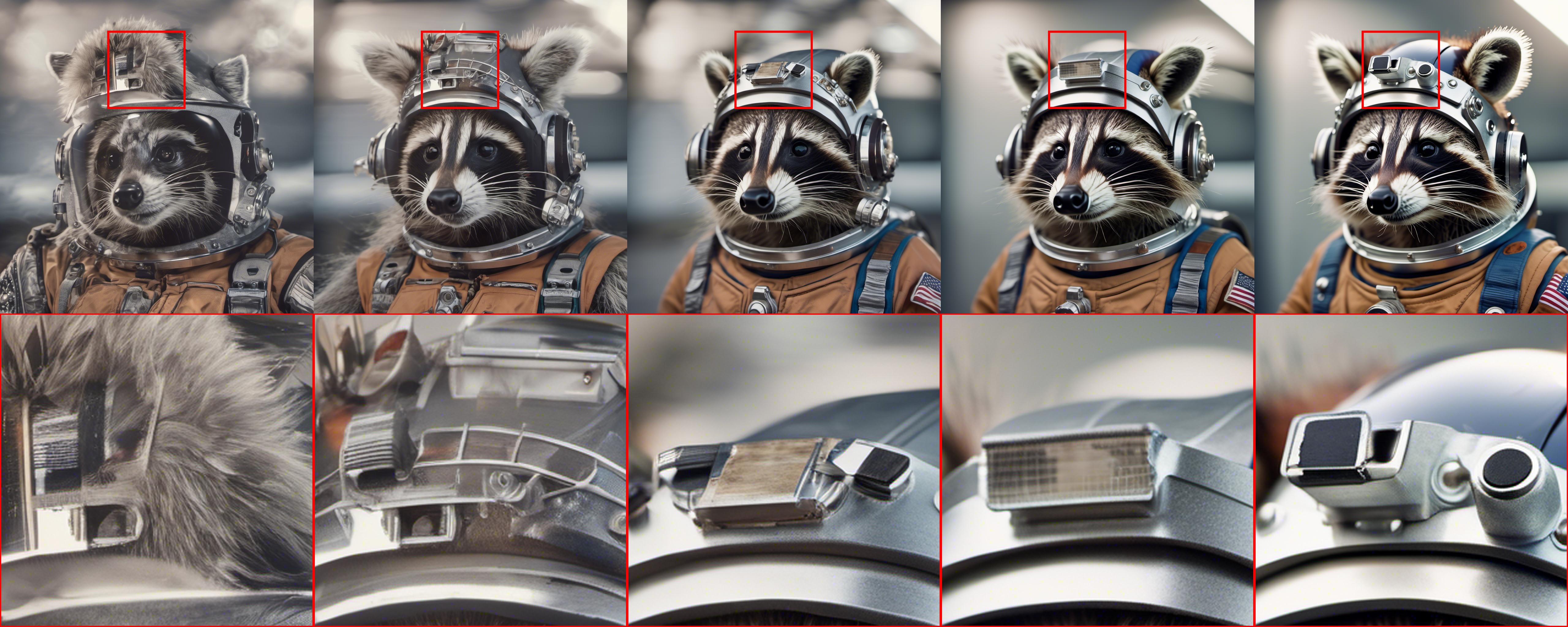}\\[-0.5mm]%
        \makebox[\hh][c]{\small\emph{A 4k dslr photo of a raccoon wearing an astronaut helmet, photorealistic.}}\\[1mm]%
        \includegraphics[width=\hh]{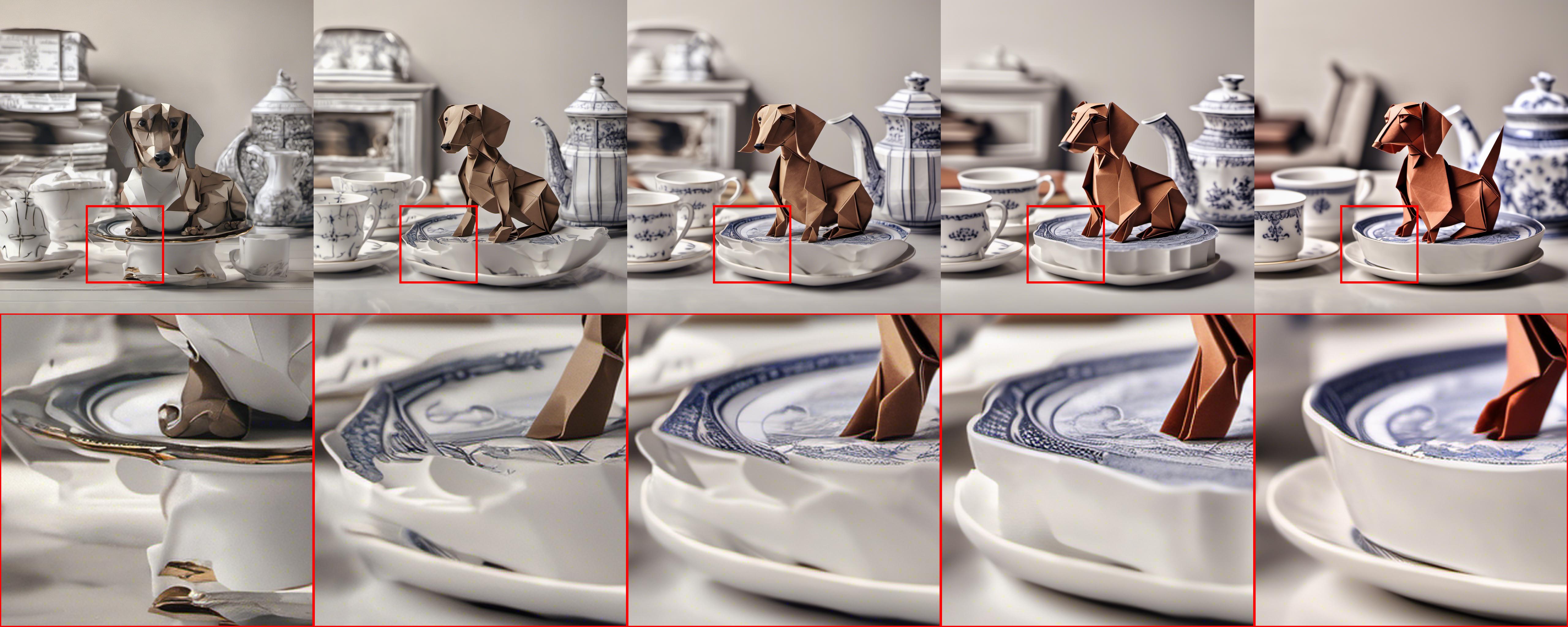}\\[-0.5mm]%
        \makebox[\hh][c]{\small\emph{A highly detailed paper origami of a Dachshund on a table next to a porcelain teapot, 4k dslr.}}\\[1mm]%
        \makebox[\hhhh][c]{$\gweight = 1$}%
        \makebox[\hhhh][c]{$\gweight = 3$}%
        \makebox[\hhhh][c]{$\gweight = 5$}\hfill%
        \makebox[\hhhh][c]{$\gweight = 1$}%
        \makebox[\hhhh][c]{$\gweight = 3$}%
        \makebox[\hhhh][c]{$\gweight = 5$}\\%
        \includegraphics[width=\hhh]{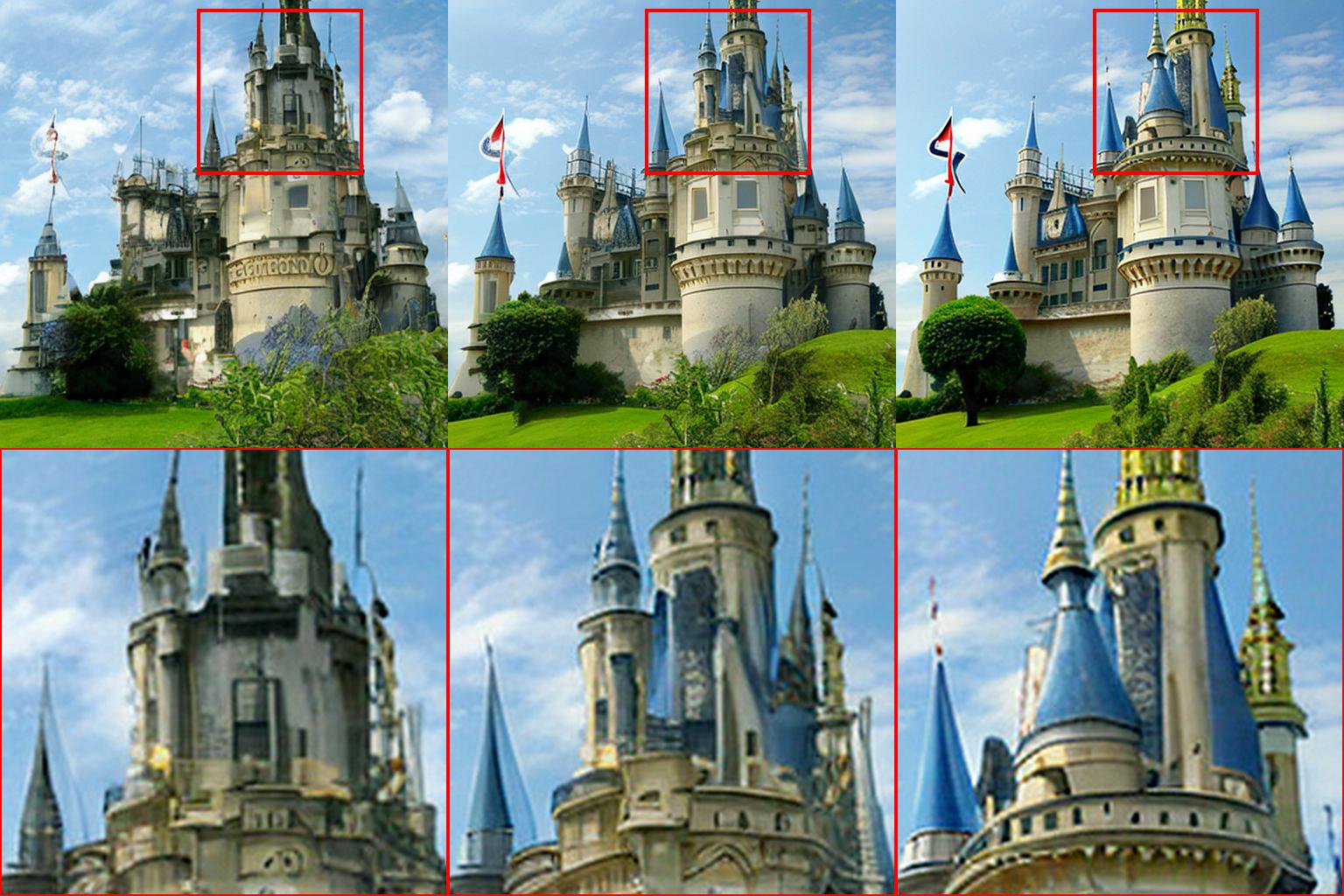}\hfill%
        \includegraphics[width=\hhh]{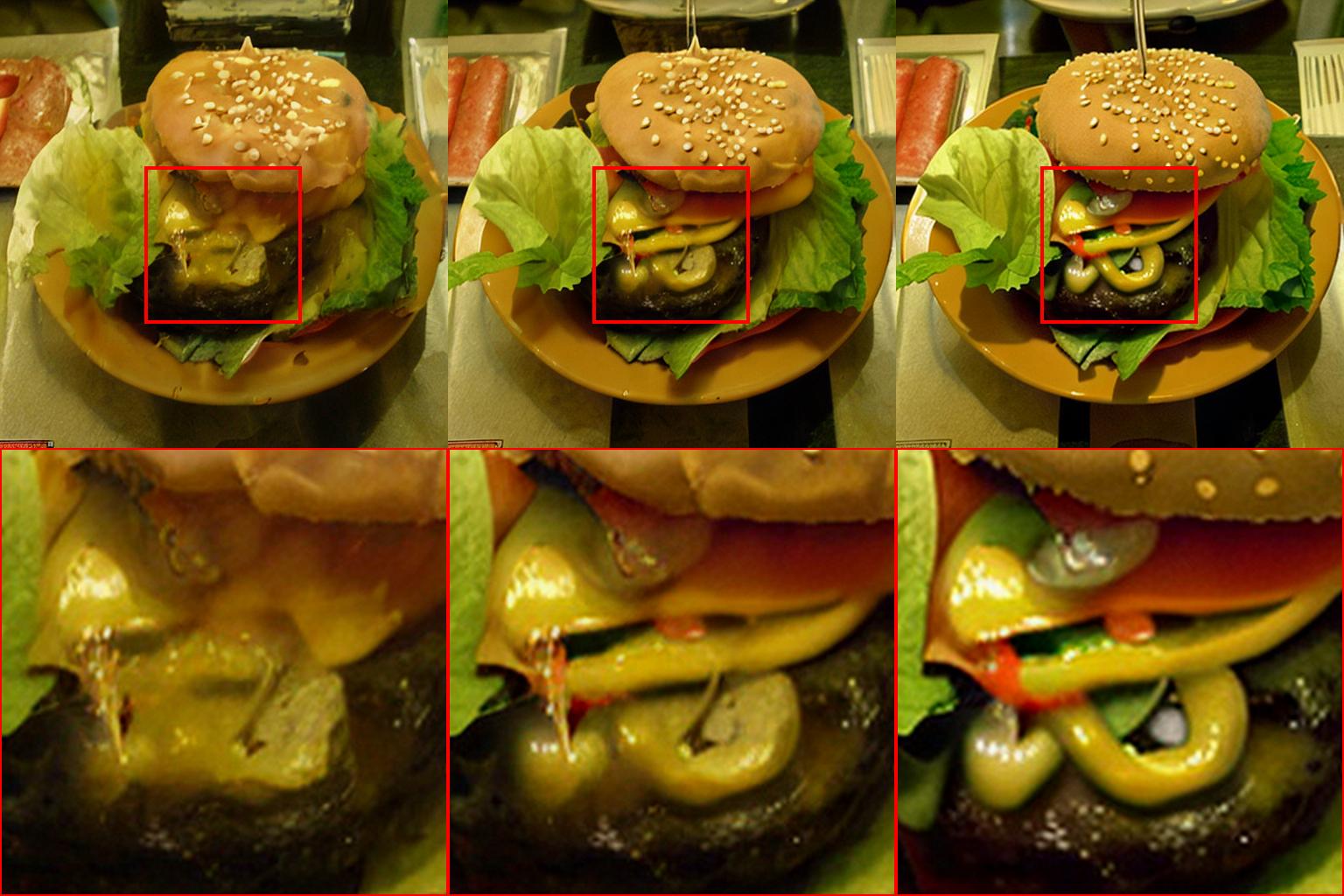}\\[-0.5mm]%
        \makebox[\hhh][c]{\small ImageNet class 483: \emph{castle}}\hfill%
        \makebox[\hhh][c]{\small ImageNet class 933: \emph{cheeseburger}}%
        \caption{%
            Effect of guidance weight $\gweight$ with our method.
            We limit the guidance to $\sigma \in \left(0.28,5.42\right]$ with SD-XL (top) and to $\sigma \in \left(0.19,1.61\right]$ with EDM2-XXL (bottom).
            Higher $\gweight$ leads to clearer and more well-defined image details while keeping the color palette and overall composition unchanged.
        }\vspace{-3mm}%
        \label{fig:sdxl_imagenet_qualitative_results_B}
    \end{figure}
}

\newcommand{\figQualitativeResultsC}{
    \renewcommand{\h}{0.2\linewidth}
    \renewcommand{\hh}{\linewidth}
    \begin{figure}[p]%
        \makebox[\h][c]{$\sigmahi = \infty$}%
        \makebox[\h][c]{$\sigmahi = 8.39$}%
        \makebox[\h][c]{$\sigmahi = \underline{5.42}$}%
        \makebox[\h][c]{$\sigmahi = 4.24$}%
        \makebox[\h][c]{$\sigmahi = 1.75$}\\
        \includegraphics[width=\hh]{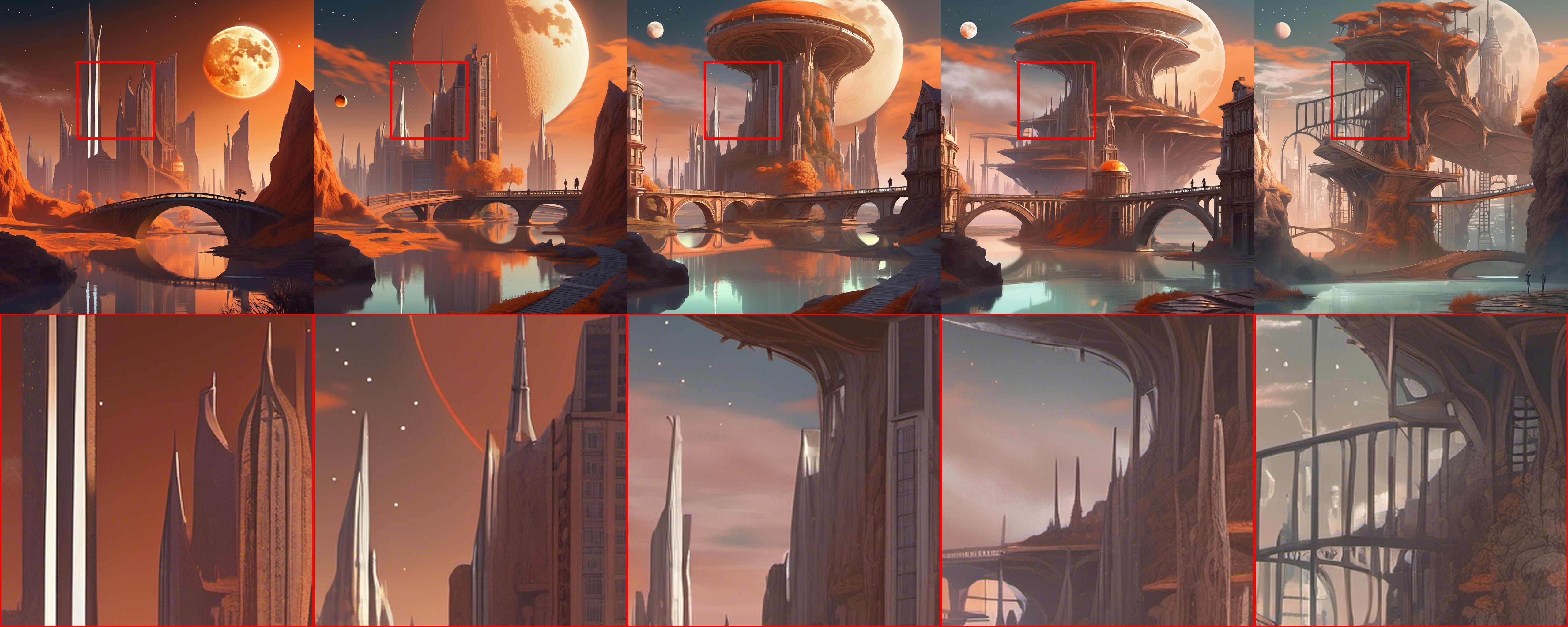}\\%
        \makebox[\h][c]{$\sigmalo = 0$}%
        \makebox[\h][c]{$\sigmalo = \underline{0.28}$}%
        \makebox[\h][c]{$\sigmalo = 1.00$}%
        \makebox[\h][c]{$\sigmalo = 2.42$}%
        \makebox[\h][c]{$\sigmalo = 3.72$}\\
        \includegraphics[width=\hh]{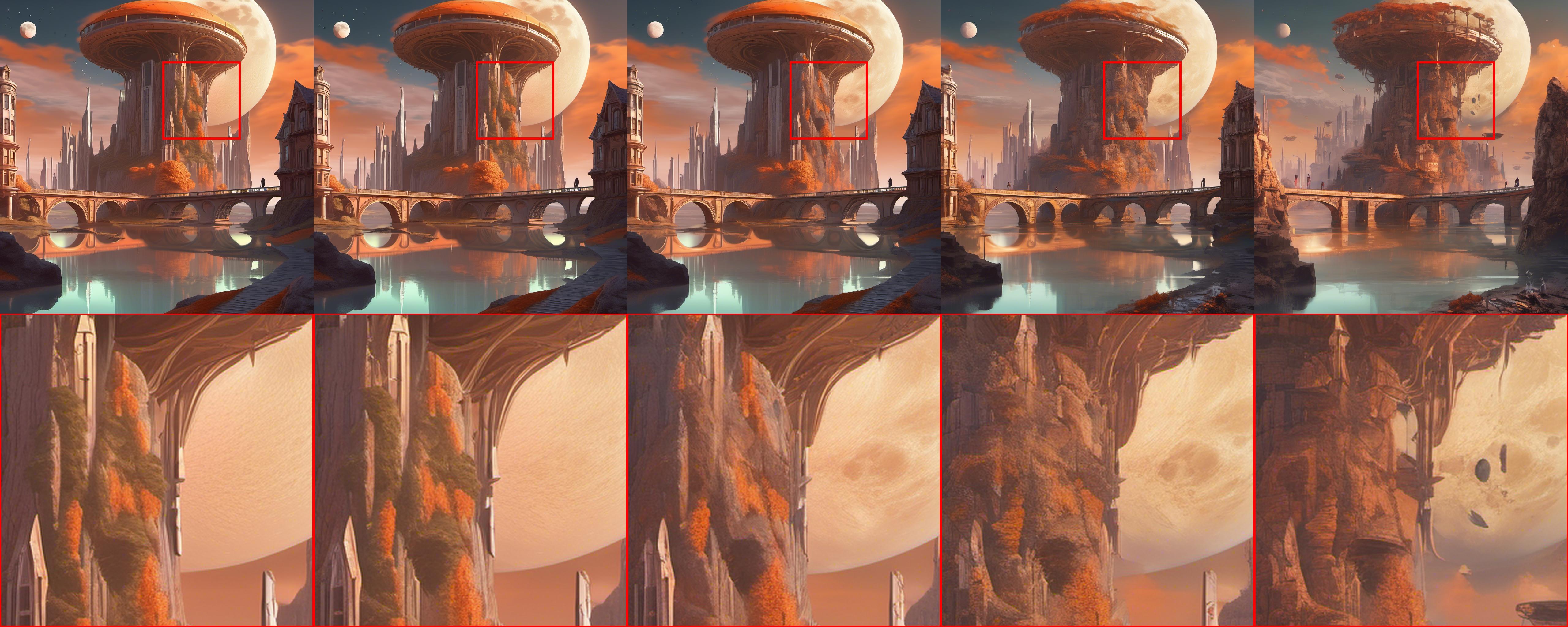}\\[0.5mm]%
        \makebox[\hh][c]{\begin{minipage}{0.9\linewidth}\small\emph{A fantasy landscape on an alien planet in which there are many buildings. There is a beautiful bridge with a pond in the center. There is one large moon in the sky. The sky is orange. Digital art, artstation}\end{minipage}}%
        \caption{%
            Effect of changing the guidance interval $\left(\sigmalo,\sigmahi\right]$ with $\gweight = 16$.
            \textbf{Top:} Decreasing $\sigmahi$, i.e., disabling guidance at high noise levels, while keeping $\sigmalo = 0.28$.
            High values lead to simplified image composition and oversaturated colors (left); low values cause the image to become increasingly convoluted (right).
            \textbf{Bottom:} Increasing $\sigmalo$, i.e., disabling guidance at low noise levels, while keeping $\sigmahi = 5.42$.
            The value can be made relatively high with no noticeable impact, reducing sampling cost.
        }%
        \label{fig:sdxl_qualitative_results_C}
    \end{figure}
}

\newcommand{\figQualitativeResultsD}{
    \renewcommand{\h}{4mm}
    \renewcommand{\hh}{0.48\linewidth}
    \renewcommand{\hhh}{0.160\linewidth}
    \begin{figure}[p]%
        \makebox[0.5\linewidth][c]{\ \ \ \ \ \ CFG}\hfill%
        \makebox[0.5\linewidth][c]{\,\ \ Ours, $\sigma \in \left(0.19, 1.61\right]$}\\
        \makebox[\h][l]{\rotatebox[origin=l]{90}{\makebox[0mm][c]{}}}%
        \makebox[\hhh][c]{$\gweight = 1$}%
        \makebox[\hhh][c]{$\gweight = 3$}%
        \makebox[\hhh][c]{$\gweight = 5$}\hfill%
        \makebox[\hhh][c]{$\gweight = 1$}%
        \makebox[\hhh][c]{$\gweight = 3$}%
        \makebox[\hhh][c]{$\gweight = 5$}\\%
        \makebox[\h][l]{\rotatebox[origin=l]{90}{\makebox[0mm][c]{\hspace*{42.5mm}\small ImageNet class 483: \emph{castle}}}}%
        \includegraphics[width=\hh]{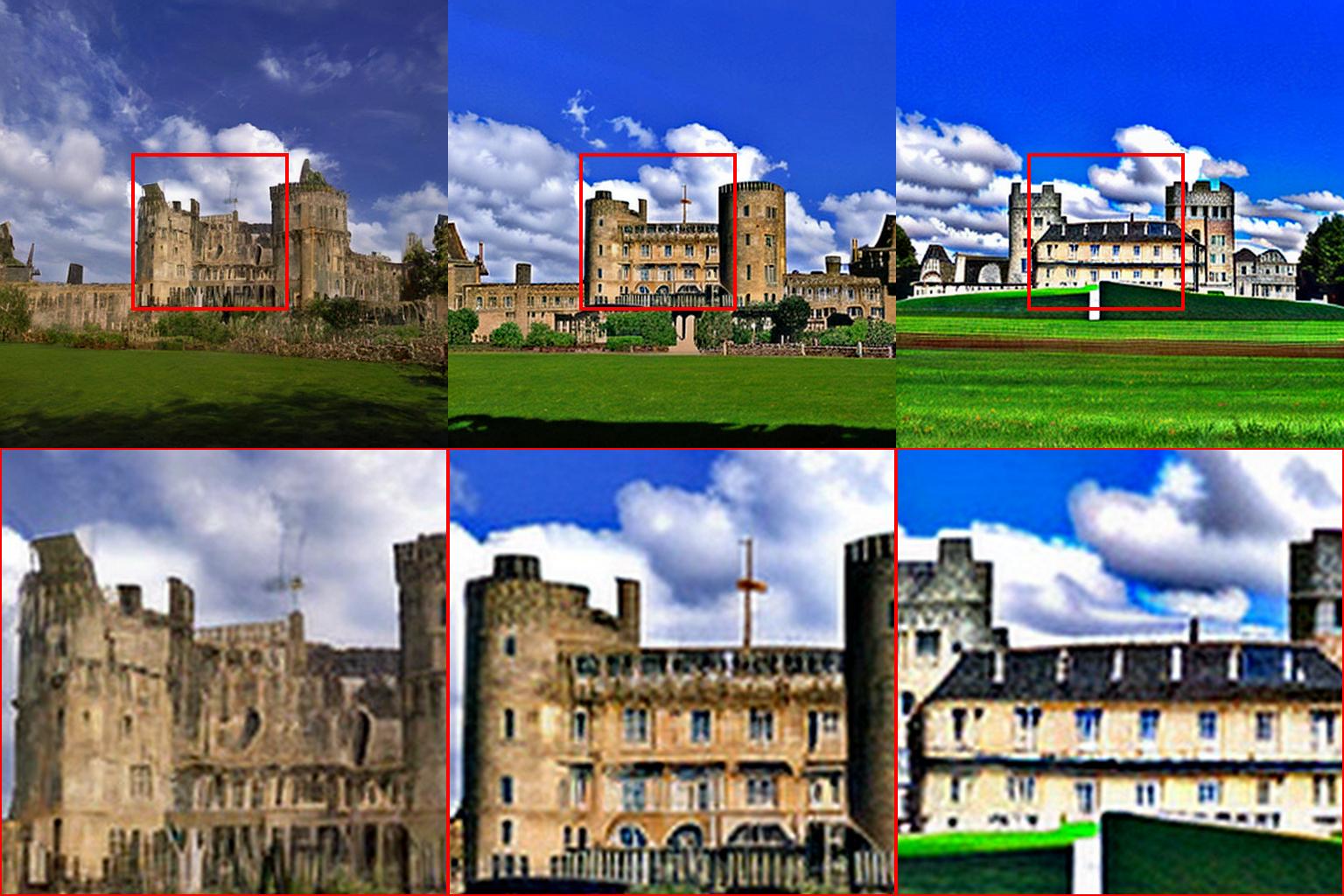}\hfill%
        \includegraphics[width=\hh]{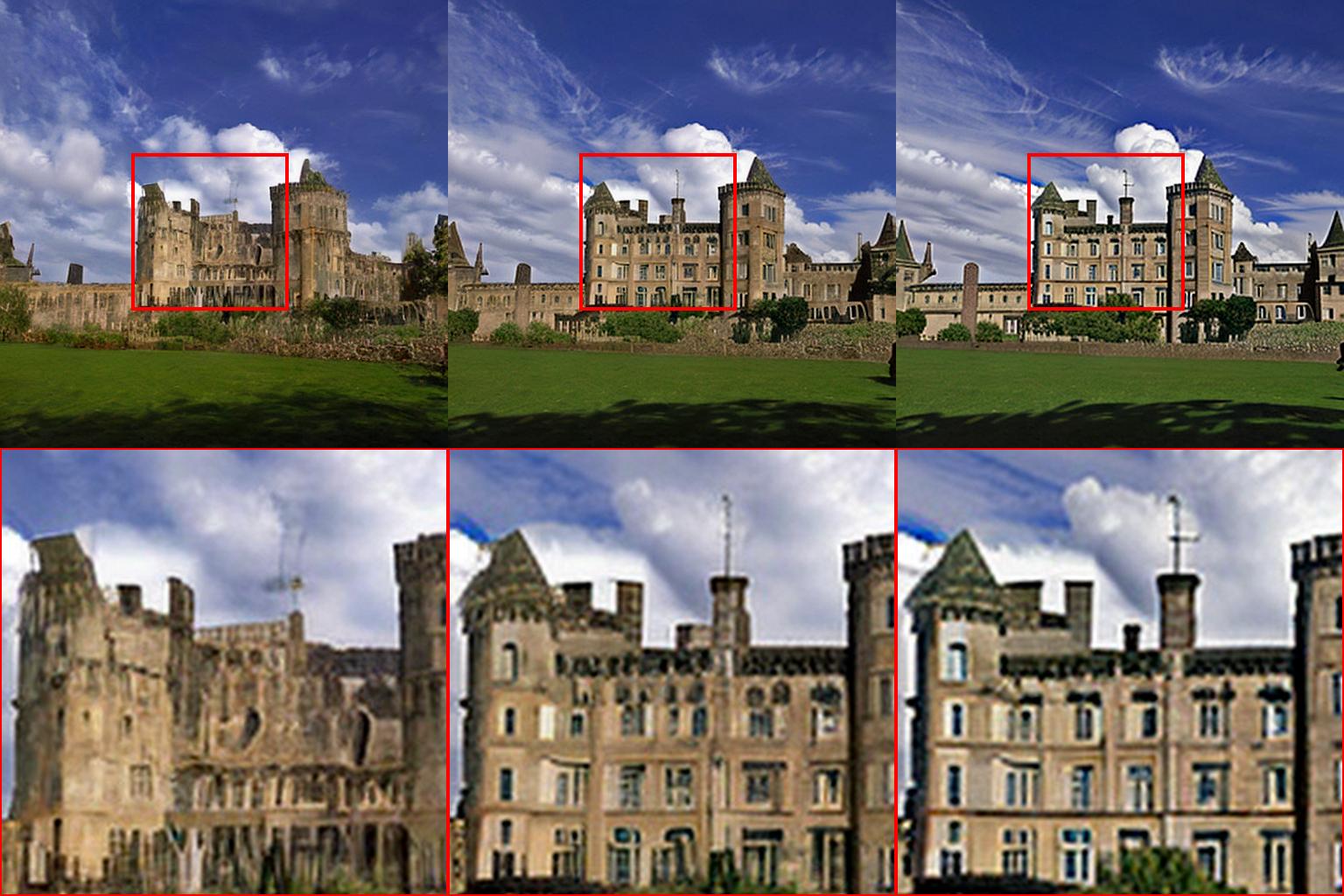}%
        \caption{Effect of increasing guidance weight $\gweight$ with CFG vs.~our method. \textbf{Left:} Increasing the guidance weight with CFG leads to changes in image composition and contrast. \textbf{Right:} With our method, increasing $\gweight$ improves image details but retains the overall composition and realistic colors.}
        \label{fig:qualitative_results_D}
    \end{figure}
}

\newcommand{\figQualitativeResultsAImageNetSupp}{
    \renewcommand{\h}{0.313\linewidth}
    \renewcommand{\hh}{4mm}
    \begin{figure}[t]
        \makebox[\hh][l]{}%
        \makebox[\h]{CFG with low guidance}\hfill%
        \makebox[\h]{CFG with high guidance}\hfill%
        \makebox[\h]{Ours with high guidance}\\%
        \makebox[\hh][l]{}%
        \makebox[\h]{\small $\gweight=1$\,\,\,(no guidance)}\hfill%
        \makebox[\h]{\small $\gweight=5$,\,\,\,$\sigma\in\left(0,\infty\right)$}\hfill%
        \makebox[\h]{\small $\gweight=5$,\,\,\,$\sigma\in\left(0.19,1.61\right]$}\\%
        \makebox[\hh]{}%
        \makebox[\h]{\small{\color{red}fuzzy details}, {\color{darkgreen}high diversity}}\hfill%
        \makebox[\h]{\small{\color{darkgreen}crisp details}, {\color{red}low diversity}}\hfill%
        \makebox[\h]{\small{\color{darkgreen}crisp details}, {\color{darkgreen}high diversity}}\\%
        \makebox[\hh][l]{\rotatebox[origin=l]{90}{\makebox[0mm][c]{\hspace*{65mm}\small ImageNet class 17: \emph{jay}}}}%
        \includegraphics[width=\h]{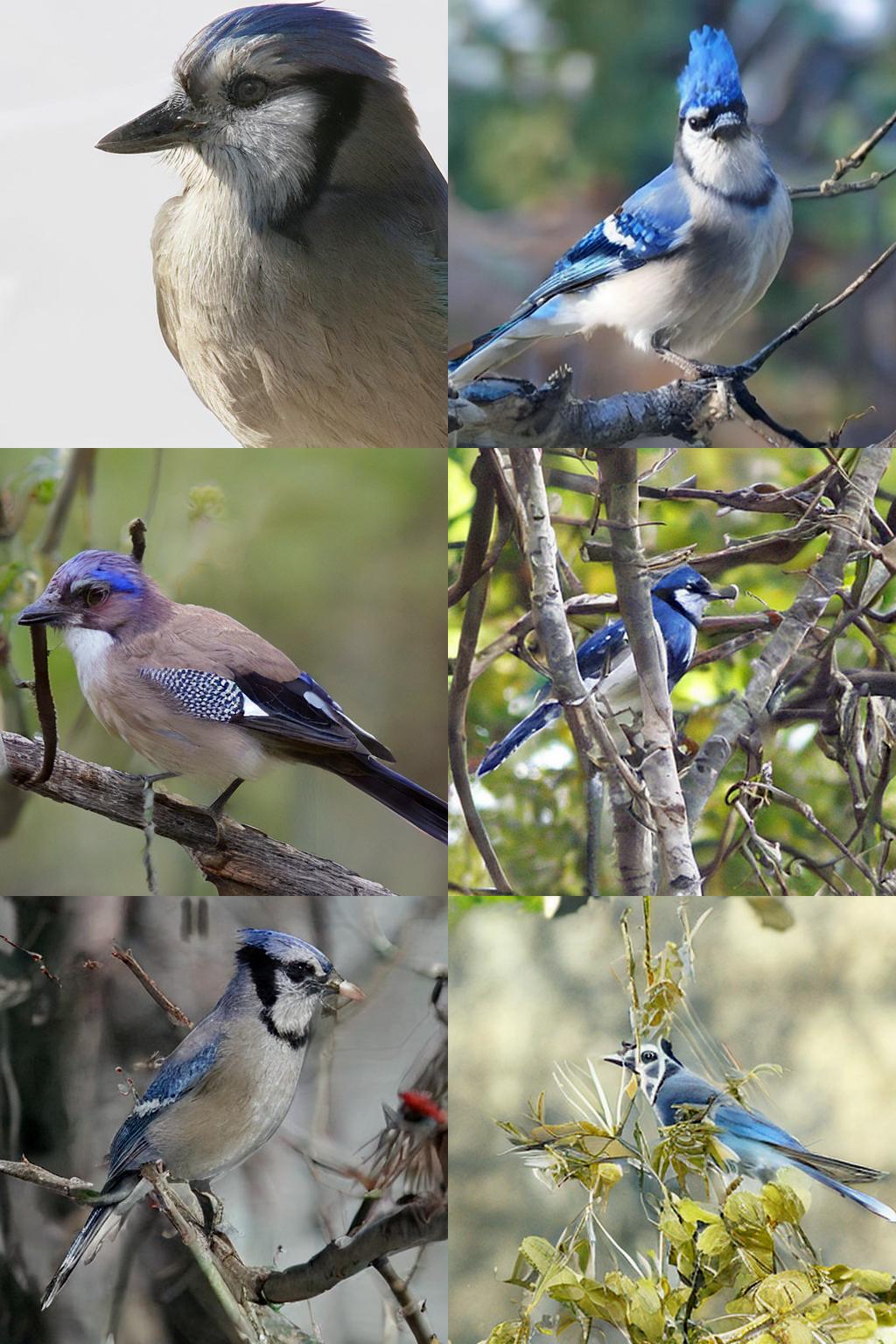}\hfill%
        \includegraphics[width=\h]{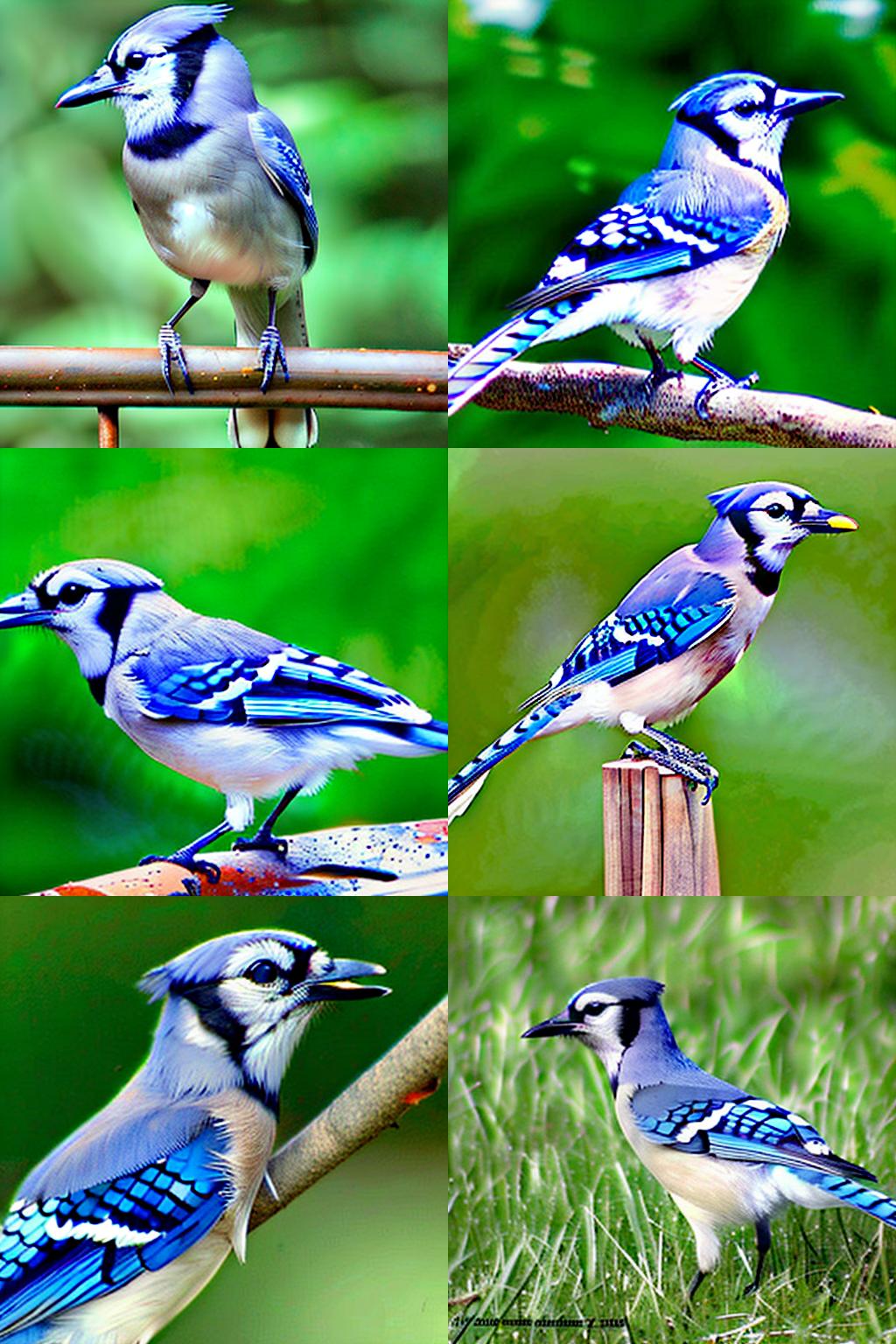}\hfill%
        \includegraphics[width=\h]{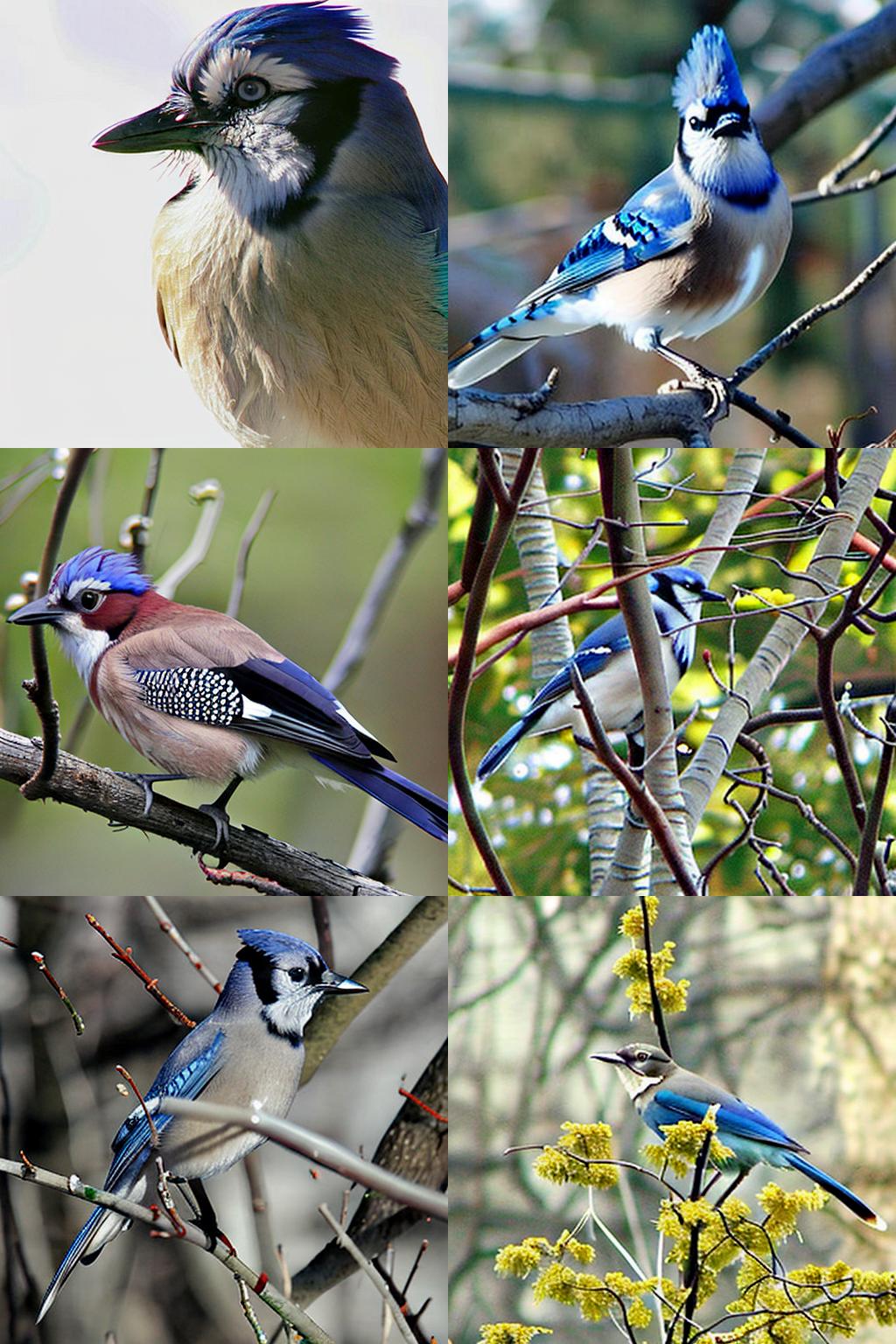}\hfill\\
        \makebox[\hh][l]{\rotatebox[origin=l]{90}{\makebox[0mm][c]{\hspace*{65mm}\small ImageNet class 33: \emph{loggerhead}}}}%
        \includegraphics[width=\h]{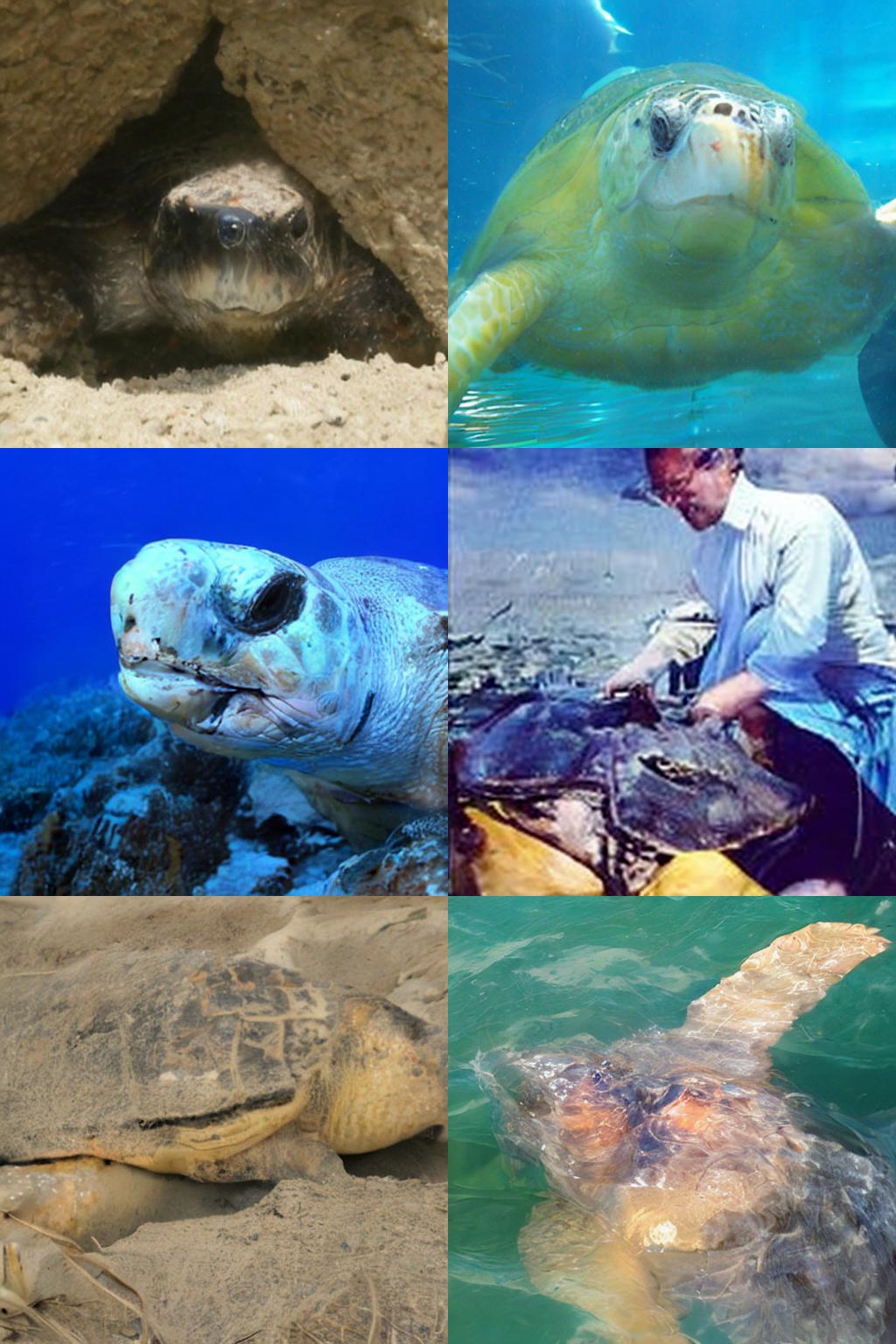}\hfill%
        \includegraphics[width=\h]{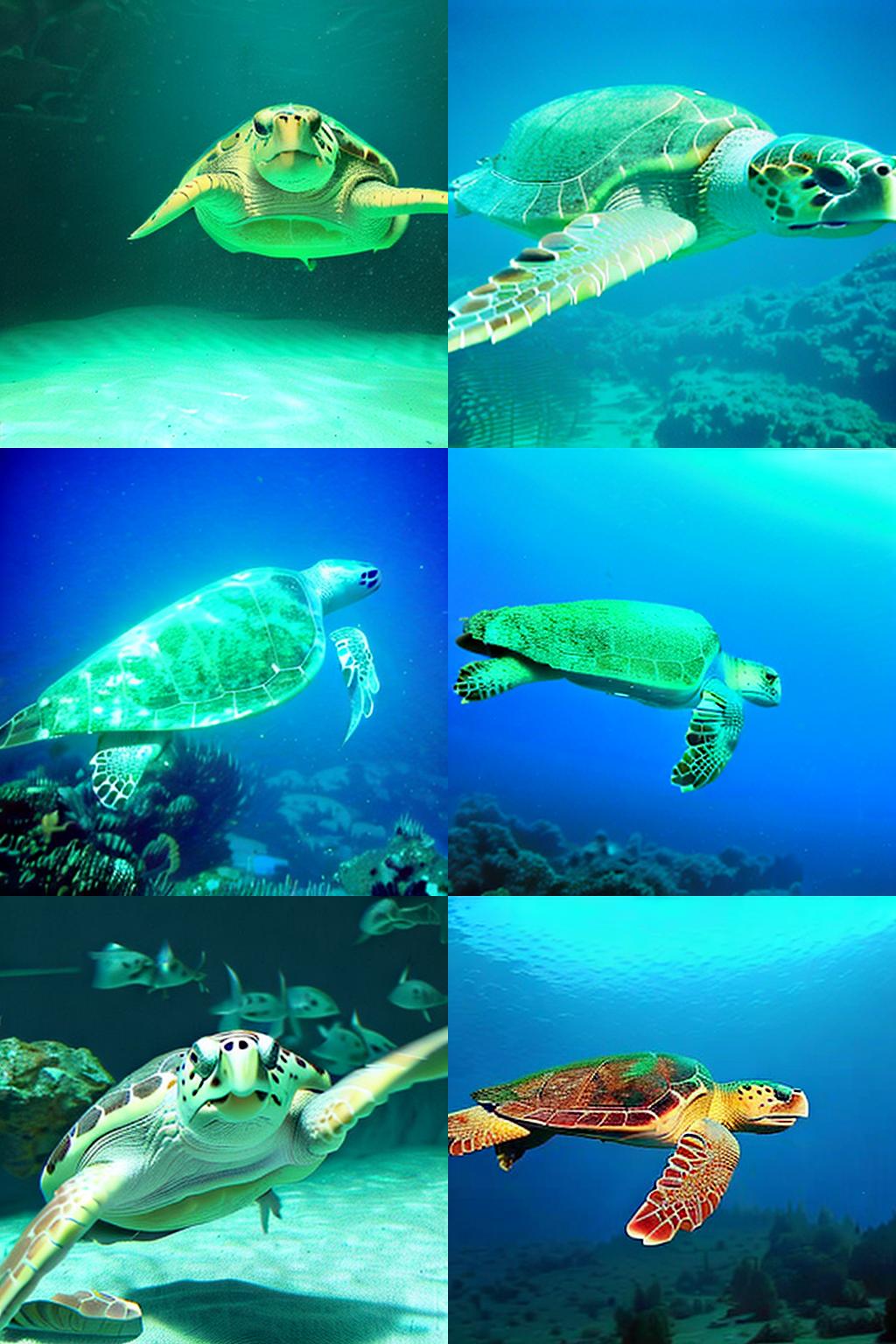}\hfill%
        \includegraphics[width=\h]{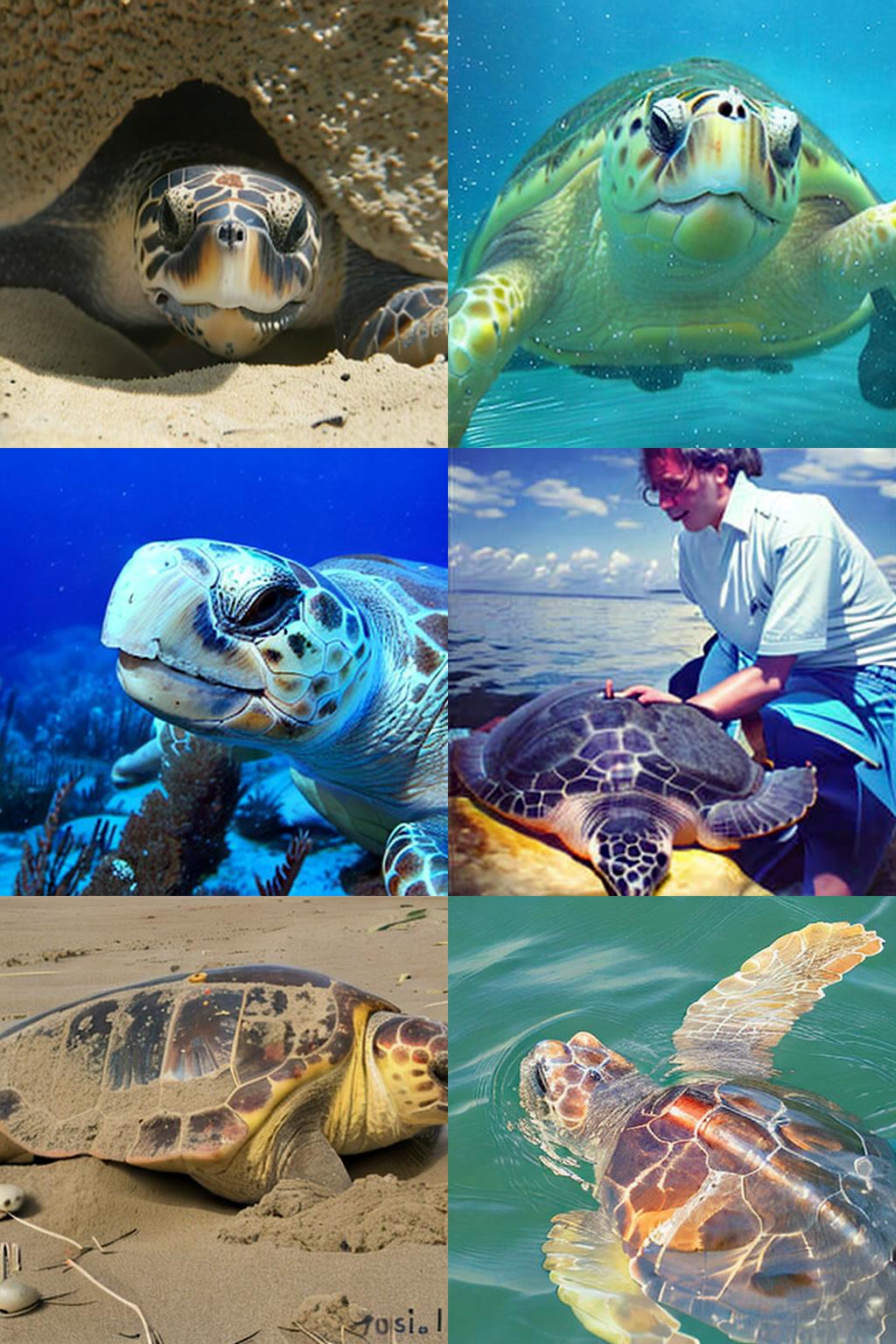}\\
        \makebox[\hh][l]{\rotatebox[origin=l]{90}{\makebox[0mm][c]{\hspace*{65mm}\small ImageNet class 497: \emph{church}}}}%
        \includegraphics[width=\h]{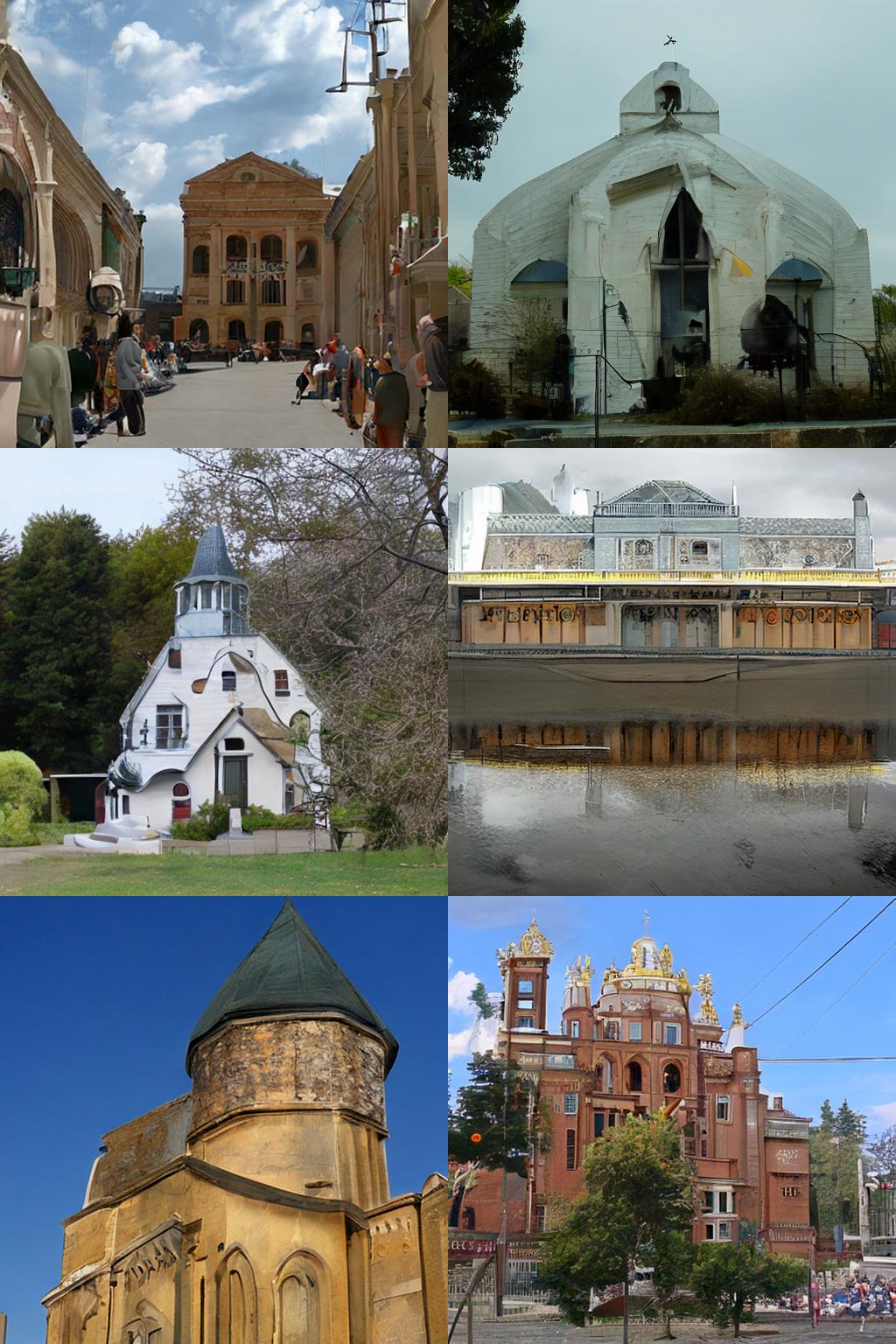}\hfill%
        \includegraphics[width=\h]{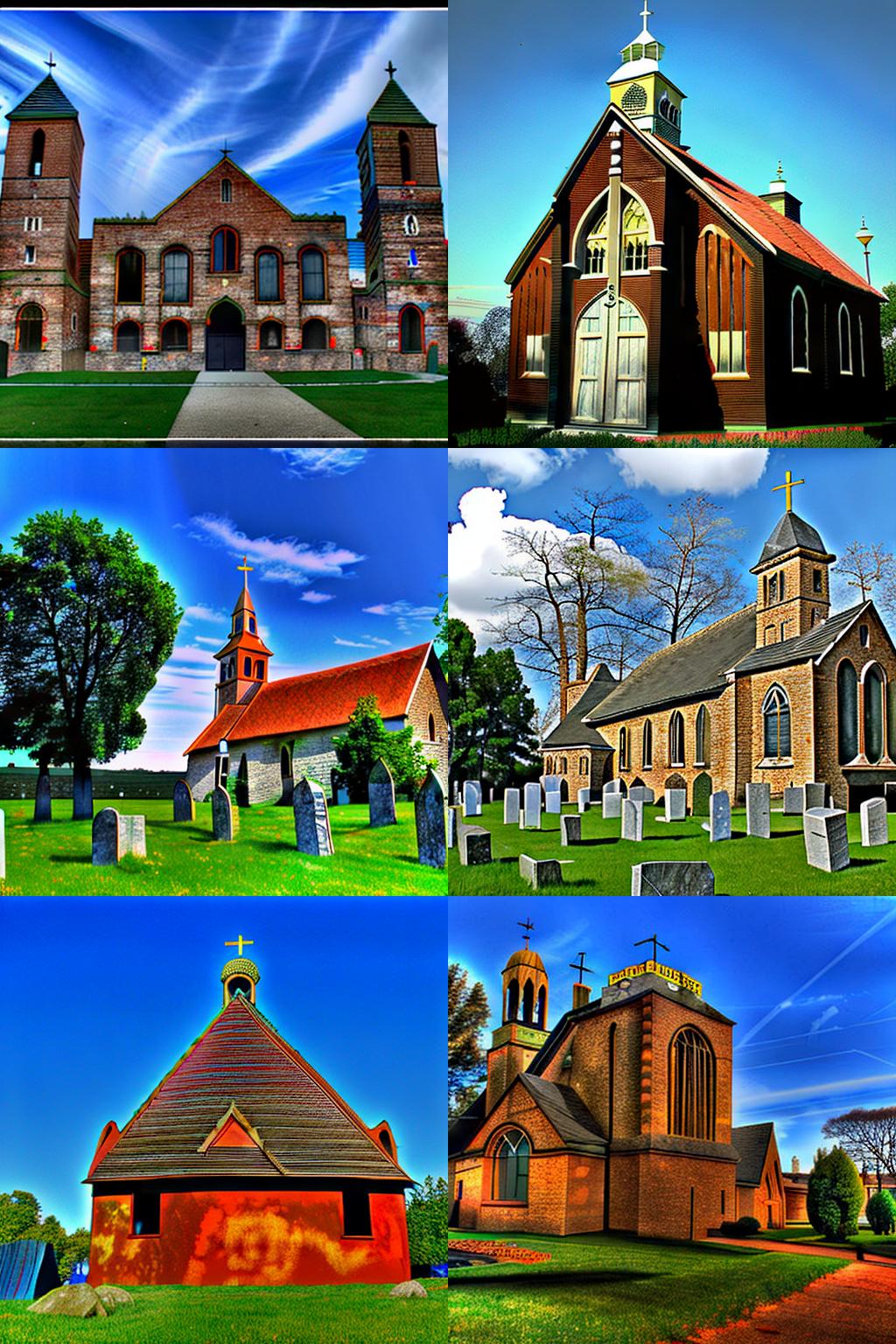}\hfill%
        \includegraphics[width=\h]{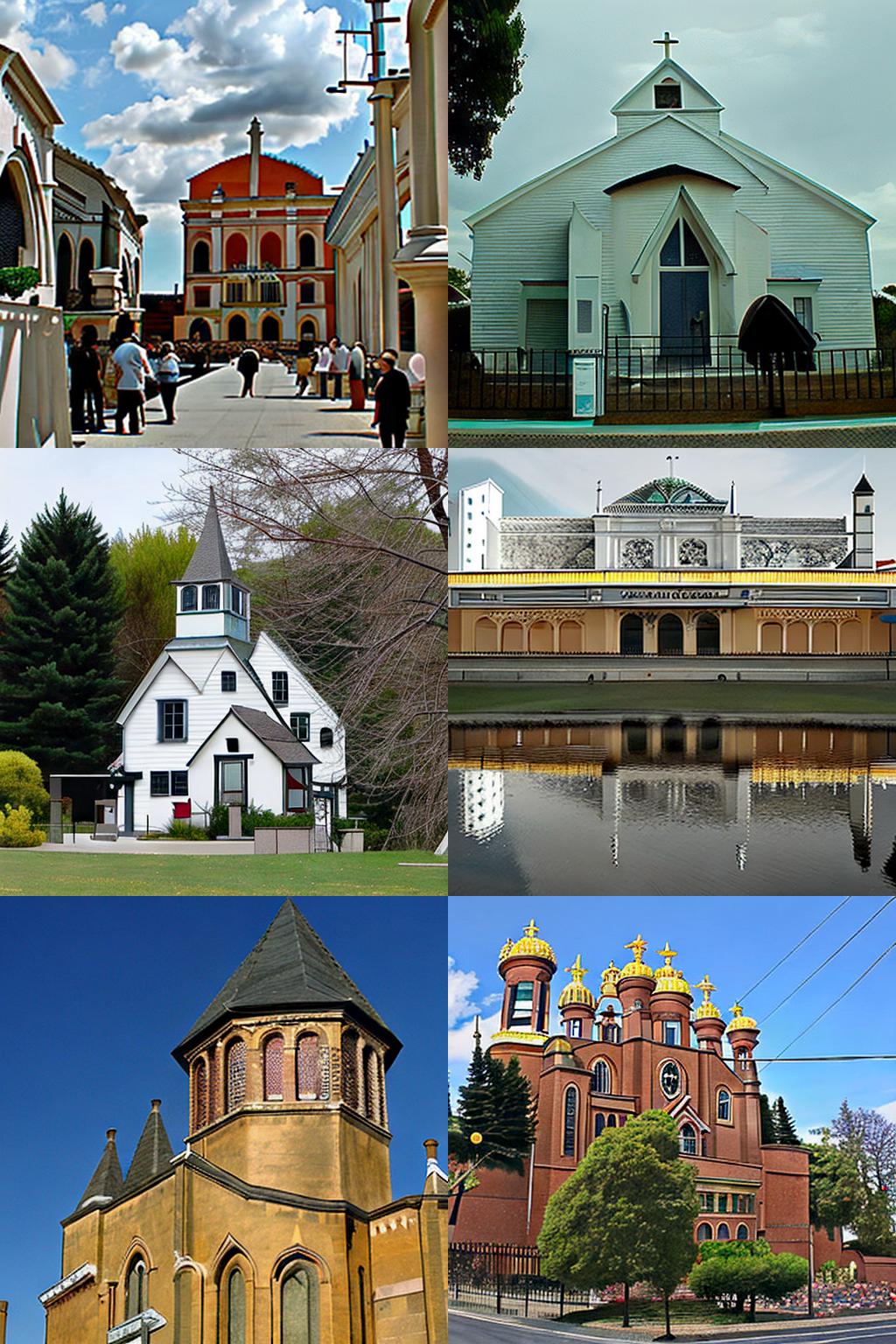}
        \caption{Additional EDM2-XXL results that demonstrate how CFG with low $\gweight$ yields fuzzy images that lack detail (left) and CFG with high $\gweight$ leads to reduced diversity and oversaturated colors. Our method (right) produces images with crisp details while maintaining natural colors.}
        \label{fig:sdxl_imagenet_qualitative_results_A_imagenet_supp}
    \end{figure}
}

\newcommand{\figQualitativeResultsASDXLSupp}{
    \renewcommand{\h}{0.3\linewidth}
    \renewcommand{\hh}{12mm}
    \begin{figure}[t]
        \makebox[\hh][l]{}%
        \makebox[\h]{CFG with low guidance}\hfill%
        \makebox[\h]{CFG with high guidance}\hfill%
        \makebox[\h]{Ours with high guidance}\\%
        \makebox[\hh][l]{}%
        \makebox[\h]{\small $\gweight=2$,\,\,\,$\sigma\in\left(0,\infty\right)$}\hfill%
        \makebox[\h]{\small $\gweight=16$,\,\,\,$\sigma\in\left(0,\infty\right)$}\hfill%
        \makebox[\h]{\small $\gweight=16$,\,\,\,$\sigma\in\left(0.28,5.42\right]$}\\%
        \makebox[\hh]{}%
        \makebox[\h]{\small{\color{red}fuzzy details}, {\color{darkgreen}high diversity}}\hfill%
        \makebox[\h]{\small{\color{darkgreen}crisp details}, {\color{red}low diversity}}\hfill%
        \makebox[\h]{\small{\color{darkgreen}crisp details}, {\color{darkgreen}high diversity}}\\%
        \makebox[\hh][l]{\rotatebox[origin=l]{90}{\hspace*{1mm}\begin{minipage}{55mm}\small\emph{A pointillist painting of a raccoon looking at the sea.}\end{minipage}}}%
        \includegraphics[width=\h]{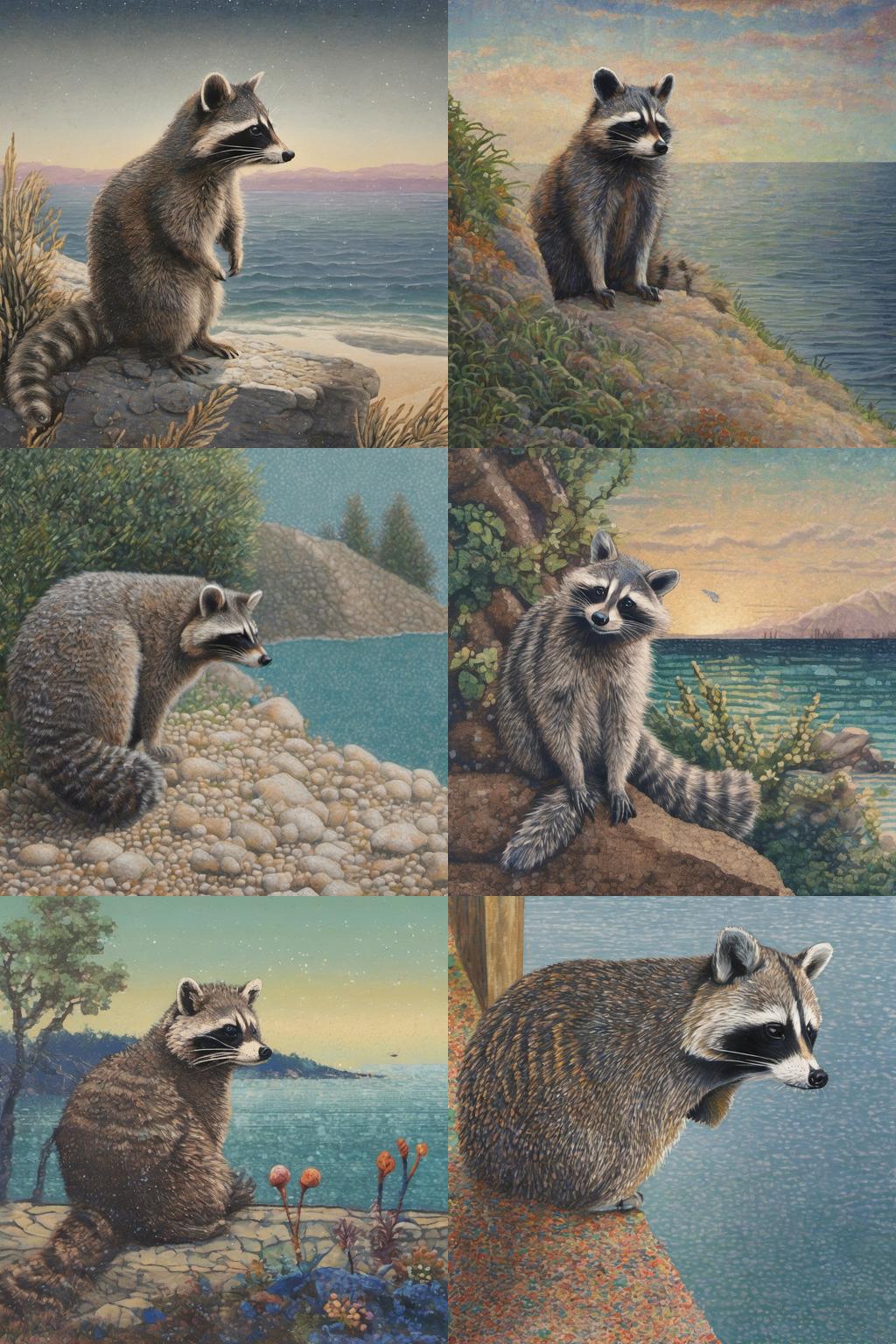}\hfill%
        \includegraphics[width=\h]{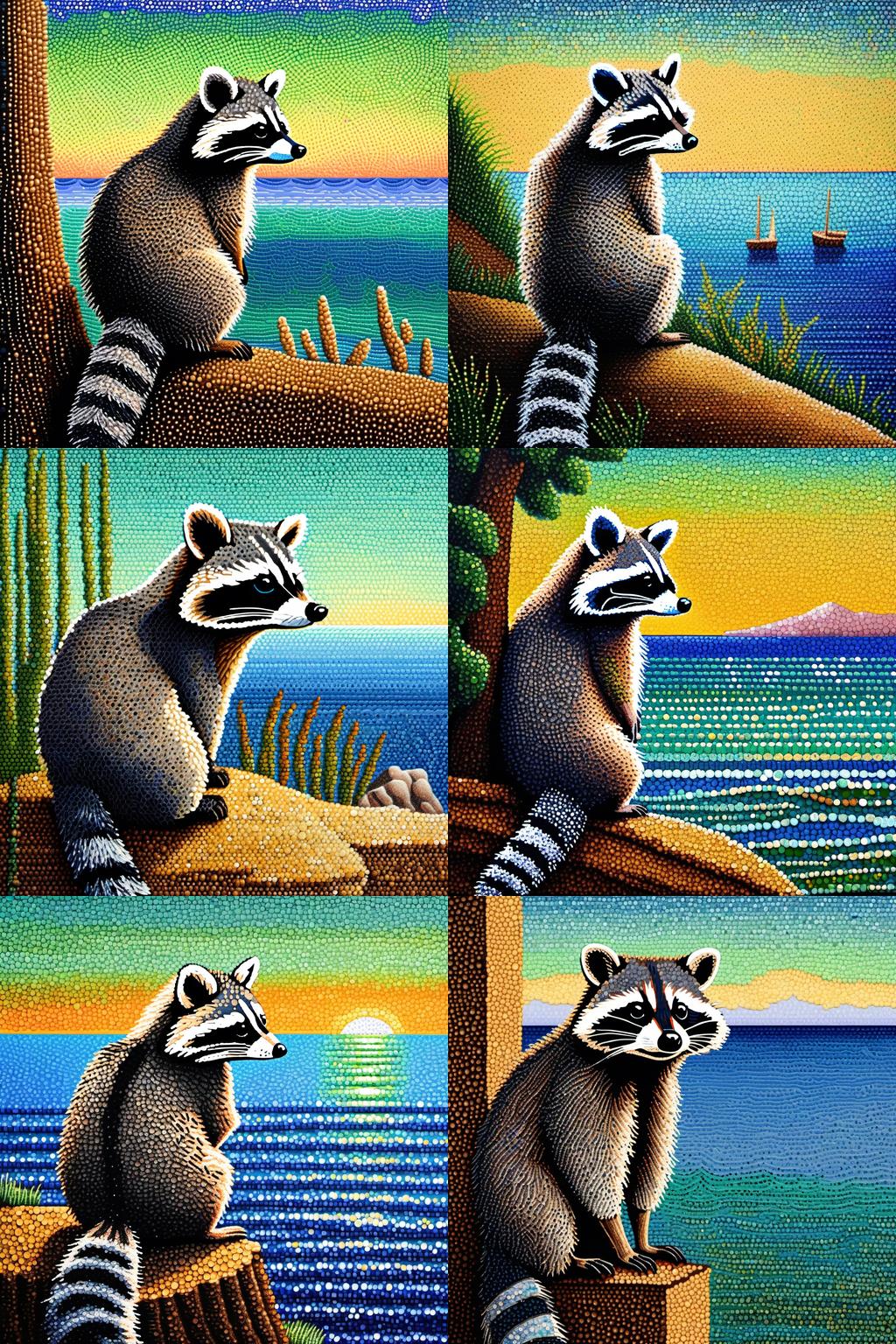}\hfill%
        \includegraphics[width=\h]{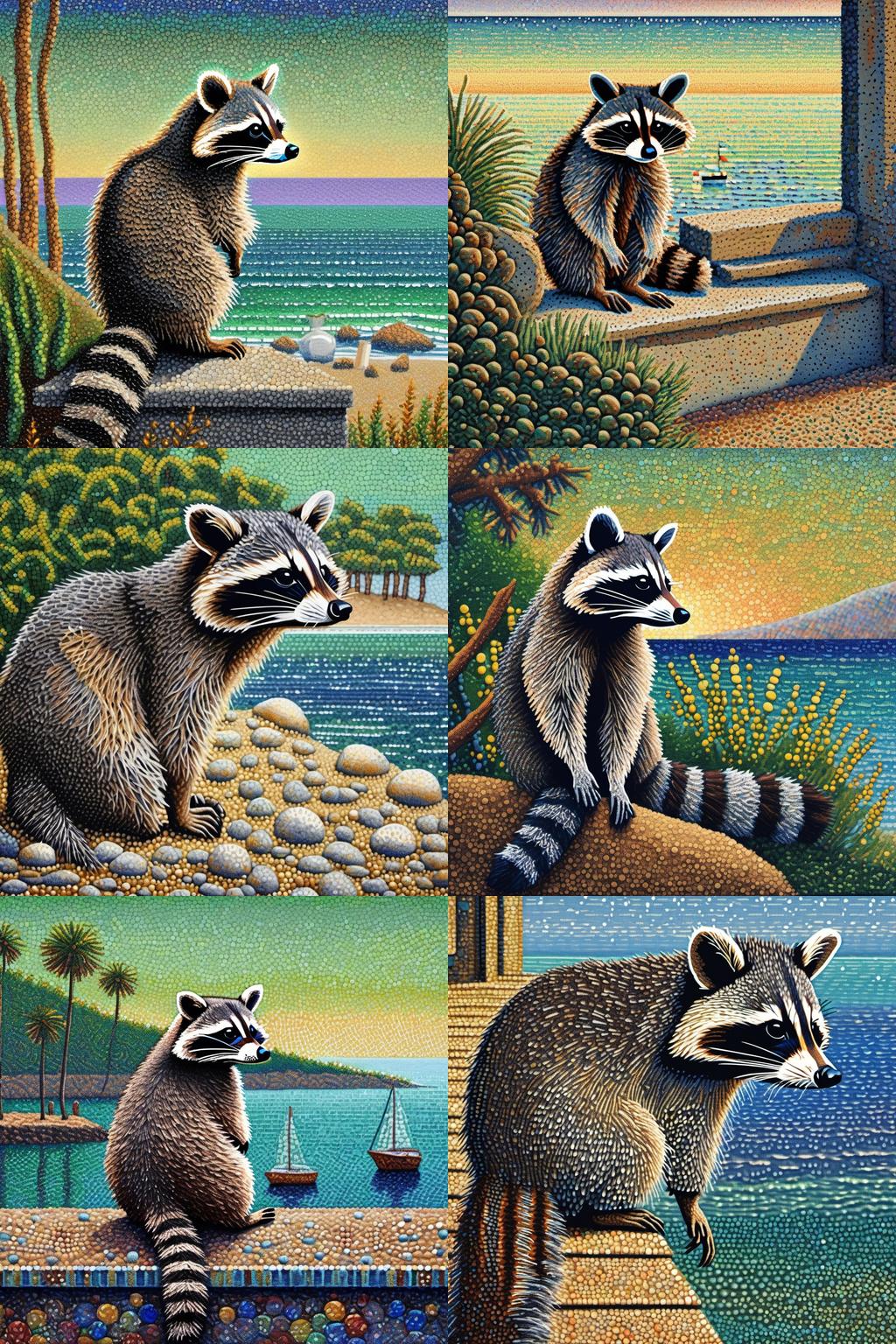}\\%
        \makebox[\hh][l]{\rotatebox[origin=l]{90}{\hspace*{1mm}\begin{minipage}{55mm}\small\emph{A wild west town with cowboys and saloons, set at sunset}\end{minipage}}}%
        \includegraphics[width=\h]{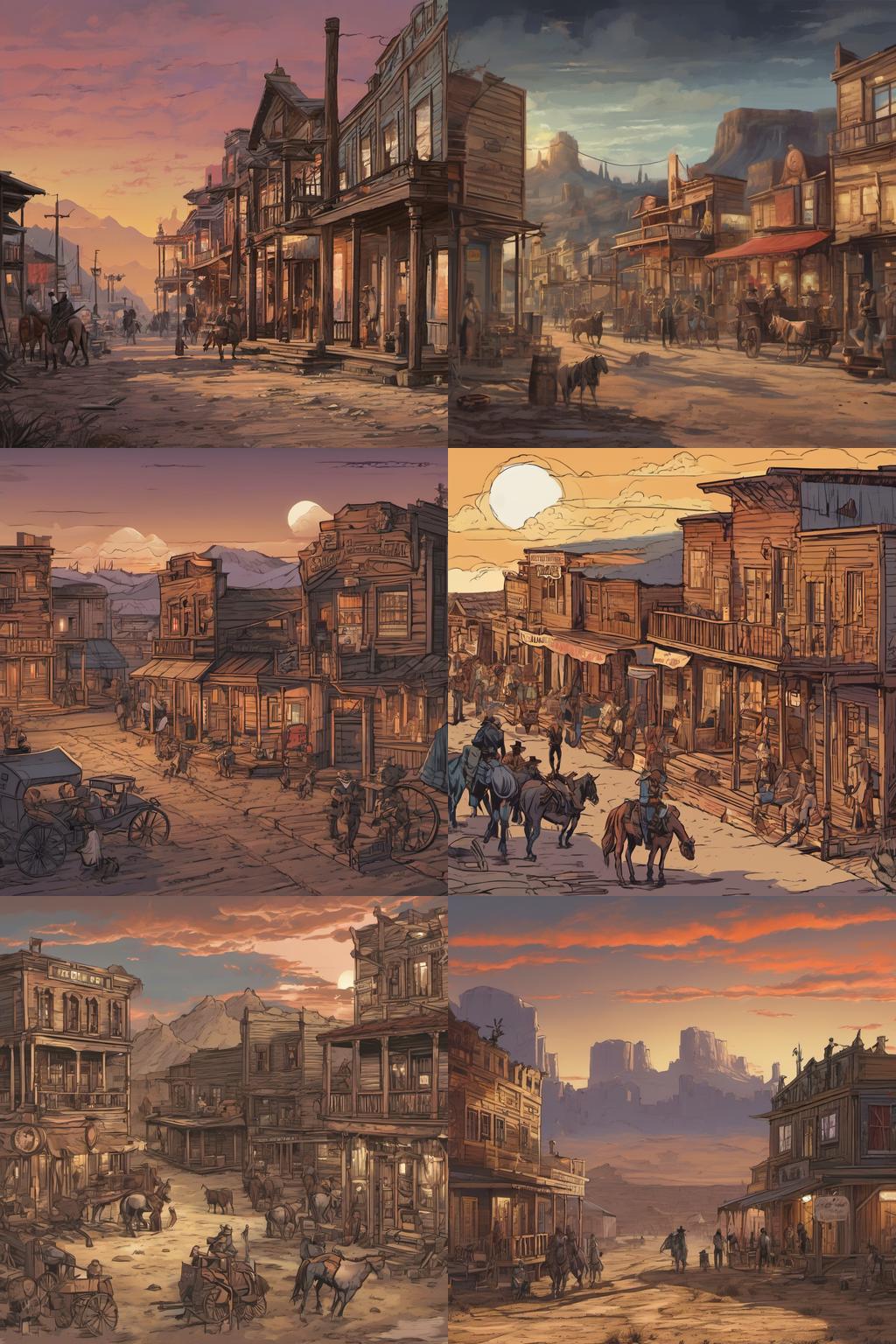}\hfill%
        \includegraphics[width=\h]{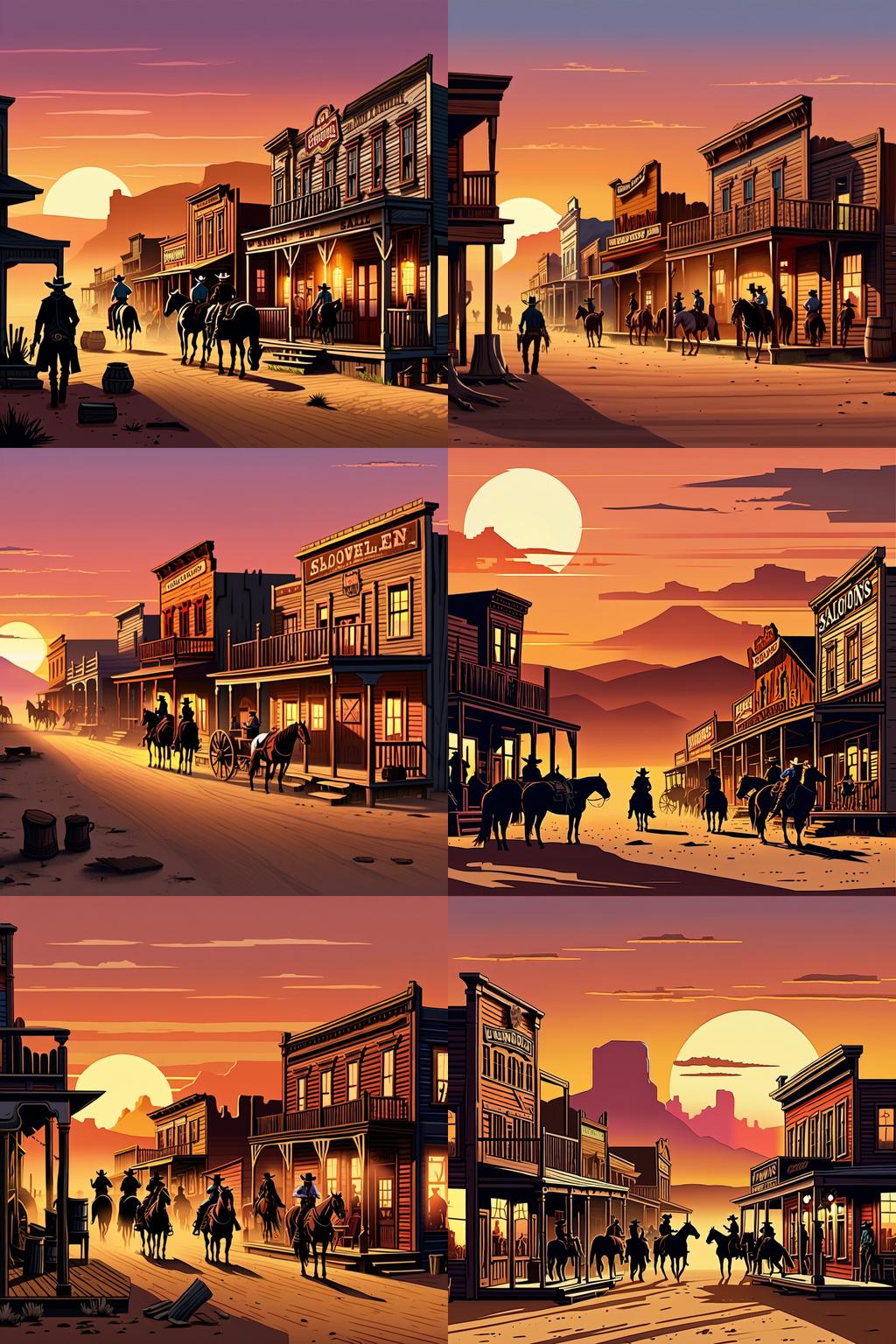}\hfill%
        \includegraphics[width=\h]{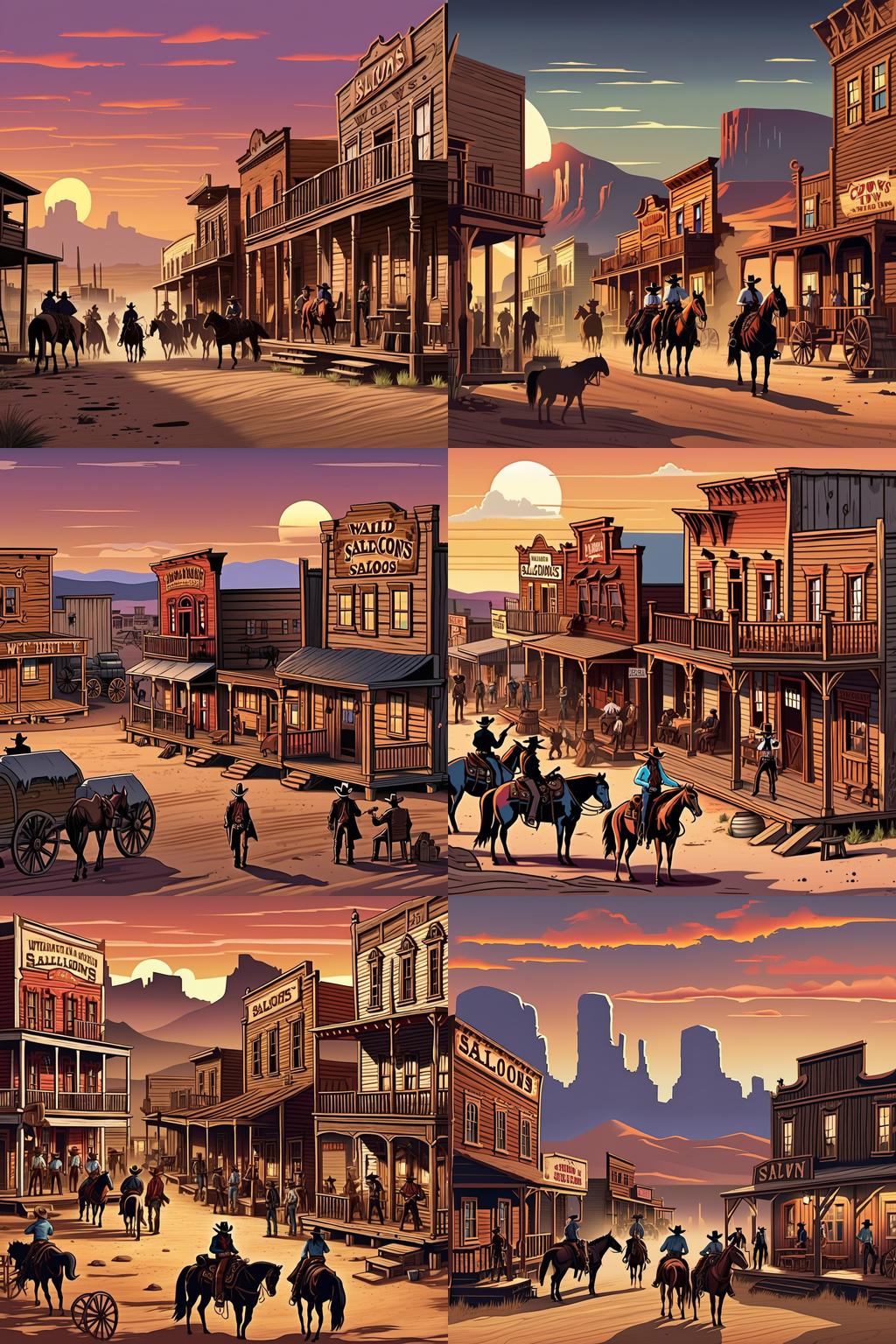}\\%
        \makebox[\hh][l]{\rotatebox[origin=l]{90}{\hspace*{1mm}\begin{minipage}{55mm}\small\emph{A blue jay standing on a large basket of rainbow macarons.}\end{minipage}}}%
        \includegraphics[width=\h]{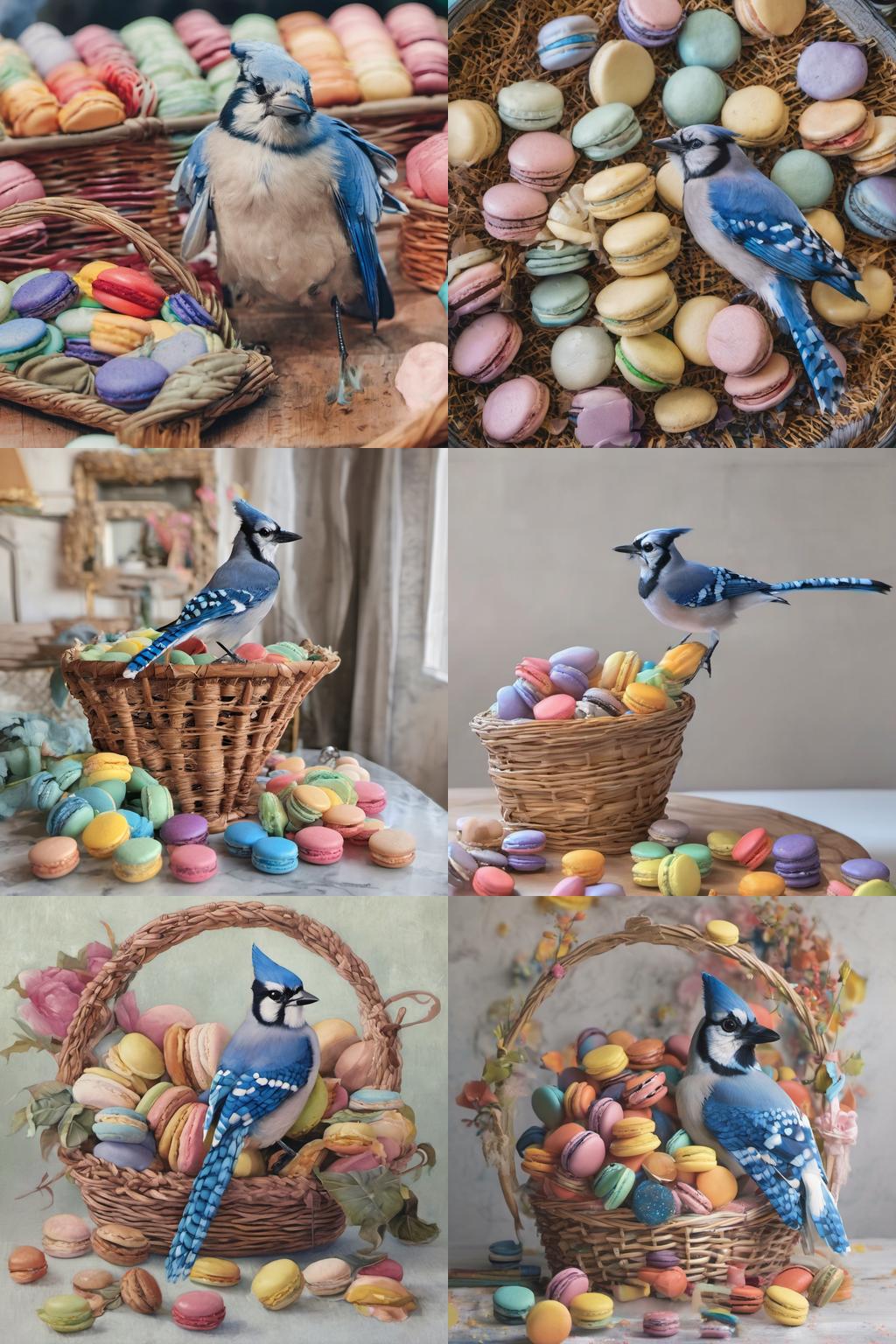}\hfill%
        \includegraphics[width=\h]{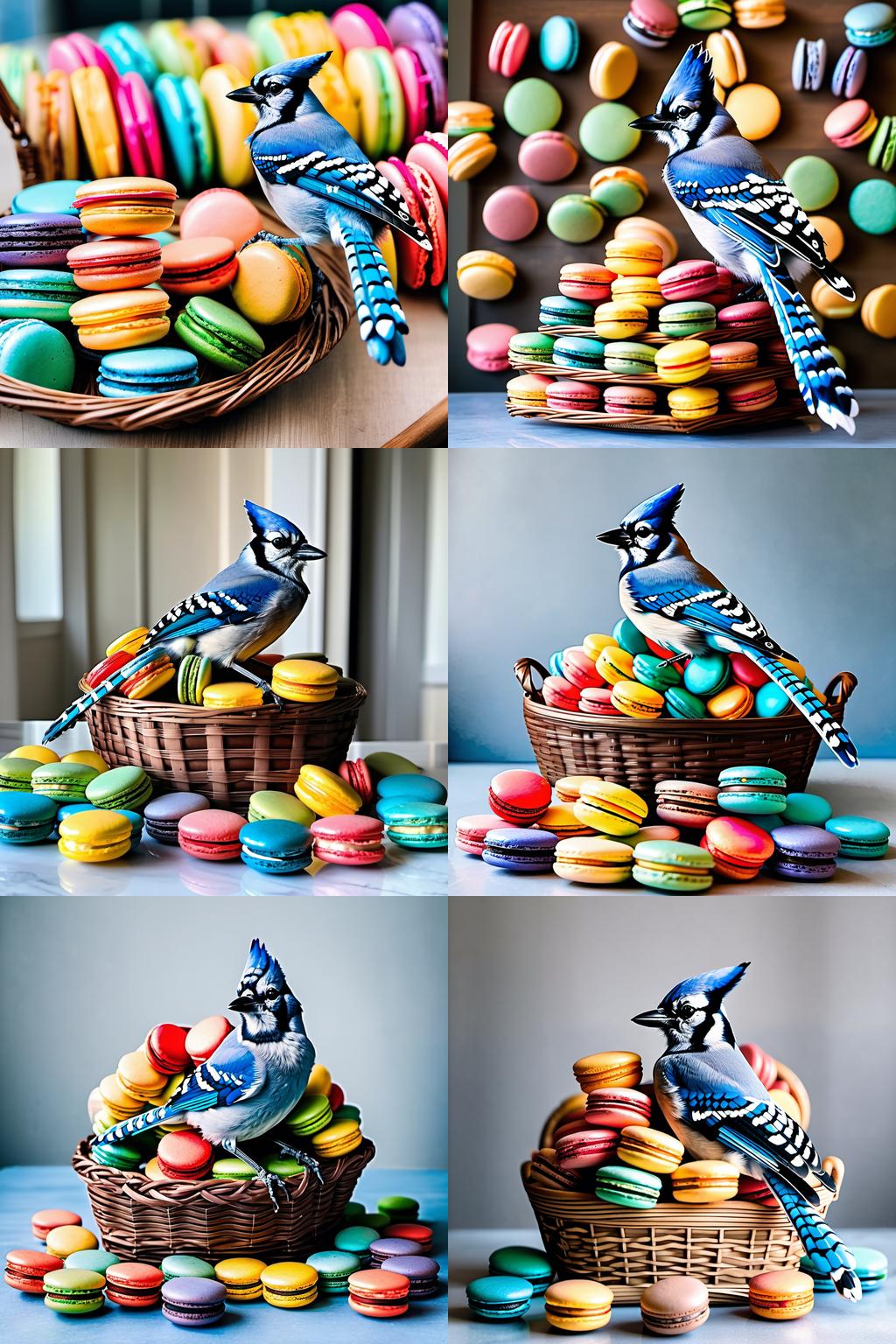}\hfill%
        \includegraphics[width=\h]{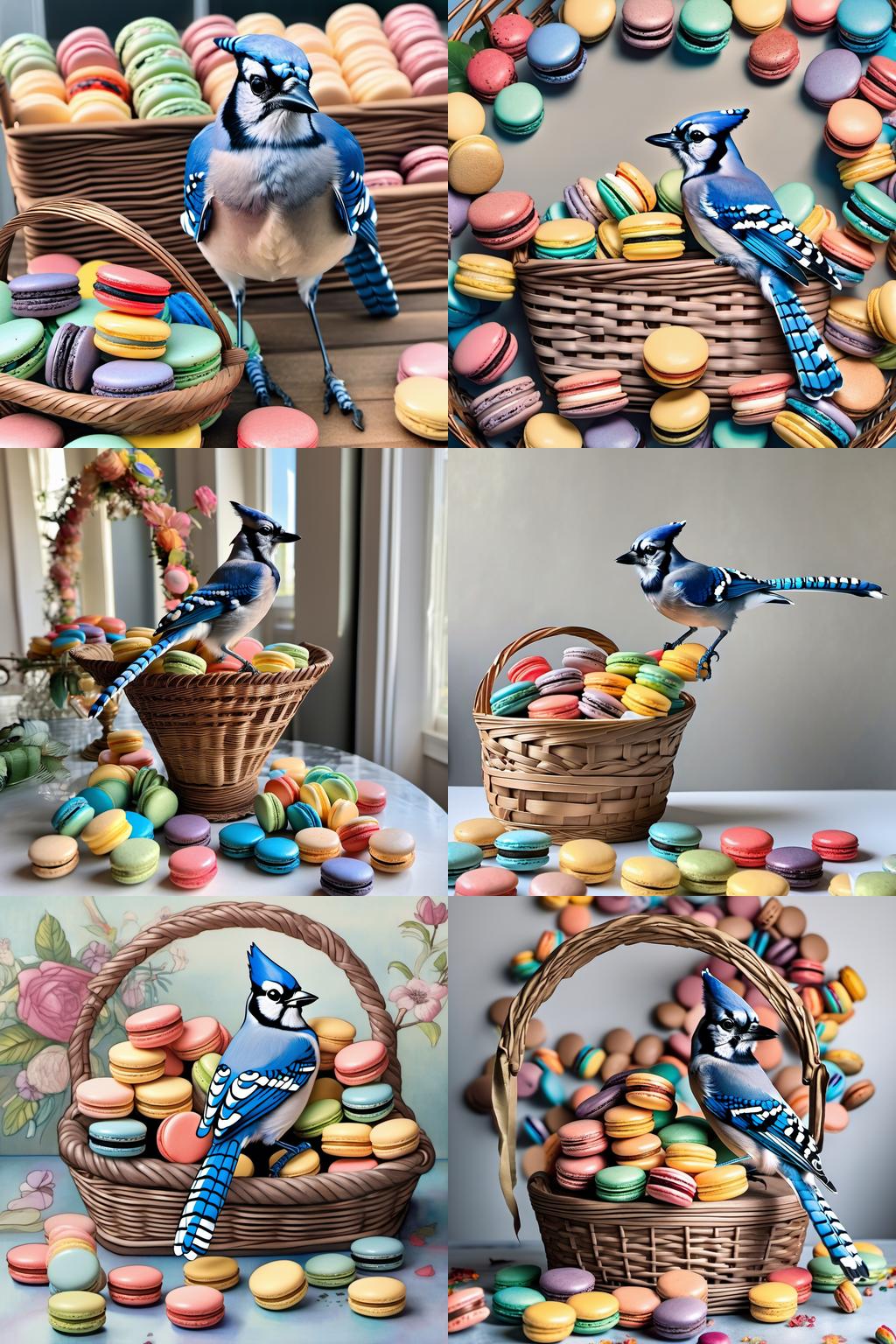}\\%
        \caption{More SD-XL results that demonstrate how CFG with low $\gweight$ yields fuzzy images that lack detail (left) and CFG with high $\gweight$ leads to reduced diversity and oversaturated colors. Our method (right) produces images with crisp details while maintaining natural colors. The degree of the negative effects with CFG varies between prompts.}
        \label{fig:sdxl_imagenet_qualitative_results_A_sdxl_supp}
    \end{figure}
}

\newcommand{\figQualitativeResultsBImageNetSupp}{
    \renewcommand{\h}{0.2\linewidth}
    \renewcommand{\hh}{\linewidth}
    \renewcommand{\hhh}{0.495\linewidth}
    \renewcommand{\hhhh}{0.167\linewidth}
    \begin{figure}[t]%
        \makebox[\hhhh][c]{$\gweight = 1$}%
        \makebox[\hhhh][c]{$\gweight = 3$}%
        \makebox[\hhhh][c]{$\gweight = 5$}\hfill%
        \makebox[\hhhh][c]{$\gweight = 1$}%
        \makebox[\hhhh][c]{$\gweight = 3$}%
        \makebox[\hhhh][c]{$\gweight = 5$}\\%
        \includegraphics[width=\hhh]{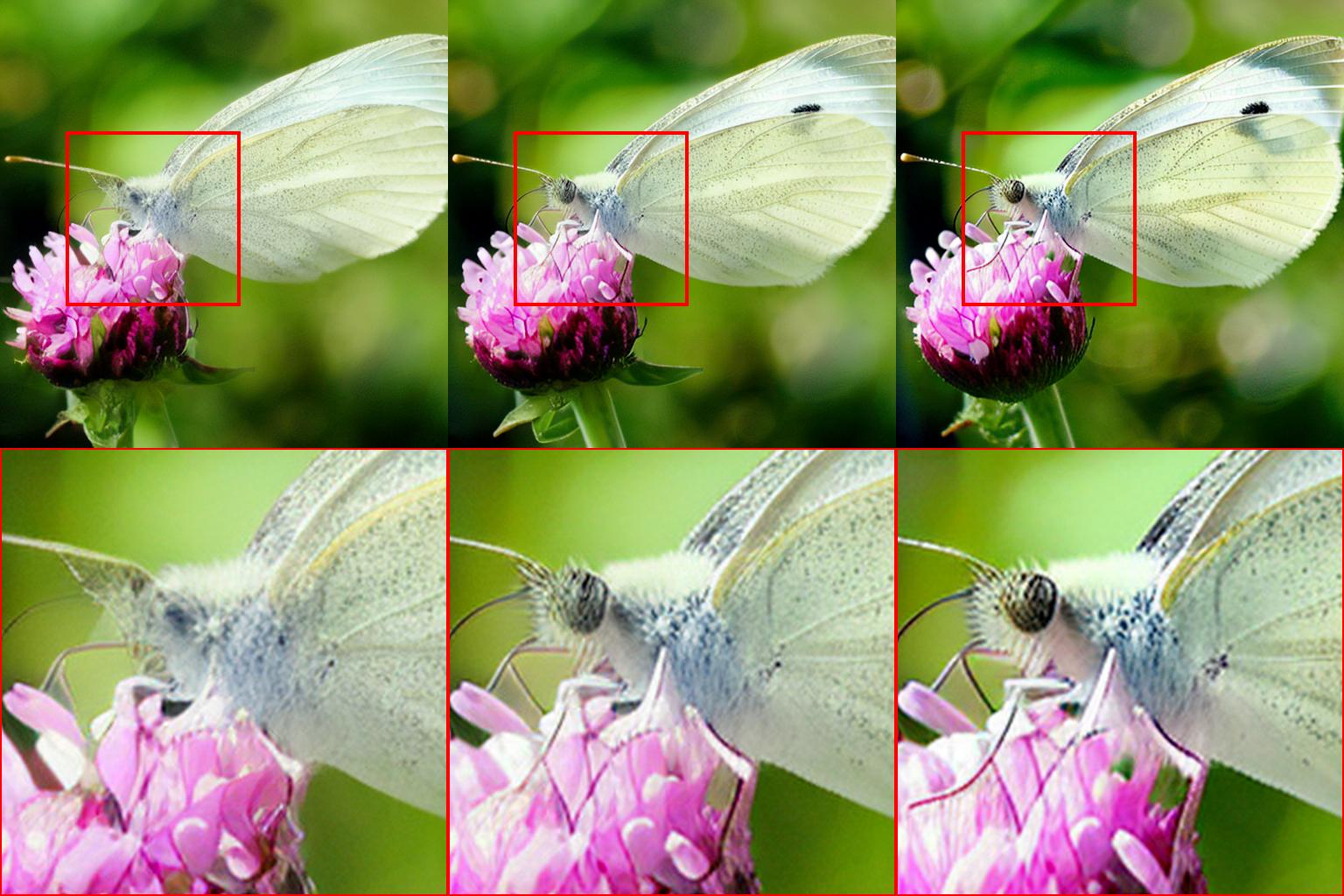}\hfill%
        \includegraphics[width=\hhh]{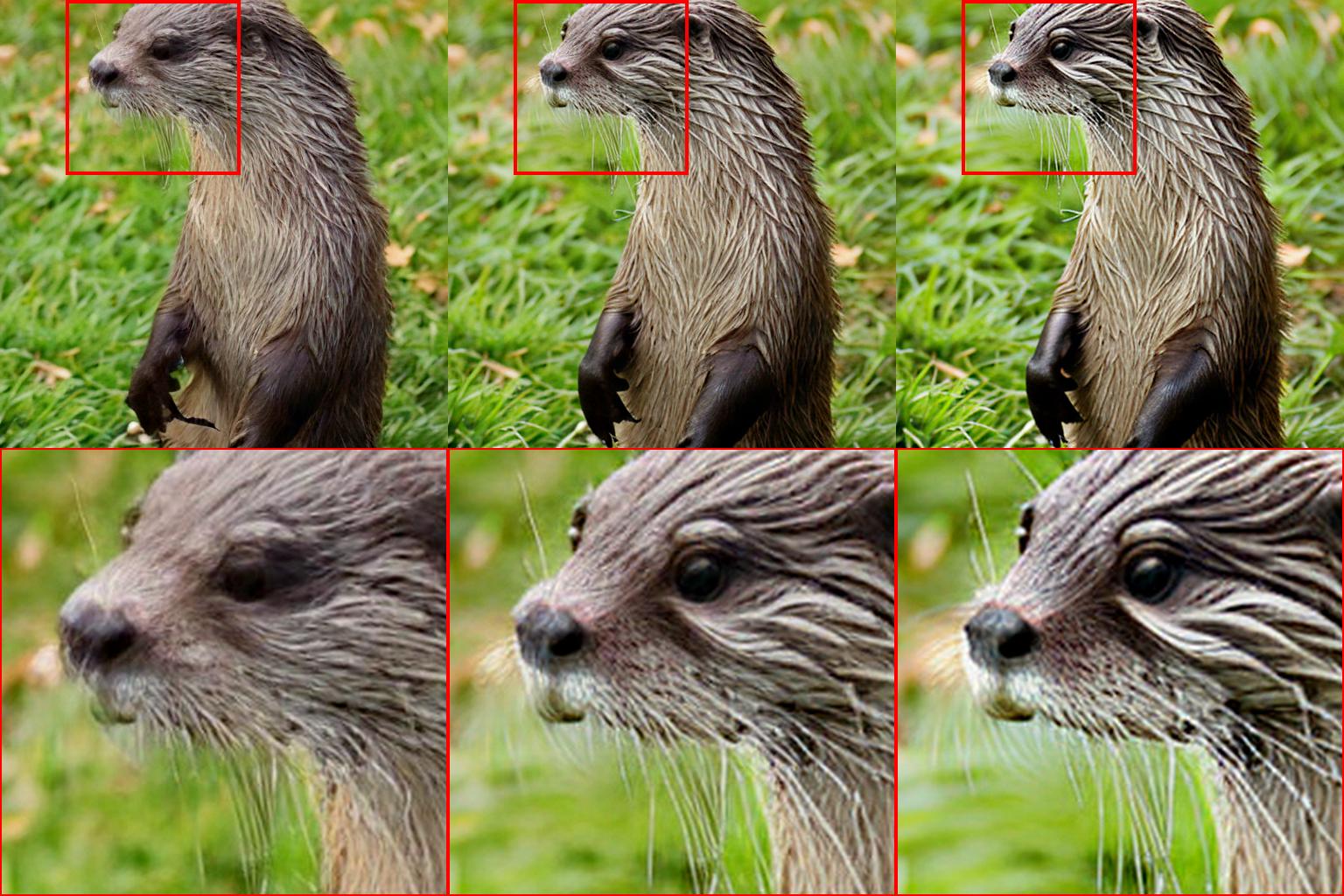}\\[-0.5mm]%
        \makebox[\hhh][c]{\small ImageNet class 324: \emph{cabbage butterfly}}\hfill%
        \makebox[\hhh][c]{\small ImageNet class 360: \emph{otter}}\\[1.0mm]%
        \includegraphics[width=\hhh]{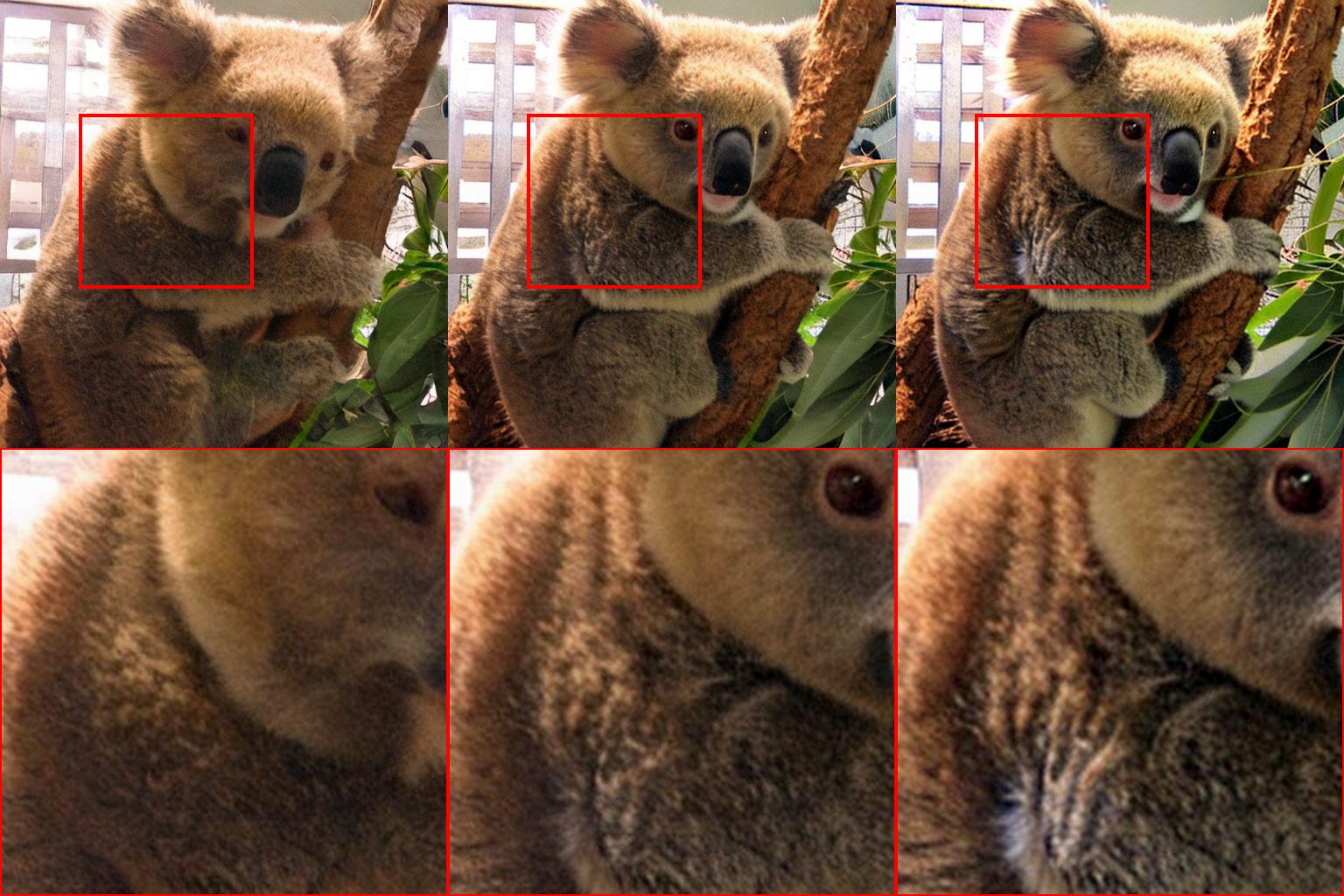}\hfill%
        \includegraphics[width=\hhh]{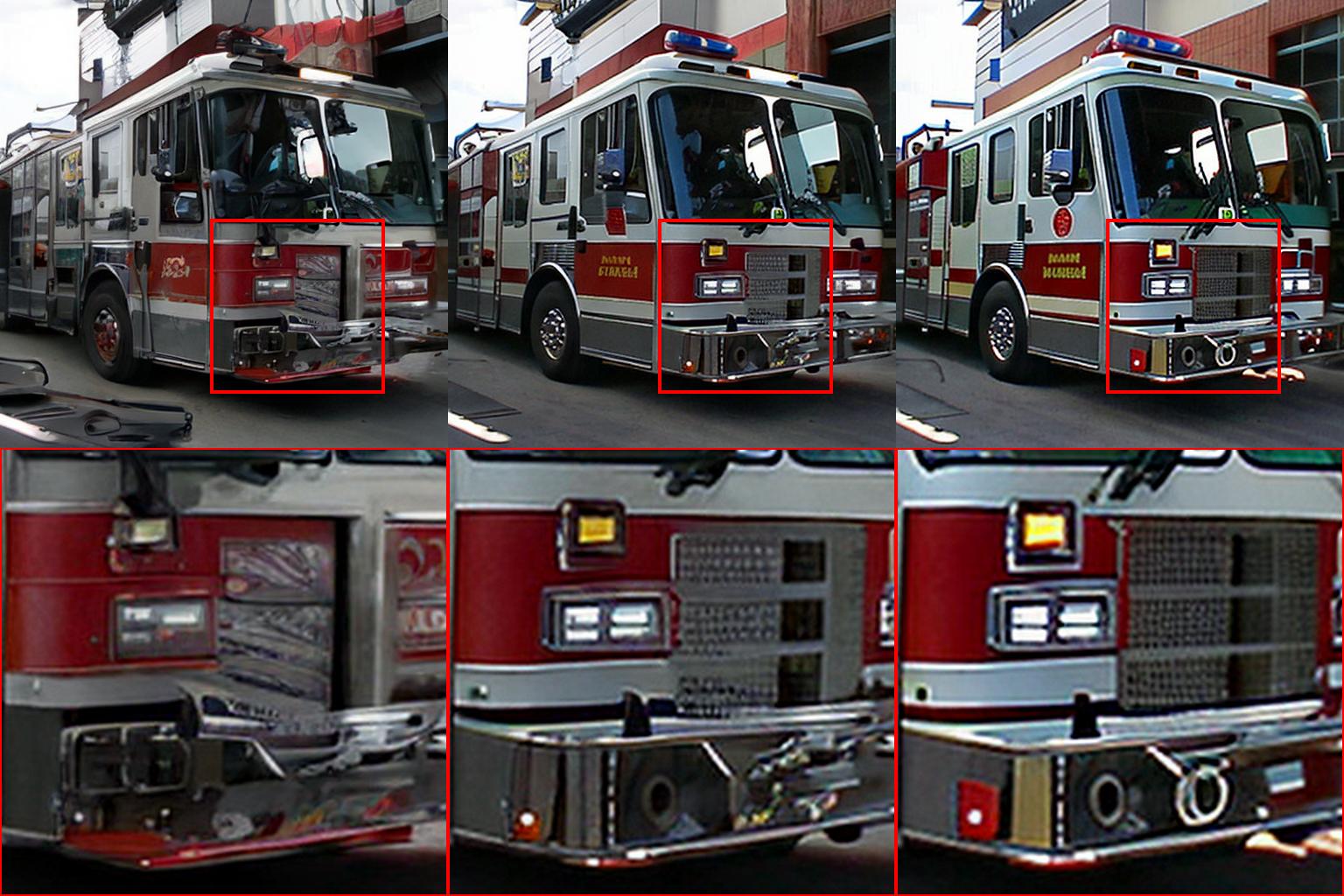}\\[-0.5mm]%
        \makebox[\hhh][c]{\small ImageNet class 105: \emph{koala}}\hfill%
        \makebox[\hhh][c]{\small ImageNet class 555: \emph{fire truck}}\\[1.0mm]%
        \includegraphics[width=\hhh]{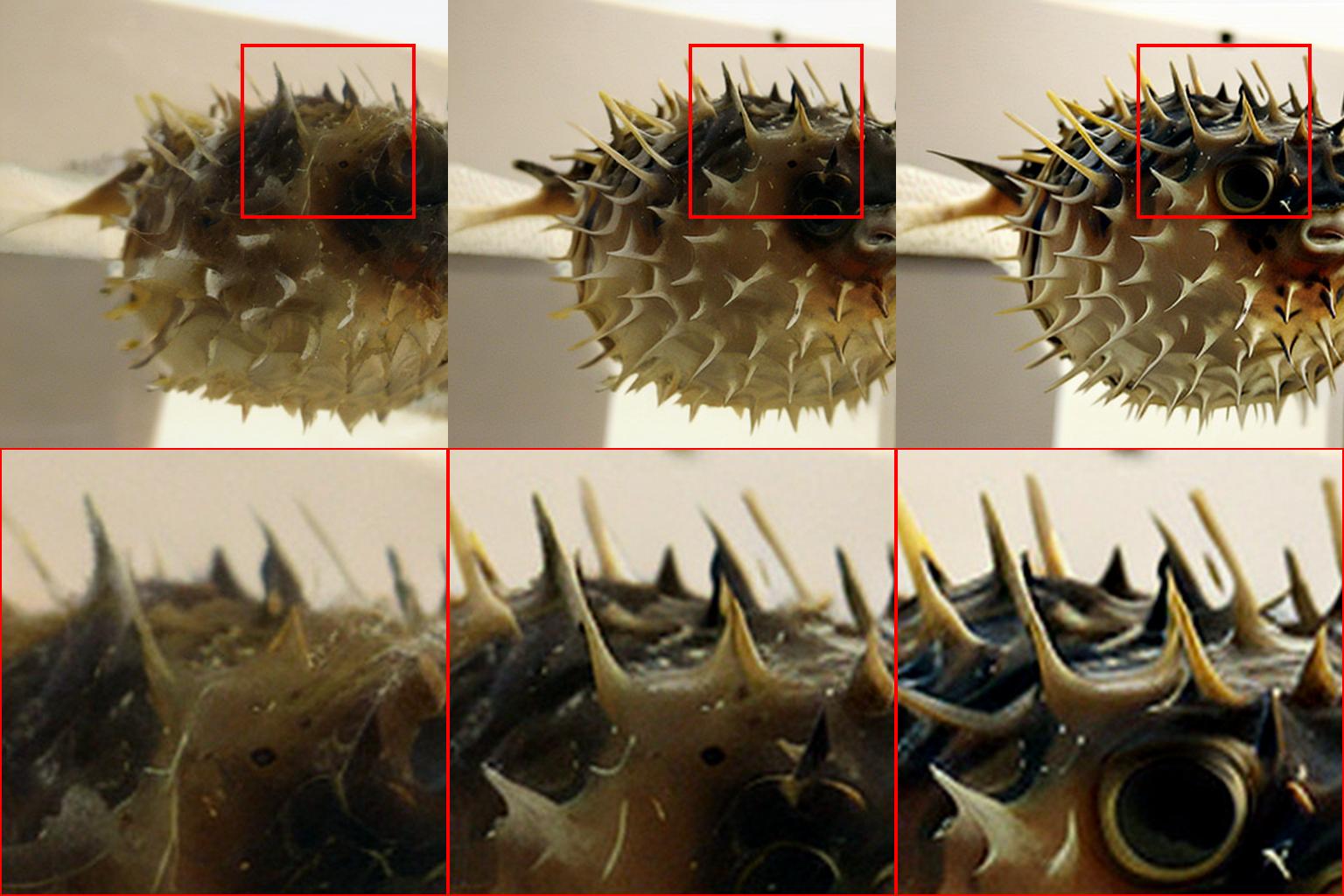}\hfill%
        \includegraphics[width=\hhh]{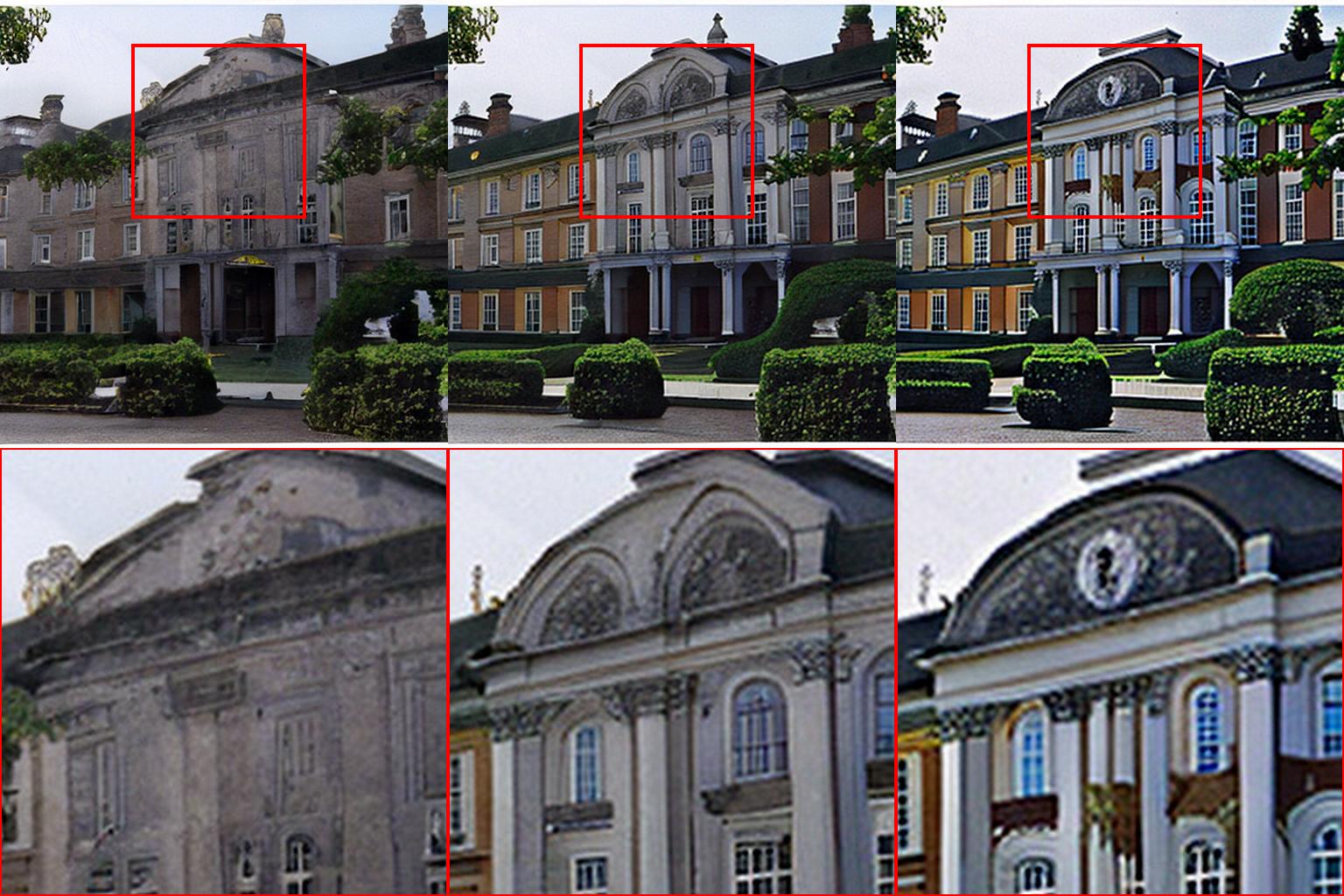}\\[-0.5mm]%
        \makebox[\hhh][c]{\small ImageNet class 397: \emph{pufferfish}}\hfill%
        \makebox[\hhh][c]{\small ImageNet class 698: \emph{palace}}\\[1.0mm]%
        \includegraphics[width=\hhh]{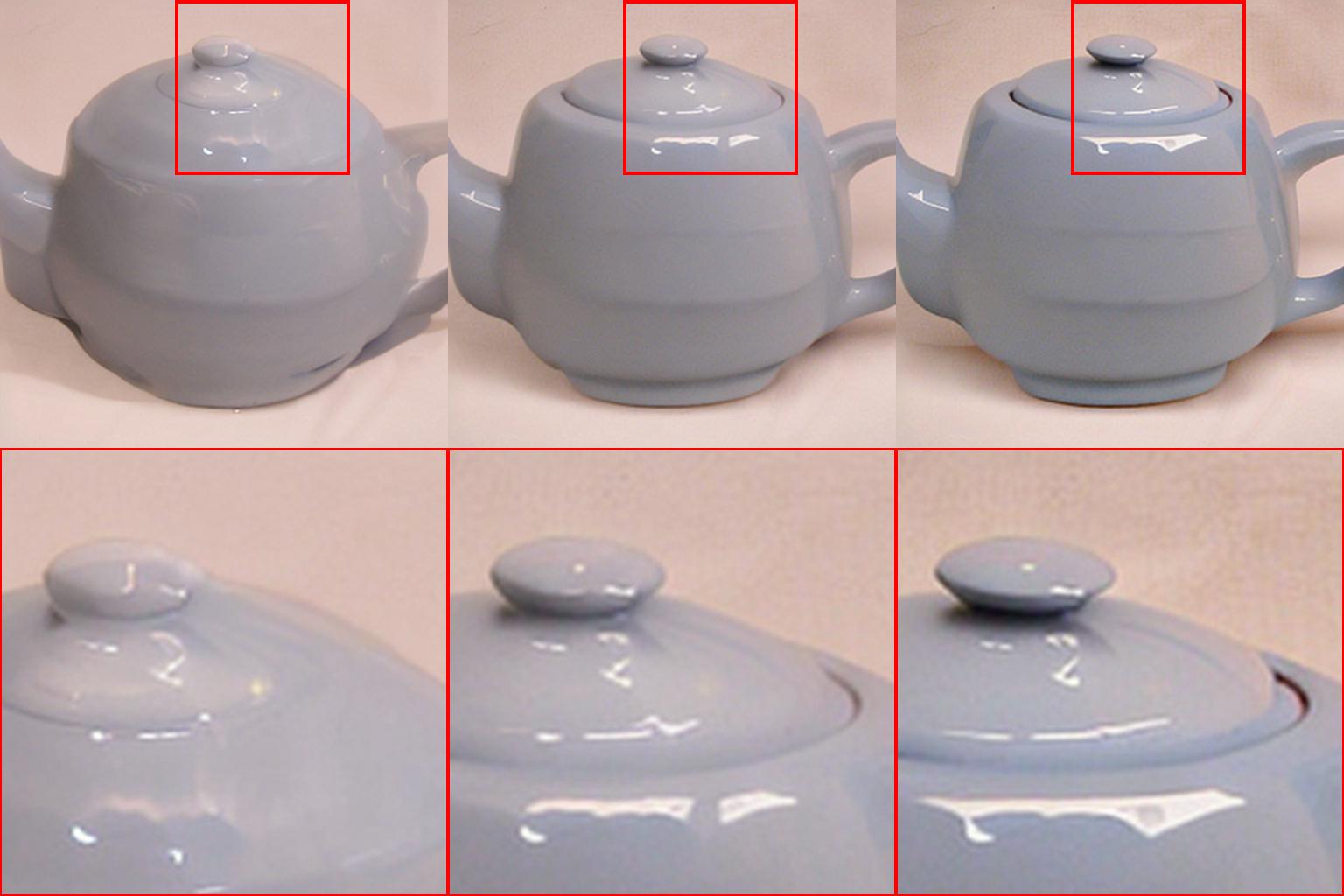}\hfill%
        \includegraphics[width=\hhh]{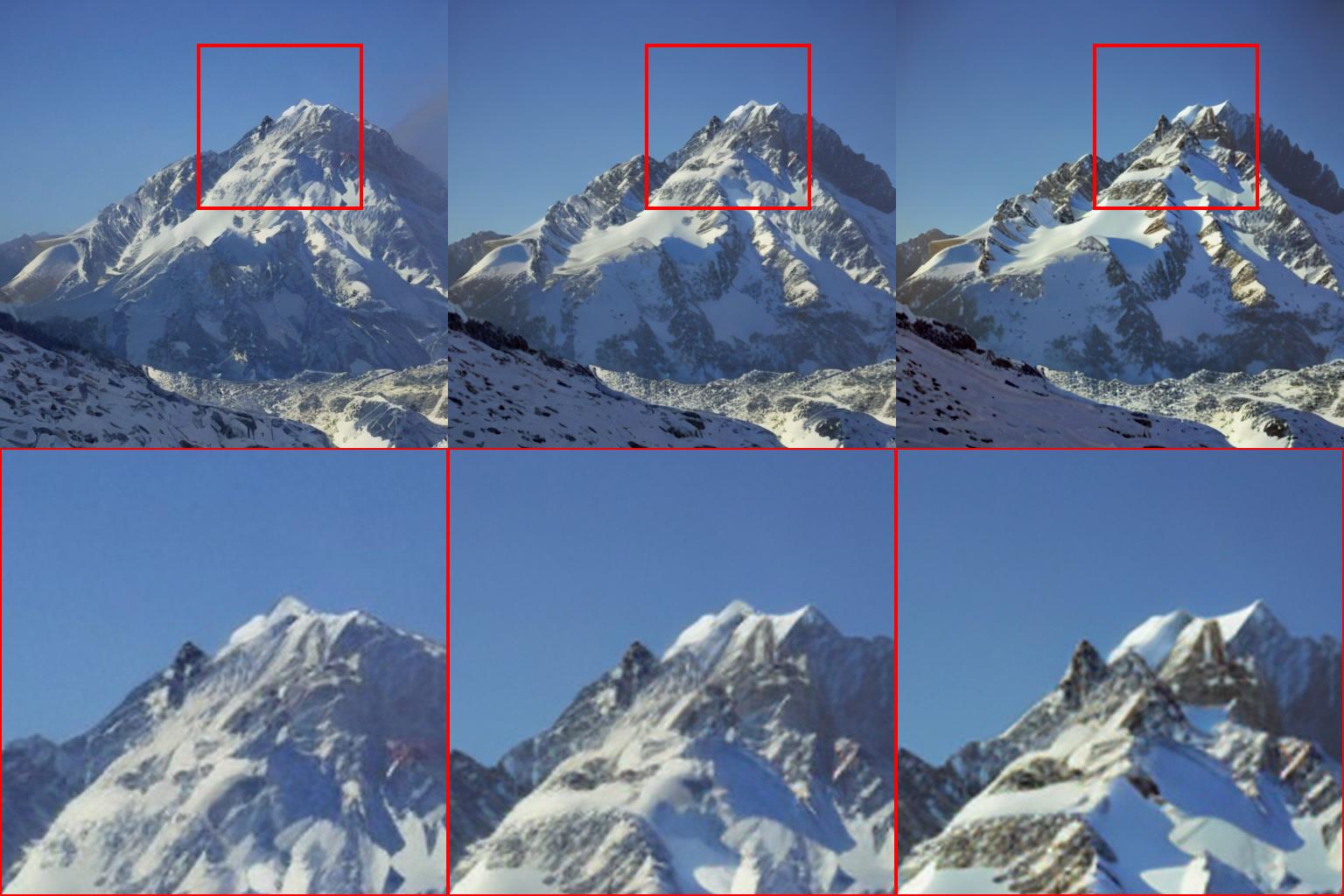}\\[-0.5mm]%
        \makebox[\hhh][c]{\small ImageNet class 849: \emph{teapot}}\hfill%
        \makebox[\hhh][c]{\small ImageNet class 970: \emph{alp}}%
        \caption{More EDM2-XXL results showing the effect of changing $\gweight$ with our method. We limit the guidance to $\sigma \in (0.19, 1.61]$. Increasing $\gweight$ produces images with more well-defined details while maintaining the color palette and the original image composition.}
        \label{fig:qualitative_results_B_imagenet_supp}
    \end{figure}
}

\newcommand{\figQualitativeResultsBSDXLSupp}{
    \renewcommand{\h}{0.2\linewidth}
    \renewcommand{\hh}{\linewidth}
    \renewcommand{\hhh}{0.495\linewidth}
    \renewcommand{\hhhh}{0.167\linewidth}
    \begin{figure}[t]%
        \makebox[\h][c]{$\gweight = 2$}%
        \makebox[\h][c]{$\gweight = 4$}%
        \makebox[\h][c]{$\gweight = 8$}%
        \makebox[\h][c]{$\gweight = 12$}%
        \makebox[\h][c]{$\gweight = 16$}\\%
        \includegraphics[width=\hh]{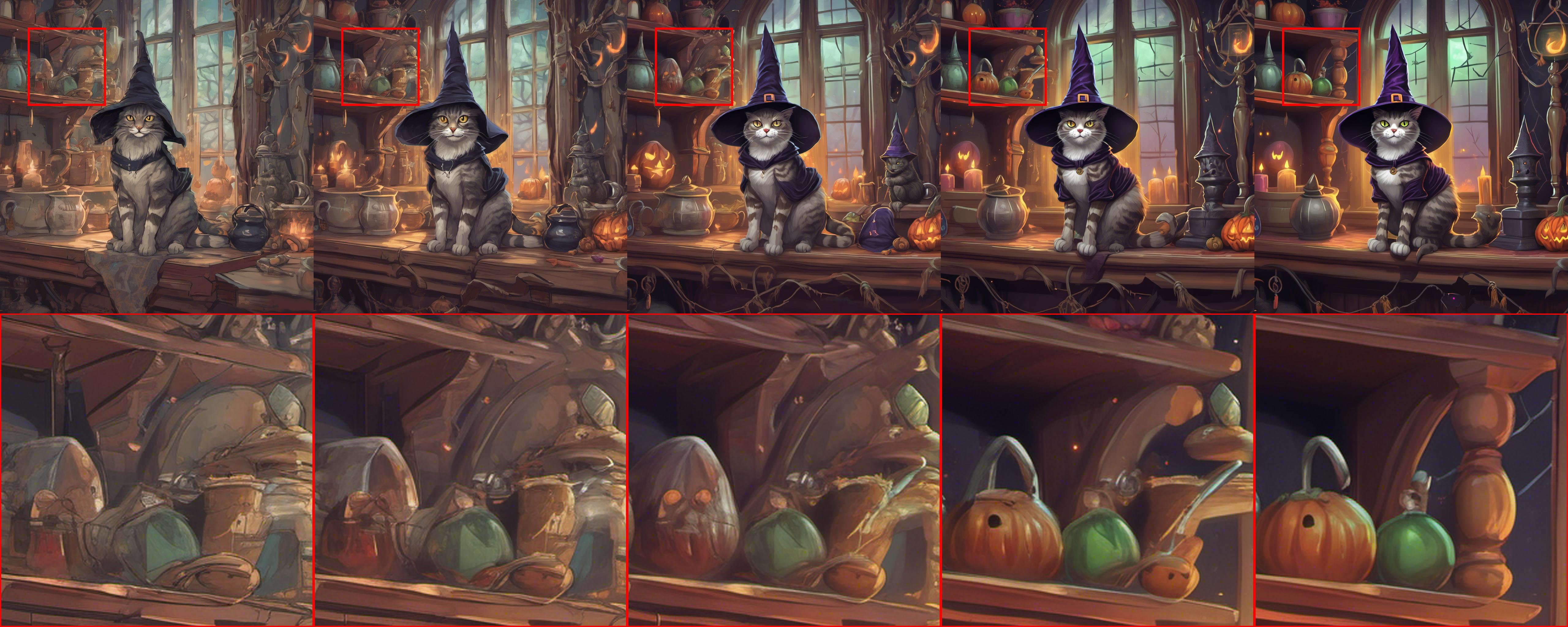}\\[0.5mm]%
        \makebox[\hh][c]{\begin{minipage}{0.9\linewidth}\small\emph{A highly detailed zoomed-in digital painting of a cat dressed as a witch wearing a wizard hat in a haunted house, artstation.}\end{minipage}}\\[1mm]%
        \includegraphics[width=\hh]{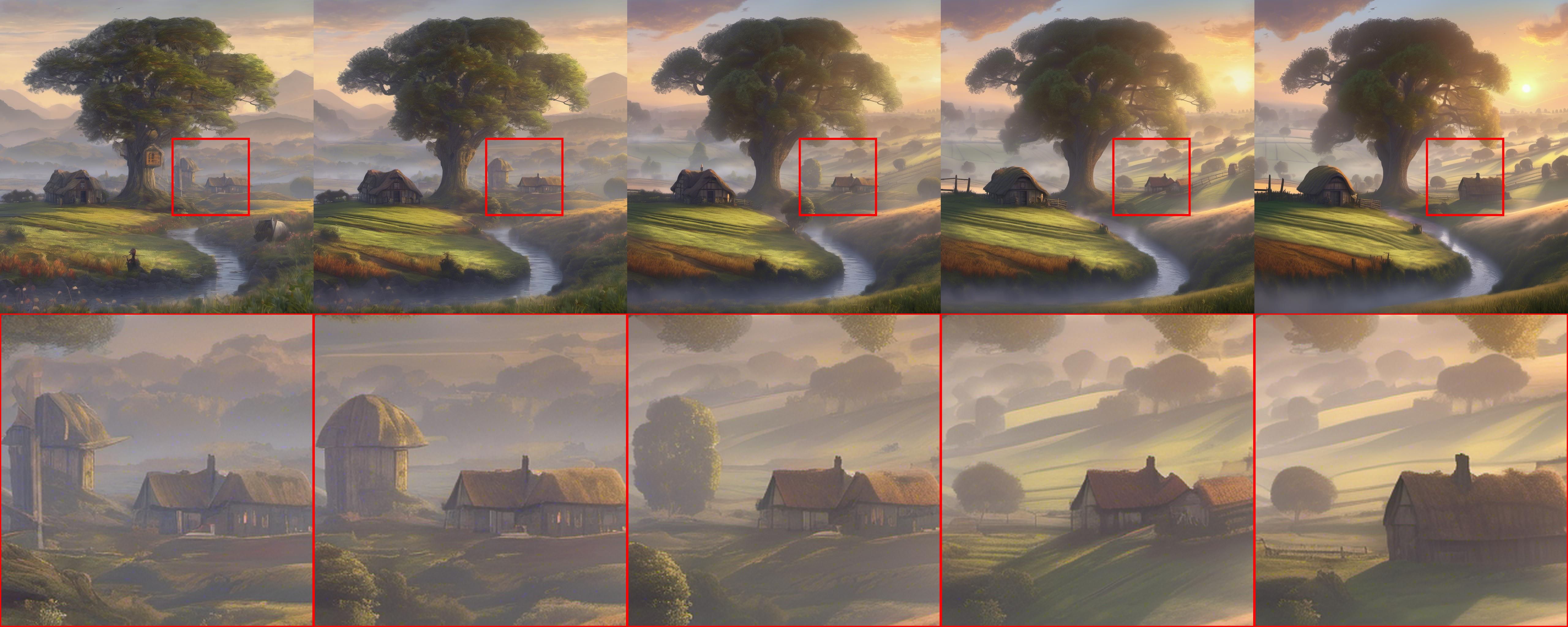}\\[0.5mm]%
        \makebox[\hh][c]{\begin{minipage}{0.9\linewidth}\small\emph{A fantasy landscape of the Shire during sunrise. The Sun is near the horizon and there is fog over farm fields. Highly detailed fantasy art, artstation.}\end{minipage}}\\[1mm]
        \includegraphics[width=\hh]{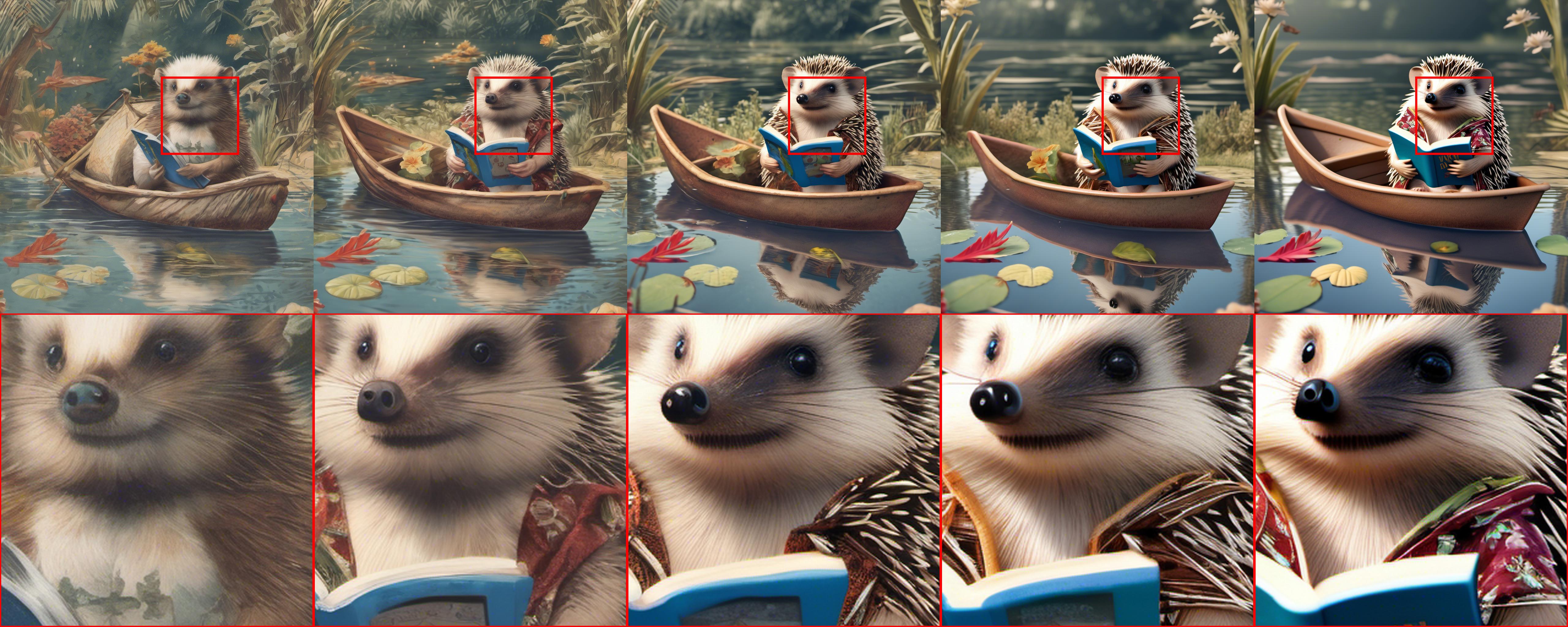}\\[0.5mm]%
        \makebox[\hh][c]{\begin{minipage}{0.9\linewidth}\small\emph{A 4K dslr photo of a hedgehog sitting in a small boat in the middle of a pond. It is wearing a Hawaiian shirt and a straw hat. It is reading a book. There are a few leaves in the background.}\end{minipage}}%
        \caption{More SD-XL results showing the effect of changing $\gweight$ with our method. We limit the guidance to $\sigma \in (0.28, 5.42]$. Increasing $\gweight$ produces images with more well-defined details while maintaining the color palette and the original image composition.}
        \label{fig:qualitative_results_B_sdxl_supp}
    \end{figure}
}

\newcommand{\figQualitativeResultsBImageNetSuppCFG}{
    \renewcommand{\h}{4mm}
    \renewcommand{\hh}{0.48\linewidth}
    \renewcommand{\hhh}{0.16\linewidth}
    \begin{figure}[t]%
        \makebox[0.5\linewidth][c]{CFG}\hfill%
        \makebox[0.5\linewidth][c]{Ours, $\sigma \in \left(0.19, 1.61\right]$}\\
        \makebox[\h][l]{\rotatebox[origin=l]{90}{\makebox[0mm][c]{}}}%
        \makebox[\hhh][c]{$\gweight = 1$}%
        \makebox[\hhh][c]{$\gweight = 3$}%
        \makebox[\hhh][c]{$\gweight = 5$}\hfill%
        \makebox[\hhh][c]{$\gweight = 1$}%
        \makebox[\hhh][c]{$\gweight = 3$}%
        \makebox[\hhh][c]{$\gweight = 5$}\\%
        \makebox[\h][l]{\rotatebox[origin=l]{90}{\makebox[0mm][c]{\hspace*{42.5mm}\small ImageNet class 33: \emph{loggerhead}}}}%
        \includegraphics[width=\hh]{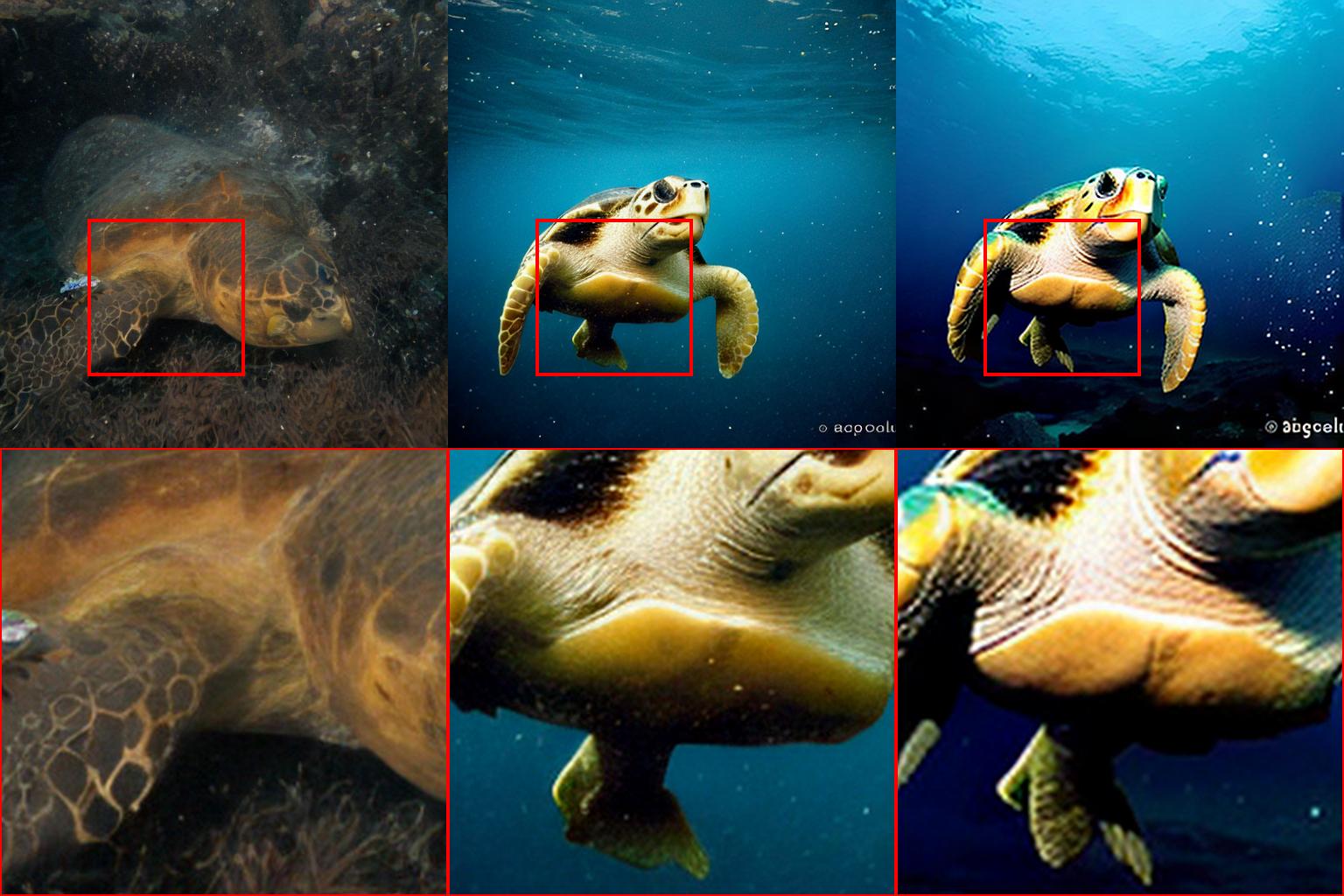}\hfill%
        \includegraphics[width=\hh]{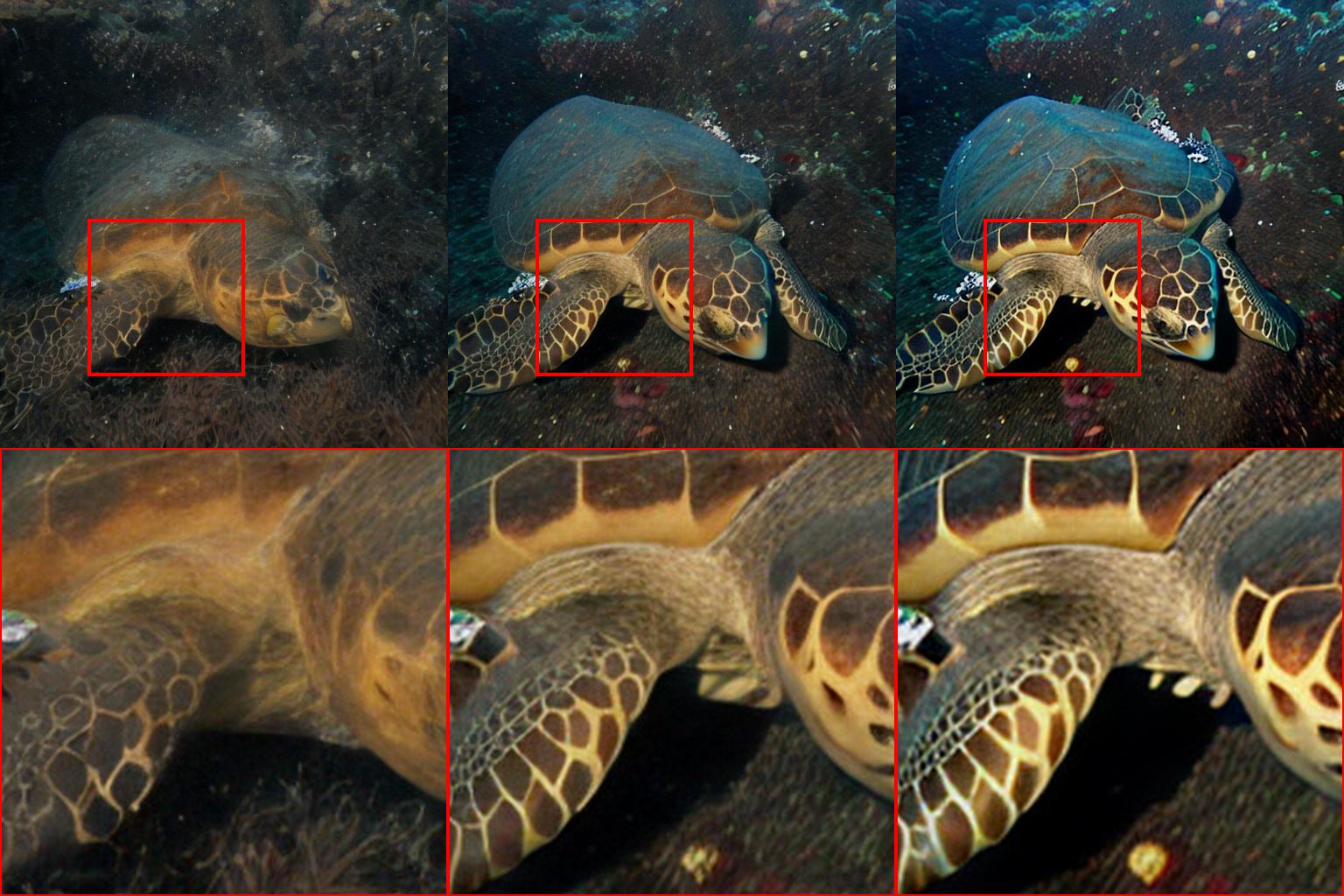}\\[1mm]%
        \makebox[\h][l]{\rotatebox[origin=l]{90}{\makebox[0mm][c]{\hspace*{42.5mm}\small ImageNet class 64: \emph{green mamba}}}}%
        \includegraphics[width=\hh]{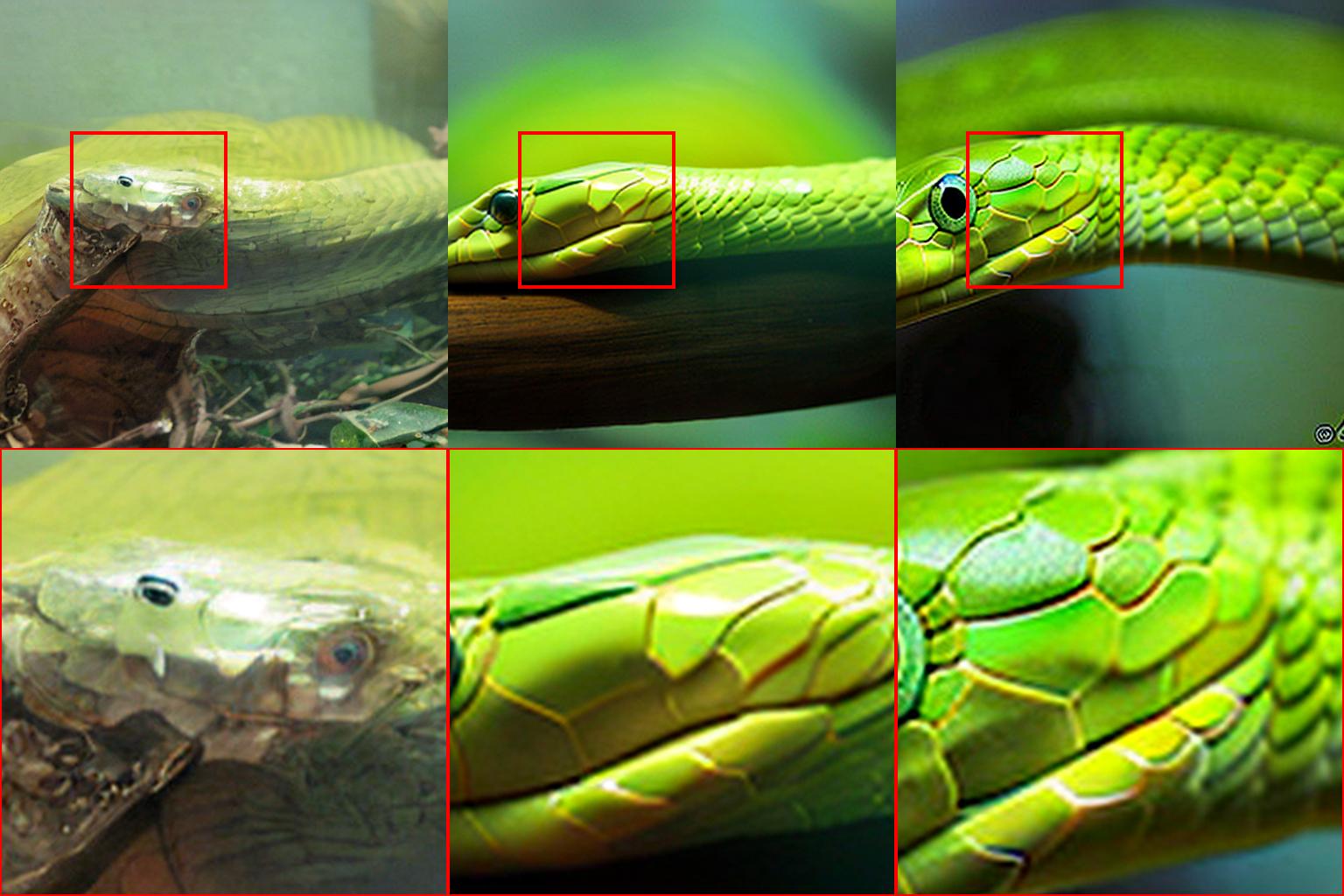}\hfill%
        \includegraphics[width=\hh]{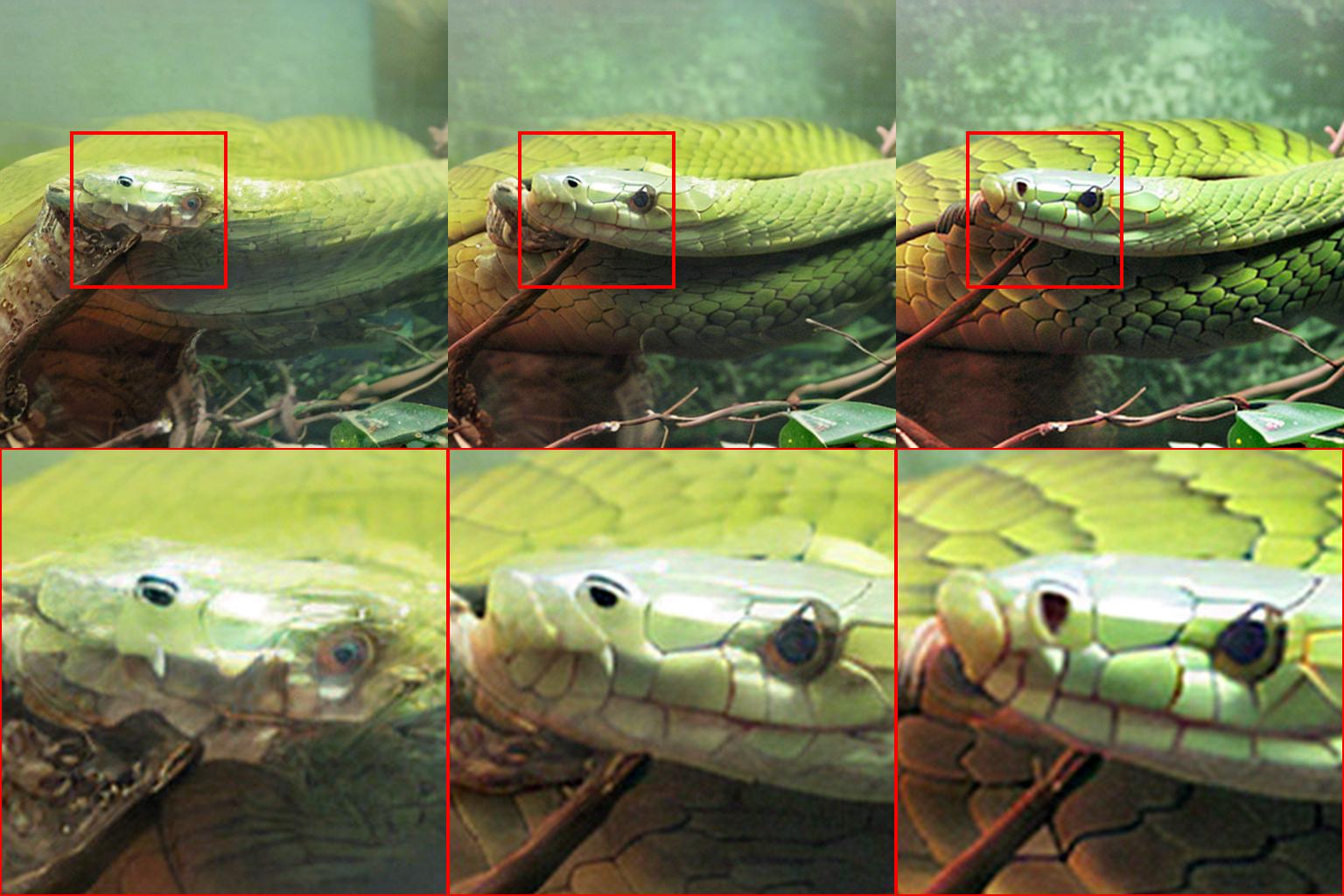}\\[1mm]%
        \makebox[\h][l]{\rotatebox[origin=l]{90}{\makebox[0mm][c]{\hspace*{42.5mm}\small ImageNet class 1: \emph{goldfish}}}}%
        \includegraphics[width=\hh]{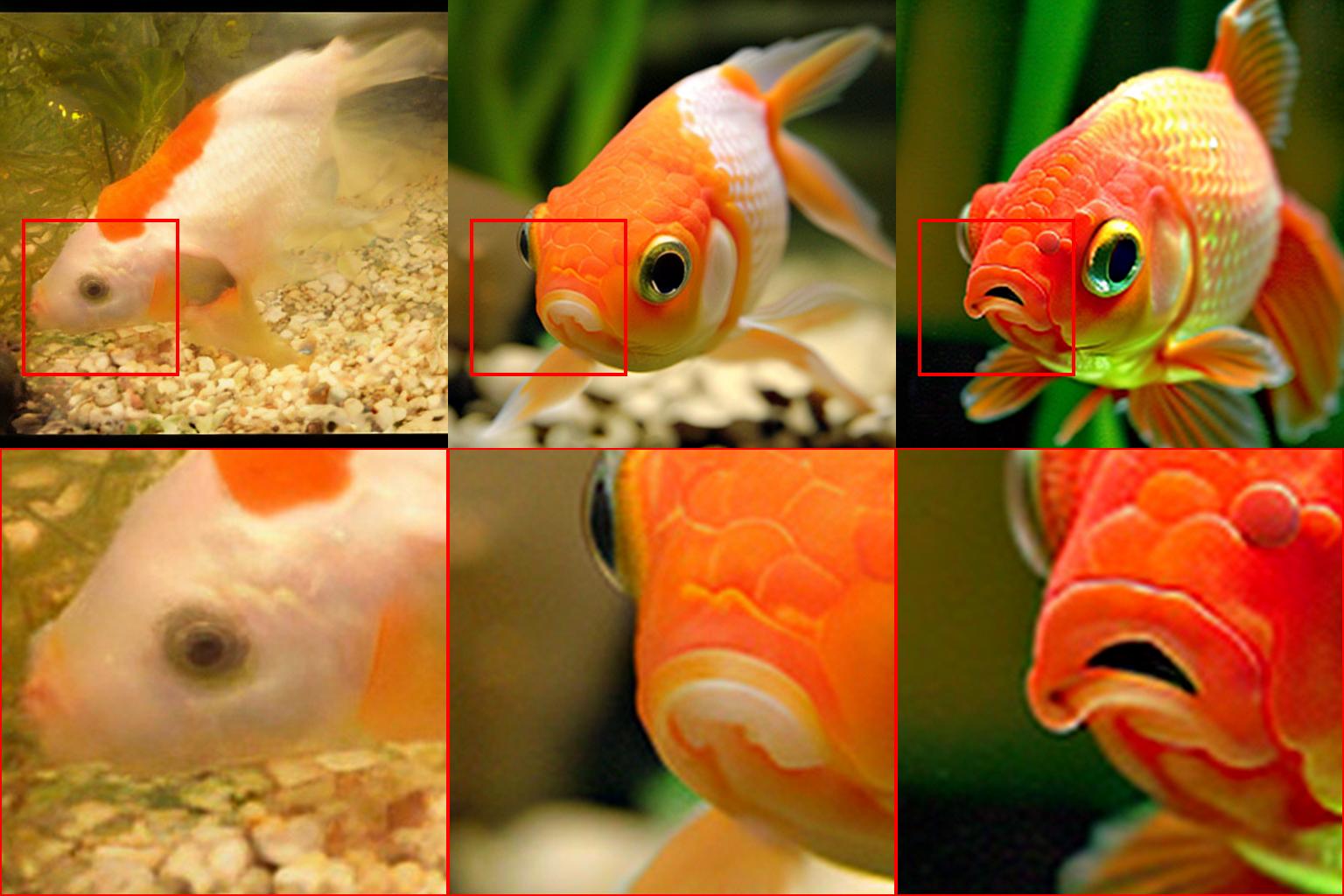}\hfill%
        \includegraphics[width=\hh]{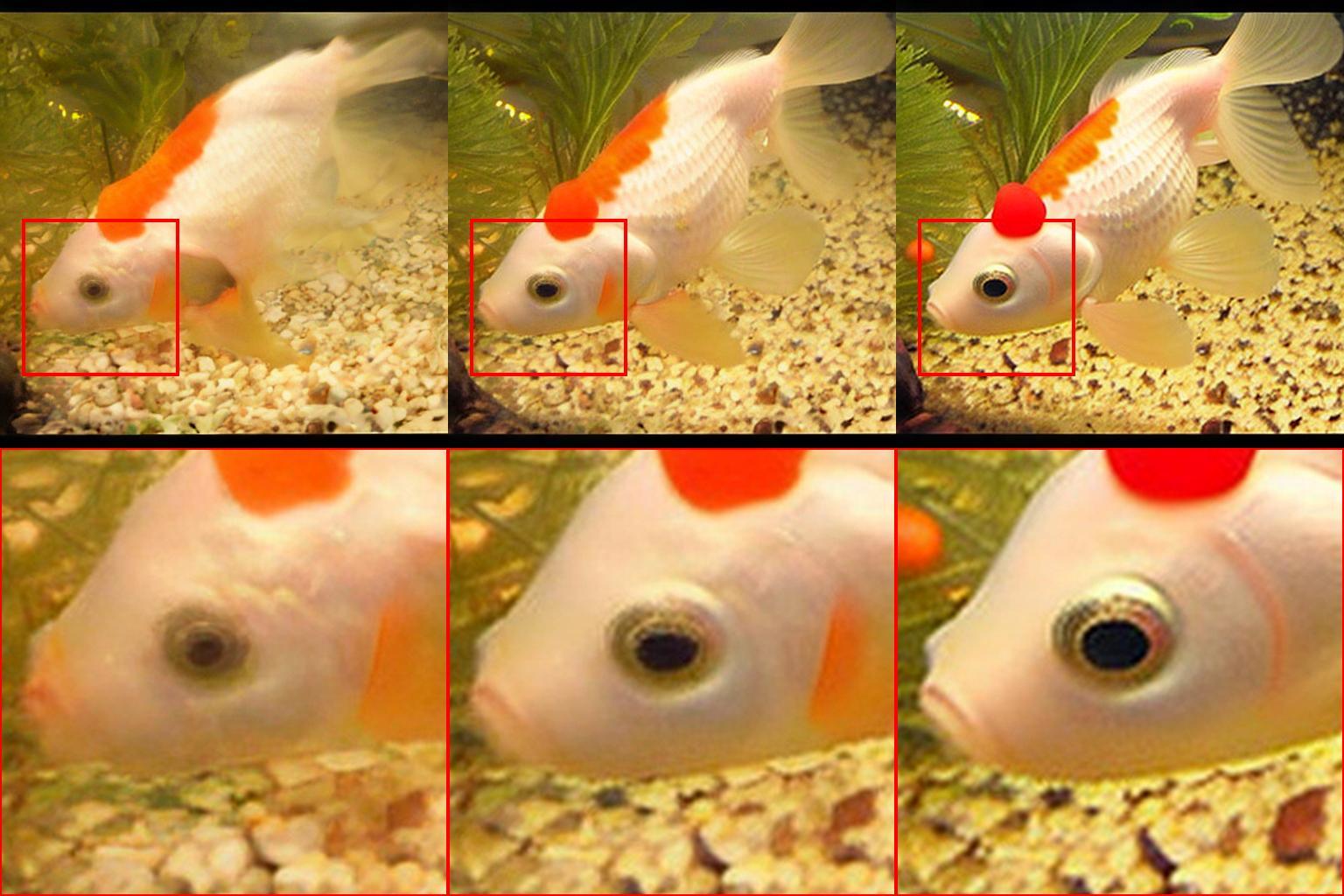}\\[1mm]%
        \makebox[\h][l]{\rotatebox[origin=l]{90}{\makebox[0mm][c]{\hspace*{42.5mm}\small ImageNet class 504: \emph{coffee mug}}}}%
        \includegraphics[width=\hh]{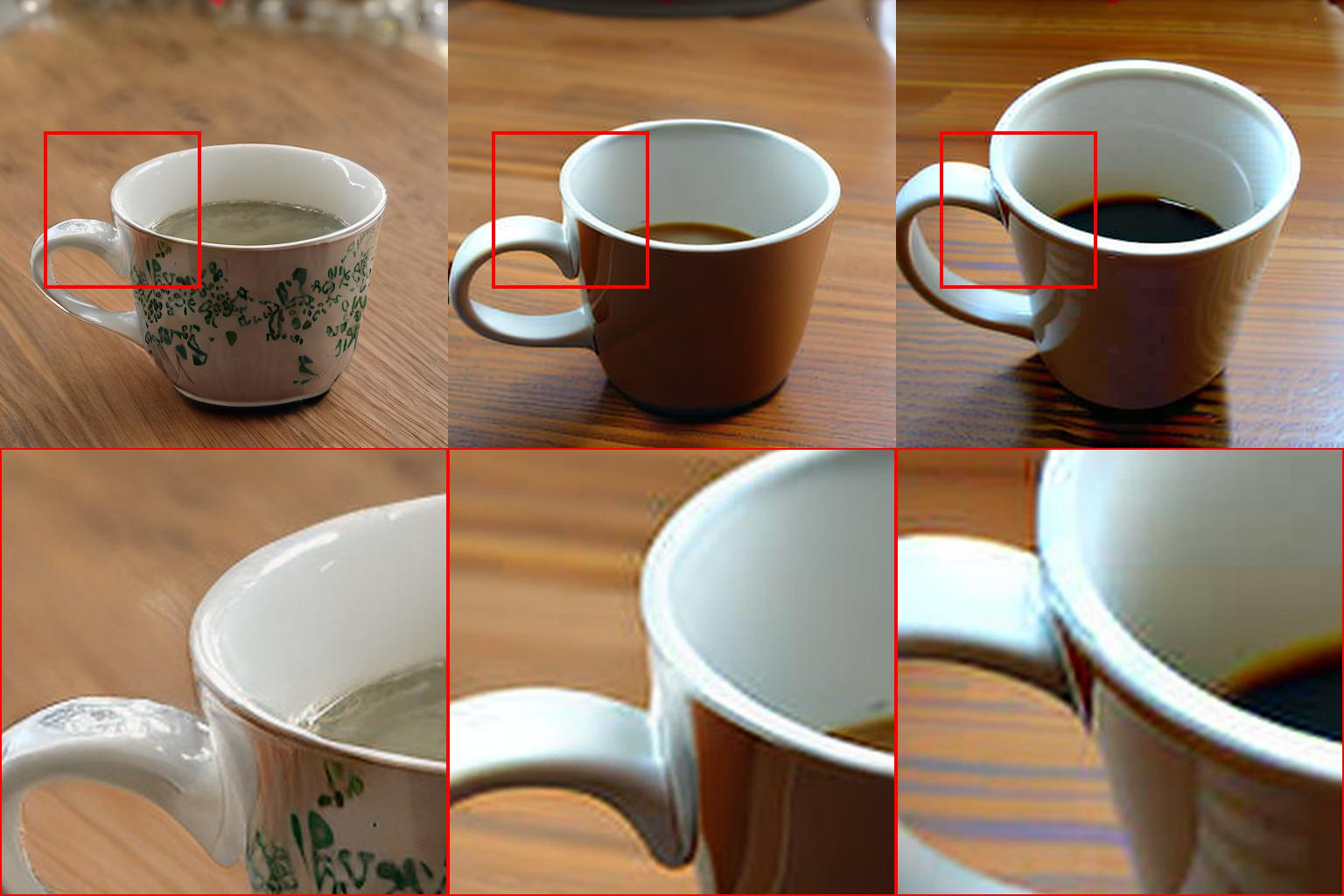}\hfill%
        \includegraphics[width=\hh]{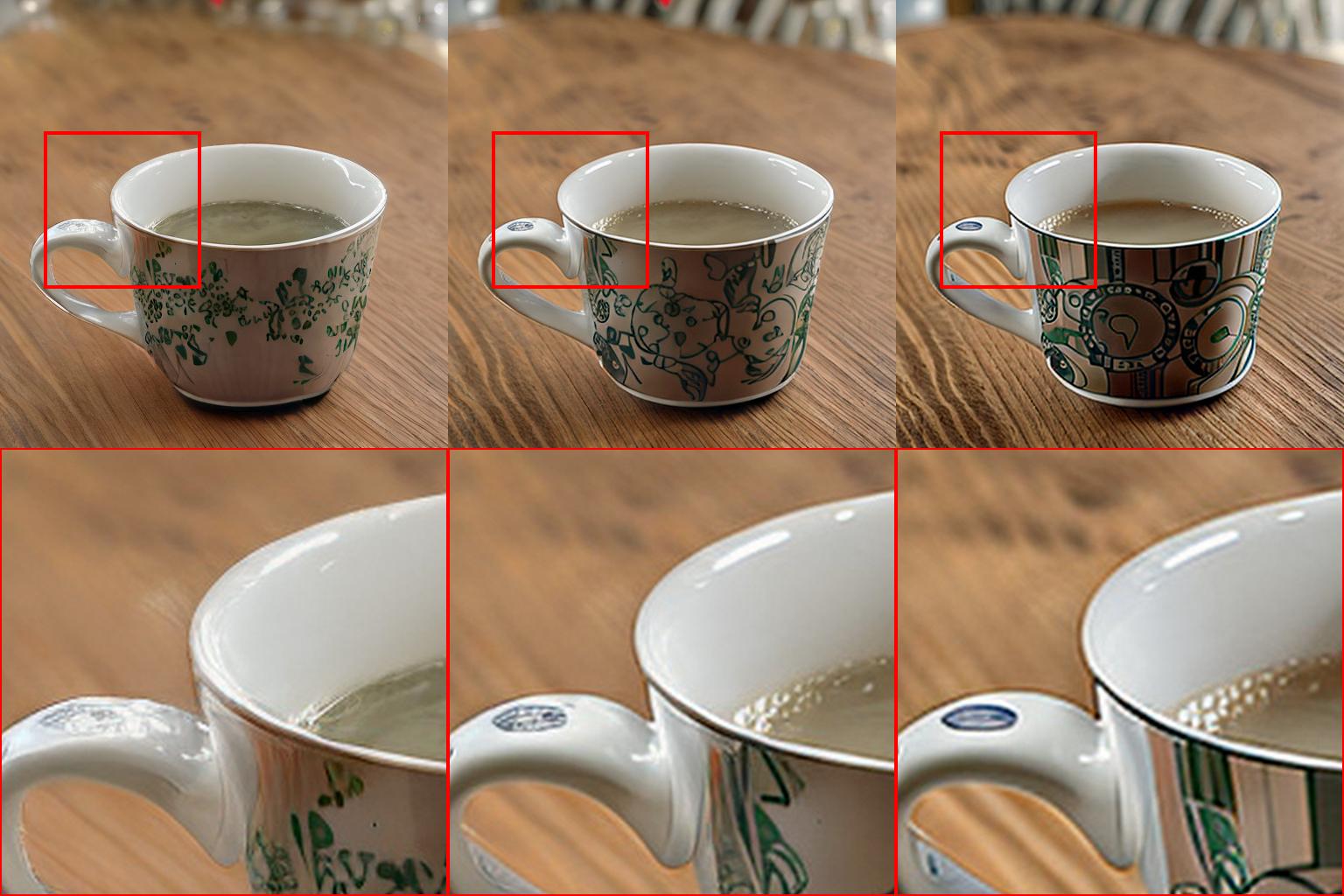}
        \caption{Effect of increasing guidance weight $\gweight$ with CFG vs. our method. \textbf{Left:} Increasing the guidance weight with CFG leads to large changes in the image composition. \textbf{Right:} Our method leads to well-defined image details and retains the overall composition to a significant degree.}
        \label{fig:qualitative_results_B_imagenet_cfg_supp}
    \end{figure}
}

\newcommand{\figQualitativeResultsBSDXLSuppCFGA}{
    \renewcommand{\h}{0.95\linewidth}
    \renewcommand{\hh}{4mm}
    \renewcommand{\hhh}{\linewidth}
    \renewcommand{\hhhh}{0.1583\linewidth}
    \begin{figure}[t]%
        \makebox[\hh][c]{}%
        \makebox[\hhhh][c]{$\gweight = 1$}%
        \makebox[\hhhh][c]{$\gweight = 2$}%
        \makebox[\hhhh][c]{$\gweight = 4$}%
        \makebox[\hhhh][c]{$\gweight = 8$}%
        \makebox[\hhhh][c]{$\gweight = 12$}%
        \makebox[\hhhh][c]{$\gweight = 16$}\\%
        \makebox[\hh][l]{\rotatebox[origin=l]{90}{\makebox[0mm][c]{\hspace*{40mm}\small CFG}}}%
        \includegraphics[width=\h]{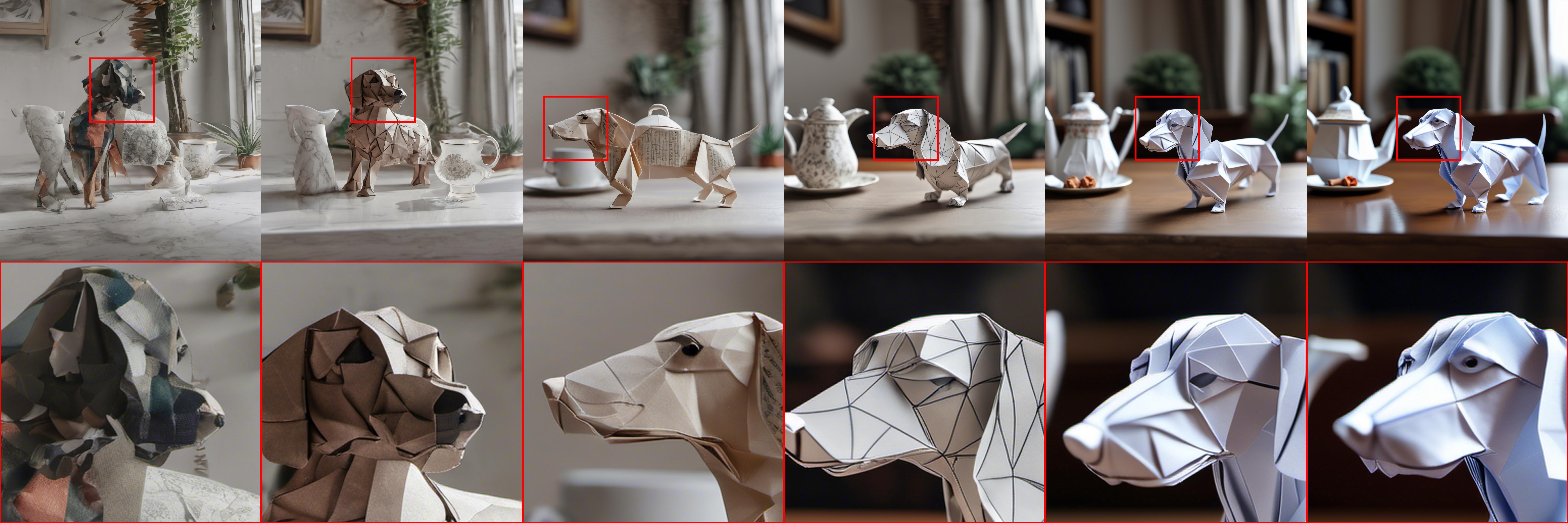}\\[0.5mm]
        \makebox[\hh][l]{\rotatebox[origin=l]{90}{\makebox[0mm][c]{\hspace*{40mm}\small Ours, $\sigma \in \left(0.28, 5.42\right]$}}}%
        \includegraphics[width=\h]{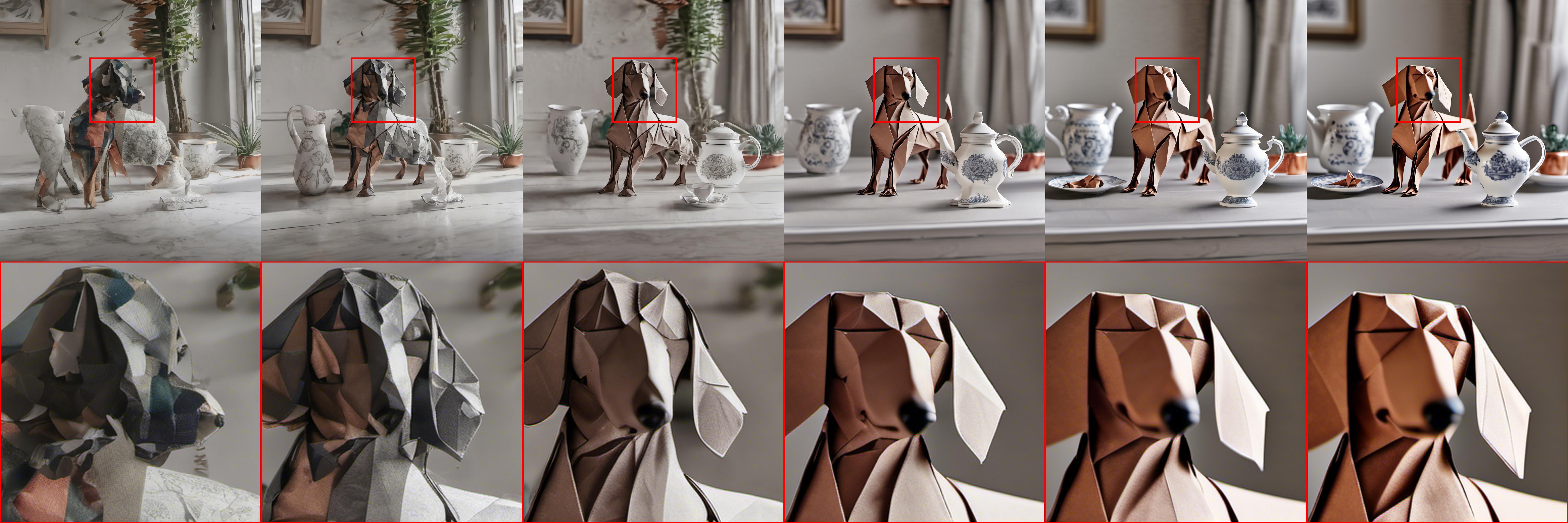}\\[0.5mm]%
        \makebox[\hh]{}\hfill%
        \makebox[\hhh][c]{\begin{minipage}{0.9\linewidth}\small\emph{A highly detailed paper origami of a Dachshund on a table next to a porcelain teapot, 4k dslr.}\end{minipage}}
        \caption{Effect of increasing guidance weight $\gweight$ with CFG vs. our method. \textbf{Top:} Increasing the guidance weight with CFG leads to large changes in the image composition. Note how the dog's head moves as $\gweight$ changes. \textbf{Bottom:} Our method leads to well-defined image details and retains the overall composition to a significant degree.}
        \label{fig:qualitative_results_B_sdxl_supp_cfga}
    \end{figure}
}

\newcommand{\figQualitativeResultsBSDXLSuppCFGB}{
    \renewcommand{\h}{0.95\linewidth}
    \renewcommand{\hh}{4mm}
    \renewcommand{\hhh}{\linewidth}
    \renewcommand{\hhhh}{0.1583\linewidth}
    \begin{figure}[t]%
        \makebox[\hh][c]{}%
        \makebox[\hhhh][c]{$\gweight = 1$}%
        \makebox[\hhhh][c]{$\gweight = 2$}%
        \makebox[\hhhh][c]{$\gweight = 4$}%
        \makebox[\hhhh][c]{$\gweight = 8$}%
        \makebox[\hhhh][c]{$\gweight = 12$}%
        \makebox[\hhhh][c]{$\gweight = 16$}\\%
        \makebox[\hh][l]{\rotatebox[origin=l]{90}{\makebox[0mm][c]{\hspace*{40mm}\small CFG}}}%
        \includegraphics[width=\h]{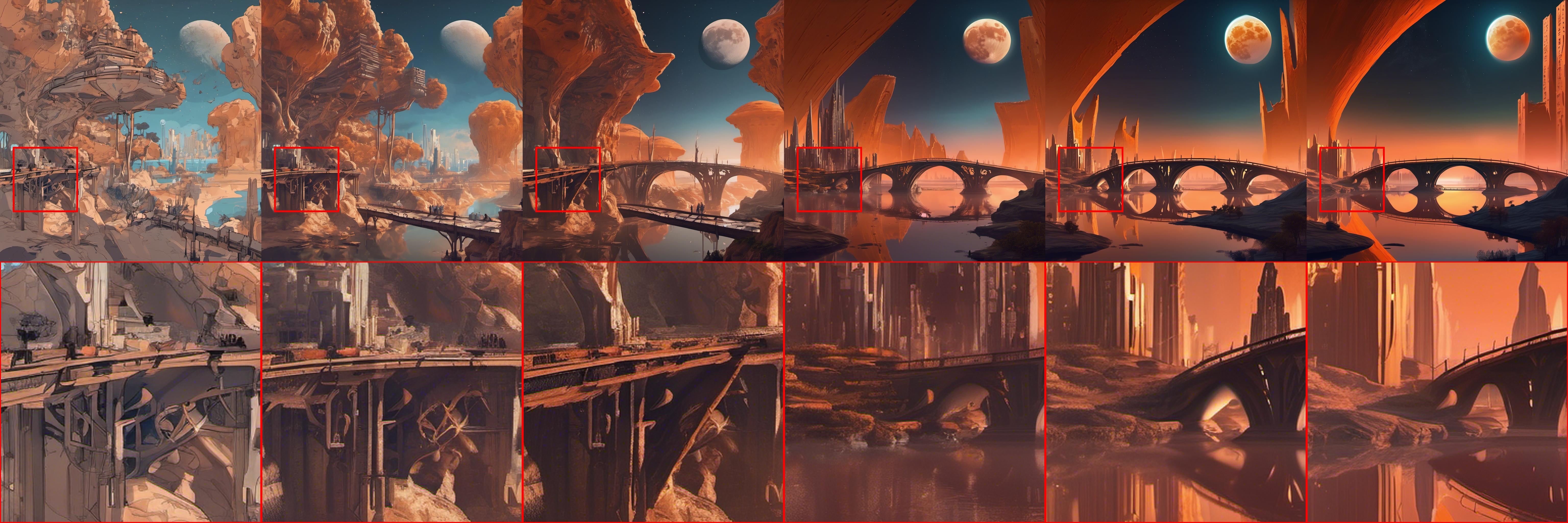}\\[0.5mm]%
        \makebox[\hh][l]{\rotatebox[origin=l]{90}{\makebox[0mm][c]{\hspace*{40mm}\small Ours, $\sigma \in \left(0.28, 5.42\right]$}}}%
        \includegraphics[width=\h]{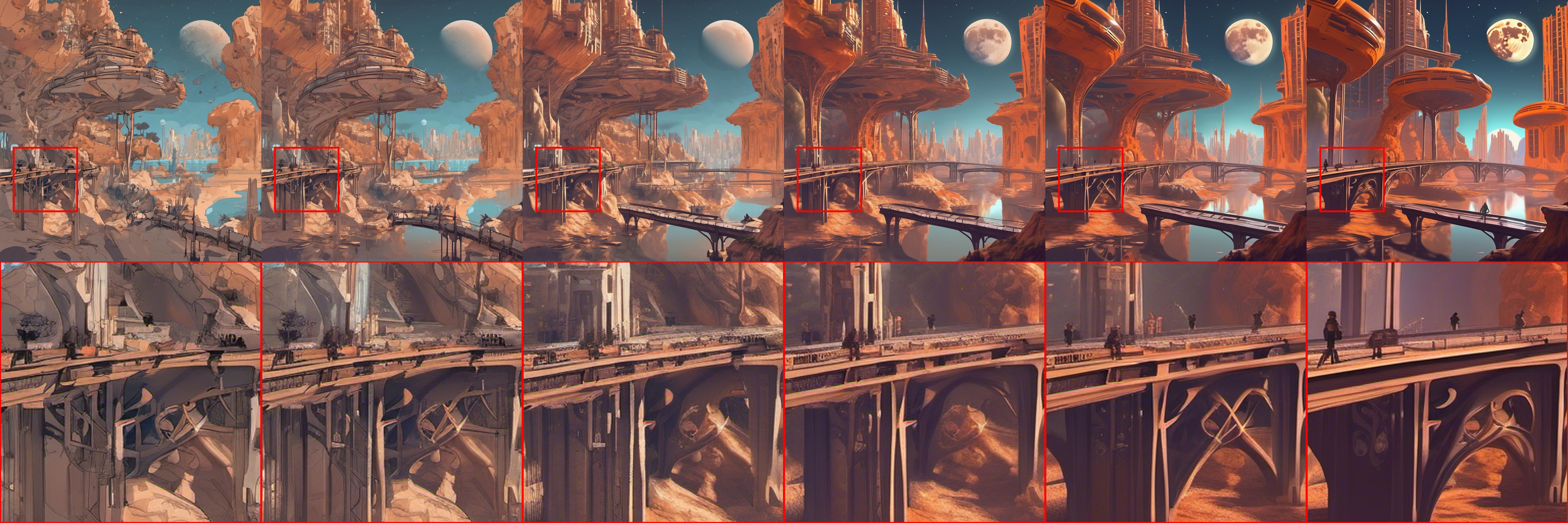}\\[0.5mm]%
        \makebox[\hhh][c]{\begin{minipage}{0.9\linewidth}\small\emph{A fantasy landscape on an alien planet in which there are many buildings. There is a beautiful bridge with a pond in the center. There is one large moon in the sky. The sky is orange. Digital art, artstation}\end{minipage}}
        \caption{Effect of increasing guidance weight $\gweight$ with CFG vs. our method. \textbf{Top:} Increasing the guidance weight with CFG leads to large changes in the image composition. \textbf{Bottom:} Our method leads to well-defined image details and retains the overall composition to a significant degree.}
        \label{fig:qualitative_results_B_sdxl_supp_cfgb}
    \end{figure}
}

\newcommand{\trajxmax}{0.9}
\newcommand{\trajymin}{-1.5}
\newcommand{\trajymax}{2.2}
\newcommand{\trajh}{42mm}
\newcommand{\trajhs}{31mm}

\newcommand{\plotTrajA}{%
\centering\footnotesize%
\begin{tikzpicture}%
\begin{axis}[
  width={1.34\columnwidth}, height={\trajh}, grid={none}, axis on top,
  xmin={0}, xmax={\trajxmax}, xmode={linear}, xtick={0, .45, \trajxmax}, xticklabels={\rlap{0}, \raisebox{-1.3ex}{$\sigma$}, \llap{\trajxmax}},
  ymin={\trajymin}, ymax={\trajymax}, ymode={linear}, ytick={}, yticklabels={},
]
\addplot graphics [xmin=0,xmax=\trajxmax,ymin=\trajymin,ymax=\trajymax] {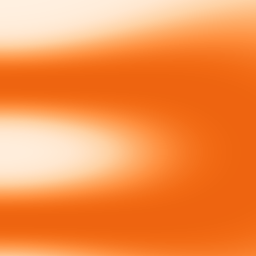};
\foreach \col in {1,...,16}{\addplot[black, thin, forget plot, opacity=0.5] table[x index=0, y index=\col] {figures/trajectory/traj_full.txt};}
\end{axis}
\begin{scope}[shift={(.35\columnwidth, .03\columnwidth)}]\begin{axis}[
  width={0.95\columnwidth}, height={\trajhs}, grid={none}, axis on top,
  xmin={0}, xmax={8}, xmode={linear}, xtick={}, xticklabels={},
  ymin={-15}, ymax={15}, ymode={linear}, ytick={}, yticklabels={},
]
\addplot graphics [xmin=0,xmax=8,ymin=-15,ymax=15] {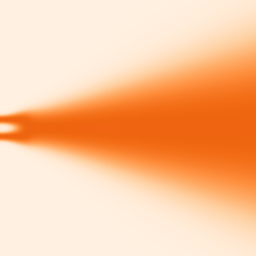};
\foreach \col in {1,...,16}{\addplot[black, thin, forget plot, opacity=0.5] table[x index=0, y index=\col] {figures/trajectory/traj_far.txt};}
\draw[cyan, semithick] (axis cs:0.0,-1.8) rectangle (axis cs:1.1,2.4);
\end{axis}\end{scope}
\end{tikzpicture}%
}%

\newcommand{\plotTrajB}{%
\centering\footnotesize%
\begin{tikzpicture}%
\begin{axis}[
  width={1.34\columnwidth}, height={\trajh}, grid={none}, axis on top,
  xmin={0}, xmax={\trajxmax}, xmode={linear}, xtick={0, .45, .617, \trajxmax}, xticklabels={\rlap{0}, \raisebox{-1.3ex}{$\sigma$}, 0.6, \llap{\trajxmax}},
  ymin={\trajymin}, ymax={\trajymax}, ymode={linear}, ytick={}, yticklabels={},
]
\addplot graphics [xmin=0,xmax=\trajxmax,ymin=\trajymin,ymax=\trajymax] {figures/trajectory/uncond_near.png};
\fill[white, opacity=0.3] (axis cs:0.617,\trajymin) rectangle (axis cs:\trajxmax,\trajymax);
\foreach \col in {1,...,16}{\addplot[black, thin, forget plot, opacity=0.5] table[x index=0, y index=\col] {figures/trajectory/traj_low.txt};}
\draw[densely dotted, black] (axis cs:0.617,\trajymin) -- (axis cs:0.617,\trajymax);
\end{axis}
\end{tikzpicture}%
}%

\newcommand{\plotTrajC}{%
\centering\footnotesize%
\begin{tikzpicture}%
\begin{axis}[
  width={1.34\columnwidth}, height={\trajh}, grid={none}, axis on top,
  xmin={0}, xmax={\trajxmax}, xmode={linear}, xtick={0, .1, .45, .617, \trajxmax}, xticklabels={\rlap{0}, 0.1, \raisebox{-1.3ex}{$\sigma$}, 0.6, \llap{\trajxmax}},
  ymin={\trajymin}, ymax={\trajymax}, ymode={linear}, ytick={}, yticklabels={},
]
\addplot graphics [xmin=0,xmax=\trajxmax,ymin=\trajymin,ymax=\trajymax] {figures/trajectory/uncond_near.png};
\fill[white, opacity=0.3] (axis cs:0,\trajymin)   rectangle (axis cs:0.1,\trajymax);
\fill[white, opacity=0.3] (axis cs:0.617,\trajymin) rectangle (axis cs:\trajxmax,\trajymax);
\foreach \col in {1,...,16}{\addplot[black, thin, forget plot, opacity=0.5] table[x index=0, y index=\col] {figures/trajectory/traj_mid.txt};}
\draw[densely dotted, black] (axis cs:0.100,\trajymin) -- (axis cs:0.100,\trajymax);
\draw[densely dotted, black] (axis cs:0.617,\trajymin) -- (axis cs:0.617,\trajymax);
\end{axis}
\end{tikzpicture}%
}%

\newcommand{\plotTrajP}{%
\centering\footnotesize%
\begin{tikzpicture}%
\begin{axis}[
  width={1.565\columnwidth}, height={\trajh}, grid={none}, axis on top,
  xmin=-.1, xmax=1, xmode={linear}, xtick={0}, xticklabels={\vphantom{0}},
  ymin={\trajymin}, ymax={\trajymax}, ymode={linear},
  ytick={-1.5, .35, 1.95}, yticklabels={\smash{\textminus 1.5}, $\xx\ $, \smash{\trajymax}},
]
\addplot+[name path=A, draw=none, mark=none, forget plot] coordinates {(1.1,-1.6) (1.1,2.3)};
\addplot+[name path=B, draw=none, mark=none, forget plot] table[x index=1, y index=0] {figures/trajectory/traj_pdfs.txt};
\addplot[C1, opacity=0.3, forget plot] fill between[of=A and B];
\addplot[C1, forget plot] table[y index=0, x index=1] {figures/trajectory/traj_pdfs.txt};
\addplot+[name path=A, draw=none, mark=none, forget plot] coordinates {(1.1,-1.6) (1.1,2.3)};
\addplot+[name path=B, draw=none, mark=none, forget plot] table[x index=2, y index=0] {figures/trajectory/traj_pdfs.txt};
\addplot[C2, opacity=0.3, forget plot] fill between[of=A and B];
\addplot[C2, forget plot] table[y index=0, x index=2] {figures/trajectory/traj_pdfs.txt};
\end{axis}
\end{tikzpicture}%
}%

\newcommand{\figTrajectory}{
\begin{figure*}[t]
\hspace*{-1mm}%
\begin{subfigure}{.121\linewidth}\plotTrajP\vspace{-0.6mm}\caption{\small \hspace*{2mm}(a) PDFs\hspace*{-2mm}}\end{subfigure}%
\begin{subfigure}{.294\linewidth}\plotTrajA\vspace{-0.6mm}\caption{\small \hspace*{2mm}(b) Guidance everywhere}\end{subfigure}%
\begin{subfigure}{.294\linewidth}\plotTrajB\vspace{-0.6mm}\caption{\small \hspace*{2mm}(c) Guidance at $\sigma \in (0, 0.6]$}\end{subfigure}%
\begin{subfigure}{.294\linewidth}\plotTrajC\vspace{-0.6mm}\caption{\small \hspace*{2mm}(d) Guidance at $\sigma \in (0.1, 0.6]$}\end{subfigure}%
\caption{\label{fig:trajectory}%
Illustration of the detrimental effects of guidance at high $\sigma$ in a synthetic 1D scenario.
\textbf{(a)} PDFs of the unconditional (orange) and conditional (green) data distributions used in this example.
\textbf{(b)} Activating guidance (weight \mbox{$\gweight = 6$}) everywhere leads to a catastrophic mode drop.
The zoomed-out inset shows how guidance pushes the sampling trajectories outside the distribution during early sampling.
\textbf{(c)} Disabling guidance at high $\sigma$ resolves the issue and restores both modes.
\textbf{(d)} Disabling guidance at low $\sigma$ has little effect and can be done to reduce computational cost.
} %
\end{figure*}
} %

\gdef\bgtrajxmax{8}
\gdef\bgtrajymin{-15}
\gdef\bgtrajymax{15}

\newcommand{\bgvzfw}{1.565\columnwidth}
\newcommand{\bgvzpw}{1.45\columnwidth}
\newcommand{\bgvzsw}{.12\linewidth}
\newcommand{\bgvzlw}{.22\linewidth}
\newcommand{\ourgreen}{green!85!black}

\newcommand{\plotBackgroundTrajP}[2]{%
\centering\footnotesize%
\begin{tikzpicture}%
\begin{axis}[
  width={\bgvzfw}, height={\trajh}, grid={none}, axis on top,
  xmin=-.1, xmax=1, xmode={linear}, xtick={.5}, xticklabels={\vphantom{$\sigma$}},
  ymin={#1}, ymax={#2}, ymode={linear}, ytick={0}, yticklabels={$\xx$},
]
\addplot+[name path=A, draw=none, mark=none, forget plot] coordinates {(1.1,-1.6) (1.1,2.3)};
\addplot+[name path=B, draw=none, mark=none, forget plot] table[x index=1, y index=0] {figures/trajectory/traj_pdfs.txt};
\addplot[C1, opacity=0.3, forget plot] fill between[of=A and B];
\addplot[C1, forget plot] table[y index=0, x index=1] {figures/trajectory/traj_pdfs.txt};
\addplot+[name path=A, draw=none, mark=none, forget plot] coordinates {(1.1,-1.6) (1.1,2.3)};
\addplot+[name path=B, draw=none, mark=none, forget plot] table[x index=2, y index=0] {figures/trajectory/traj_pdfs.txt};
\addplot[C2, opacity=0.3, forget plot] fill between[of=A and B];
\addplot[C2, forget plot] table[y index=0, x index=2] {figures/trajectory/traj_pdfs.txt};
\end{axis}
\end{tikzpicture}%
}%

\newcommand{\plotBackgroundTraj}[4]{%
\centering\footnotesize%
\begin{tikzpicture}%
\begin{axis}[
  width={\bgvzpw}, height={\trajh}, grid={none}, axis on top,
  xmin={0}, xmax={\bgtrajxmax}, xmode={linear}, xtick={4}, xticklabels={$\sigma$},
  ymin={\bgtrajymin}, ymax={\bgtrajymax}, ymode={linear}, ytick={}, yticklabels={},
]
\addplot graphics [xmin=0,xmax=\bgtrajxmax,ymin=\bgtrajymin,ymax=\bgtrajymax] {figures/trajectory/uncond_far.png};
\foreach \col in {1,...,16}{\addplot[black, thin, forget plot, opacity=0.5] table[x index=0, y index=\col] {#1};}
\addplot[#2, thin, forget plot, opacity=1] table[x index=0, y index=#3] {#1};
\addplot[only marks, #2, thick, forget plot, opacity=1] coordinates {(5.386, -7.8745)};
\addplot[-{Latex[scale=1]}, #2, thick, forget plot, opacity=1] coordinates {#4};
\draw[black] (3.770,-10.012) rectangle (6.170,-1.012); %
\end{axis}
\end{tikzpicture}%
}%

\gdef\arrowUncond{(5.386, -7.8745)
    (4.282, -6.325)
}
\gdef\arrowCond{(5.386, -7.8745)
    (4.282, -6.090)
}
\gdef\arrowGuided{(5.386, -7.8745)
    (4.282, -4.834)
}

\newcommand{\plotBackgroundVectors}{%
\centering\footnotesize%
\begin{tikzpicture}%
\begin{axis}[
  width={\bgvzpw}, height={\trajh}, grid={none}, axis on top,
  xmin=3.770, xmax=6.170, xmode={linear}, xtick={4}, xticklabels={\vphantom{$\sigma$}},
  ymin=-10.012, ymax=-1.012, ymode={linear}, ytick={}, yticklabels={},
]
\addplot graphics [xmin=0,xmax=\bgtrajxmax,ymin=\bgtrajymin,ymax=\bgtrajymax] {figures/trajectory/uncond_far.png};
\addplot[red, thin, forget plot, opacity=1] table[x index=0, y index=2] {figures/trajectory/traj_far_uncond.txt};
\addplot[\ourgreen, thin, forget plot, opacity=1] table[x index=0, y index=2] {figures/trajectory/traj_far_cond.txt};
\addplot[blue, thin, forget plot, opacity=1] table[x index=0, y index=1] {figures/trajectory/traj_far_full.txt};
\addplot[-{Latex[scale=1]}, red, thick, forget plot, opacity=1] coordinates {\arrowUncond};
\addplot[-{Latex[scale=1]}, \ourgreen, thick, forget plot, opacity=1] coordinates {\arrowCond};
\addplot[-{Latex[scale=1]}, blue, thick, forget plot, opacity=1] coordinates {\arrowGuided};
\addplot[only marks, black, thick, forget plot, opacity=1] coordinates {(5.386, -7.8745)};
\end{axis}
\end{tikzpicture}%
}%

\newcommand{\figBackground}{
\begin{figure*}[t]
\hspace*{-1mm}%
\begin{subfigure}{\bgvzsw}\plotBackgroundTrajP{\bgtrajymin}{\bgtrajymax}\vspace{-1.0mm}\caption{\label{fig:bgPDFs}\small \hspace*{2mm}(a) PDFs}\end{subfigure}\hspace*{-1.5mm}%
\begin{subfigure}{\bgvzlw}\plotBackgroundTraj{figures/trajectory/traj_far_uncond.txt}{red}{2}{\arrowUncond}\vspace{-1.0mm}\caption{\label{fig:bgUncond}\small \hspace*{2mm}(b) Unconditional}\end{subfigure}%
\begin{subfigure}{\bgvzlw}\plotBackgroundTraj{figures/trajectory/traj_far_cond.txt}{\ourgreen}{2}{\arrowCond}\vspace{-1.0mm}\caption{\label{fig:bgCond}\small \hspace*{2mm}(c) Conditional}\end{subfigure}%
\begin{subfigure}{\bgvzlw}\plotBackgroundTraj{figures/trajectory/traj_far_full.txt}{blue}{1}{\arrowGuided}\vspace{-1.0mm}\caption{\label{fig:bgGuided}\small \hspace*{2mm}(d) Guided, $\gweight=6$}\end{subfigure}%
\begin{subfigure}{\bgvzlw}\plotBackgroundVectors\vspace{-1.0mm}\caption{\label{fig:bgTangents}\small \hspace*{2mm}(e) Tangent vectors}\end{subfigure}%
\caption{\label{fig:background}%
Visualizing the effect of guidance. \textbf{(a)} The unconditional (orange) and conditional (green) PDFs. In (b) through (e), the orange unconditional density is visualized in the background. \textbf{(b)} Sample trajectories for the unconditional distribution. \textbf{(c)} Trajectories for the conditional distribution. \textbf{(d)} Trajectories for the guided distribution with $\gweight=6$. \textbf{(e)} The tangent vectors $\mathrm{d}\xx / \mathrm{d}\sigma$ at the intersection point of the three marked trajectories. The difference of the unconditional (red) and conditional (green) vectors is magnified as per Equation~\ref{eq:cfgodeb}, causing the unexpected detour in low-probability areas and a mode drop. See Figure~\ref{fig:trajectory} for details and comparison to our approach.
} %
\end{figure*}
} %

\newcommand{\figImageNetGuidancePRCurvesB}{%
\centering\footnotesize%
\gdef\datafile{figures/imagenet512/metrics/guidance_vs_pr-dinov2.txt}%
\begin{tikzpicture}%
\begin{axis}[
  width={0.8\linewidth}, height={60mm}, grid={major},
  xmin={0.3}, xmax={0.7}, xtick={0.3, 0.4, 0.5, 0.6, 0.7}, xticklabels={$0.3$, $0.4$, $0.5$, $0.6$, $0.7$},
  ymin={0.4}, ymax={1.0}, ytick={0.4, 0.5, 0.6, 0.7, 0.8, 0.9, 1.0}, yticklabels={$0.4$, $0.5$, $0.6$, $0.7$, $0.8$, $0.9$, $1.0$},
  xlabel={Recall}, x label style={at={(axis description cs:0.5,-0.07)}, anchor=north},
  ylabel={Precision}, y label style={at={(axis description cs:-0.055,0.5)}, anchor=south},
  legend pos={north west}, legend cell align={left}, legend columns={2},
]
\addplot[C0, name path=S_ours] table[x=r_median, y=p_median, col sep=comma, filter={label}{edm2s-icfg_0.85_5.00}] {\datafile};
\addplot[black, mark=*, mark size=1pt, forget plot, nodes near coords align={north}, nodes near coords={\small\hspace{-12mm}$w=4.0$}] coordinates {(0.504,0.688)};
\addplot[C0, mark=triangle*, mark size=1.5pt, forget plot, nodes near coords align={south}, nodes near coords={\small\hspace{5mm}$46.25$}] coordinates {(0.528,0.685)};
\addlegendentry{S, Ours, $\sigma \in (0.60, 5.00]$}

\addplot[C1, name path=S_cfg] table[x=r_median, y=p_median, col sep=comma, filter={label}{edm2s-cfg_0_inf}] {\datafile};
\addplot[C1, mark=triangle*, mark size=1.5pt, forget plot, nodes near coords align={south}, nodes near coords={\small\hspace{5mm}$52.32$}] coordinates {(0.490,0.703)};
\addplot[black, mark=*, mark size=1pt, forget plot, nodes near coords align={north}, nodes near coords={}] coordinates {(0.586,0.489)};
\addplot[black, mark=*, mark size=1pt, forget plot, nodes near coords align={south}, nodes near coords={}] coordinates {(0.318,0.703)};
\addlegendentry{S, CFG}

\addplot[C2, name path=XXL_ours] table[x=r_median, y=p_median, col sep=comma, filter={label}{edm2xxl-icfg_0.85_5.00}] {\datafile};
\addplot[C2, mark=triangle*, mark size=1.5pt, forget plot, nodes near coords align={south}, nodes near coords={\small\hspace{5mm}$29.16$}] coordinates {(0.639,0.745)};
\addplot[black, mark=*, mark size=1pt, forget plot, nodes near coords align={south}, nodes near coords={\small\begin{minipage}{12mm}$w=4.0$\vspace*{1.5mm}\end{minipage}}] coordinates {(0.602,0.758)};
\addlegendentry{XXL, Ours, $\sigma \in (0.60, 5.00]$}

\addplot[C3, name path=XXL_cfg] table[x=r_median, y=p_median, col sep=comma, filter={label}{edm2xxl-cfg_0_inf}] {\datafile};
\addplot[C3, mark=triangle*, mark size=1.5pt, forget plot, nodes near coords align={north}, nodes near coords={\small\hspace{-5mm}$33.09$}] coordinates {(0.622,0.745)};
\addplot[black, mark=*, mark size=1pt, forget plot, nodes near coords align={north}, nodes near coords={\small\hspace{-20mm}$w=1.0$}] coordinates {(0.682,0.530)};
\addplot[black, mark=*, mark size=1pt, forget plot, nodes near coords align={east}, nodes near coords={\small\hspace{-20mm}$w=4.0$}] coordinates {(0.398,0.777)};
\addlegendentry{XXL, CFG}
\end{axis}
\end{tikzpicture}%
}%

\newcommand{\figImageNetGuidancePRCurves}{%
    \begin{figure}[t]%
        \centering
        \footnotesize%
        \figImageNetGuidancePRCurvesB%
        \caption{%
            Precision and recall curves for classifier-free guidance (orange, red) and our method (blue, green), when the guidance weight $w$ is varied from 1.0 to 4.0 in 0.1 increments.
            Black points indicate the minimum and maximum guidance weights in the sweep, while colored triangles show the precision/recall tradeoffs that achieve the best \FDDINO{}.
            We used the DINOv2 feature space in this plot, following the recommendation by Stein et al.~\cite{stein2023exposing}.
            The curves represent median over three evaluations.}\vspace*{-2mm}%
        \label{fig:imagenet512_guidance_vs_pr}%
    \end{figure}
}%

\begin{document}

\maketitle

\begin{abstract}

Guidance is a crucial technique for extracting the best performance out of image-generating diffusion models.
Traditionally, a constant guidance weight has been applied throughout the sampling chain of an image.
We show that guidance is clearly harmful toward the beginning of the chain (high noise levels), largely unnecessary toward the end (low noise levels), and only beneficial in the middle.
We thus restrict it to a specific range of noise levels, improving both the inference speed and result quality. 
This limited guidance interval improves the record FID in ImageNet-512 significantly, from 1.81 to 1.40.
We show that it is quantitatively and qualitatively beneficial across different sampler parameters, network architectures, and datasets, including the large-scale setting of Stable Diffusion XL.
We thus suggest exposing the guidance interval as a hyperparameter in all diffusion models that use guidance.

\end{abstract}

\section{Introduction}
\vspace{-1mm}

Denoising diffusion models \cite{Ho2020,Nichol2021a,SohlDickstein2015,Song2020ddim,Song2019gradients,Song2021sde,Vincent11,Hyvarinen05} have enabled rapid advances in high-quality image synthesis based on text prompts and other forms of input \cite{Gal2023,Ruiz2023,Zhang2023}.
They scale effortlessly to large-scale datasets \cite{Balaji2022,dalle3,Rombach2021highresolution}, and also to other modalities such as video~\cite{sora2024,Blattmann2023,imagenvideo,Ho2022}, 
3D~shapes \cite{Lin2023,Poole2023,Raj2023,Shue2023},
and audio~\cite{Kong2021,Popov2021}.

Diffusion models convert an initial image of pure noise to a novel generated image through repeated application of image denoising.
This sampling chain typically contains dozens of steps, and in each step a little bit of the denoised result is blended into the noisy image.
The sampling process first gravitates towards the mean of the training data, followed by the determination of image features in an approximate coarse-to-fine manner based on the remaining noise.
This iterative process, where the image is formed little by little, offers considerable flexibility in terms of encouraging or discouraging certain kinds of behavior at each step.

Negative prompts \cite{negativeprompt} are a widely used concept, where the sampling process is given an additional anti-goal that is to be avoided. For example, ``nudity'' might be a common negative prompt in text-based image generators.
At every sampling step, the denoiser is executed twice: once for the positive and once for the negative prompt, and the positive result is then extrapolated further away from the negative one based on a weight parameter.
This works remarkably well in practice.
Classifier-free guidance (CFG)~\cite{Ho2021classifierfree} builds on this general concept. 
It uses an unconditional model (no class information or text prompts) as a negative prompt, causing the result image to align more strongly with the conditioning signal.

In practice, all large-scale image generators rely heavily on CFG. It allows a mathematically justified way of truncating the distribution of generated images~\cite{guidance,Ho2021classifierfree}, basically trading variation for perceptually higher image quality.
By convention, the same guidance weight is used in all sampling steps.
We observe that this is sub-optimal because CFG behaves very differently on high, middle, and low noise levels.
On high noise levels, it drastically reduces the variation in the results, basically leading them towards a handful of ``template images'' per prompt.
On middle levels, it causes the sampling to more decisively choose some set of features, leading to crisper and perceptually more pleasing results.
On low levels, it is largely unnecessary.
Similar observations have been made in the Stable Diffusion community \cite{BirchTwitter, GoodwinGithub, HowardTwitter}, and
Muse \cite{Chang2023DynGuidance} and Masked DiT v2 \cite{Gao2023MaskedDiT}
observe that making guidance weight noise level-dependent improves the results.
In the context of prompt inversion, Mahajan et al.~\cite{Mahajan2024PromptInversion} notice that limiting the inversion to specific noise levels leads to improved result quality.
However, these works do not quantify the effect on distribution metrics with the exception of Sadat et al.~\cite{Sadat2024DynCFG}, whose ``dynamic CFG'' limits a linearly varying guidance weight to an interval of noise levels. Interestingly, they conclude that dynamic CFG leads to rather poor results, while a more complicated condition annealing scheme is required for good quantitative results.

We suggest that guidance should be simply limited to 
an interval of sampling steps in the middle, where the net effect is positive, without otherwise changing the guidance weight. This avoids most of the detrimental effects of guidance, while also reducing computational cost.
We show that an optimal guidance interval improves the state-of-the-art FID \cite{Karras2023edm2} in ImageNet-512 from 1.81 to 1.40 and also leads to an improved visual quality.
The benefits are consistent across sampler parameters, network architectures, and datasets, including Stable Diffusion XL. Code is available at \footnotesize{\url{https://github.com/kynkaat/guidance-interval}}\normalsize

\section{Background}
\label{sec:background}
\vspace{-1mm}
\figBackground

The concepts in this and the following section are illustrated in \autoref{fig:background} using a synthetic 1D example. In this example, generation is performed by ideal analytic denoisers, avoiding all approximations that a learned denoiser might cause. While this renders classifier-free guidance strictly harmful in the scenario, the example allows us to intuitively visualize the kinds of harm it causes.

The goal of a denoising diffusion model is to draw samples from a data distribution $\pdata(\xx)$. Let us define a series of smoothed distributions $p(\xx;\sigma)$, so that each individual distribution is the convolution between $\pdata$ and a Gaussian noise distribution with standard deviation $\sigma$. Following the EDM formulation~\cite{Karras2022elucidating}, the evolution of a sample $\xx \sim p(\xx;\sigma)$ w.r.t.~a change in~$\sigma$ is described by the ordinary differential equation (ODE):
\begin{equation}
\label{eq:ode}
\mathrm{d}\xx / \mathrm{d}\sigma = -\big(D_\theta(\xx;\sigma) - \xx\big) / \sigma\text{,}
\end{equation}
where $D_\theta$ is a denoiser model with parameters $\theta$, optimized to minimize the expected $L_2$ denoising error:
\begin{equation}
\theta = \mathrm{argmin}_\theta ~ \EE_{\yy\sim \pdata , \sigma\sim p_\text{train} , \nn\sim\mathcal{N}(\boldzero,\sigma^2\boldI)}
  \lVert D_\theta(\yy+\nn;\sigma) - \yy \rVert^2_2\text{.}\vspace*{1mm}
\end{equation}
Here, $p_\text{train}(\sigma)$ is the training distribution of noise levels, which we consider to be an implementation detail of $D_\theta$.
To generate a sample from the data distribution, we first draw an initial sample $\xx_0 \sim p(\xx;\smax)$, where $\smax$ is chosen to be large enough so that $p(\xx;\smax)$ is approximately equal to pure Gaussian distribution and thus trivial to sample from. We then follow the ODE of Equation~\ref{eq:ode} to evolve $\xx_0$ towards $\sigma=0$, i.e., the data distribution. \autoref{fig:bgPDFs} illustrates the target distribution (orange). \autoref{fig:bgUncond} depicts the diffused target distribution over the $\sigma$ axis and a set of sample trajectories computed by solving \autoref{eq:ode} from several different initial conditions.

We can think of classifier-free guidance~\cite{Ho2021classifierfree} as constructing a modified ODE where \mbox{$\mathrm{d}\xx / \mathrm{d}\sigma$} is defined as a linear combination between a conditional ODE and an unconditional ODE:
\begin{align}
\mathrm{d}\xx / \mathrm{d}\sigma &~=~ \label{eq:cfgodea}
  \gweight     \big[\!-\!\big(D_\theta(\xx|\cc;\sigma) - \xx\big) / \sigma\big] +
  (1-\gweight) \big[\!-\!\big(D_\theta(\xx;\sigma) - \xx\big) / \sigma\big] \\
  &~=~ \label{eq:cfgodeb}
  -\big(
  \gweight     D_\theta(\xx|\cc;\sigma) +
  (1-\gweight) D_\theta(\xx;\sigma)
  - \xx \big) / \sigma \text{,}
\end{align}
where $\gweight$ is the guidance weight and $\cc$ is the condition information given to the denoiser $D_\theta$ (cf.~\autoref{fig:bgCond}).
Setting \mbox{$\gweight > 1$} results in \emph{extrapolating} the effect of the condition with respect to the
  unconditional result, i.e., the sample is effectively pushed away from the unconditional result.
This extrapolation can be seen~\cite{guidance,Ho2021classifierfree} as raising the conditional likelihood $p(\cc|\xx;\sigma)$ to a power greater than one, which, intuitively, aims to concentrate the probability mass to the regions that most agree with the condition. However, as illustrated in Figure~\ref{fig:background}(d,\,e) and the next section, this ``oversteering'' may direct the trajectories away from the data distribution and cause mode drops.

Most commonly, a single denoiser model $D_\theta$ is trained to accept either conditional or unconditional input by
  dropping the conditioning information 10--20\% of the time during training.
Alternatively, we can train two separate models
  $\smash{D_{\theta\raisebox{-.15ex}{\tiny cond.}}(\xx|\cc;\sigma)}$ and
  $\smash{D_{\theta\raisebox{-.15ex}{\tiny uncond.}}(\xx;\sigma)}$.
This makes it possible to reduce the capacity of the unconditional model considerably
  to improve the overall sampling speed~\cite{Karras2023edm2}.

Sampling the ODE is done by taking a number of discrete steps that bring the noise level from $\smax$ to zero,
  giving rise to a sequence of images $\xx_0, \xx_1, \gdots, \xx_N$, each with its corresponding noise level $\sigma_i$.
Various discretization schemes and solvers have been proposed~\cite{Karras2022elucidating}.
Regardless of the specifics, the computational cost is directly proportional to the number of sampling steps $N$.

\section{Our method}
\vspace{-1mm}
\label{sec:toy}

\figTrajectory

In \autoref{fig:trajectory}, we continue to probe the downsides of CFG using the previous toy example.
We observe that applying guidance at all noise levels\,---\,as is typical\,---\,causes the sampling trajectories
  to drift quite far from the the smoothed data distribution (Figure~\ref{fig:trajectory}b).
This is caused by the unconditional trajectories effectively repelling the guided trajectories, as discussed above,
  yielding badly skewed intermediate distributions.
As a result, the sampler drops one of the modes (almost) entirely.

As most of the drift seems to be caused at high noise levels, we disable CFG in those sampling steps (Figure~\ref{fig:trajectory}c).
This correctly recovers both modes of the conditional distribution.
In addition, disabling guidance at low noise levels (Figure~\ref{fig:trajectory}d) has only a small effect on the resulting distribution,
  providing a simple way to reduce the sampling cost with minimal effect on outputs.

Although this toy example is grossly simplified, we hypothesize that broadly similar effects occur in full-scale diffusion models as well.
In Section~\ref{sec:results} we can see, e.g., image compositions becoming less varied due to guidance, somewhat akin to the mode dropping observed in the toy example. That behaviour is difficult to explain by local sharpening of probability distributions alone (Section~\ref{sec:background}).

\subsection{Practice}
\vspace{-1mm}

Motivated by the above observations, we propose to only apply guidance in a continuous interval of noise levels in the middle of the sampling chain and disable it elsewhere.
Concretely, we redefine the ODE of Equation~\ref{eq:cfgodeb} by replacing $\gweight$ with a piecewise constant function:
\begin{align}
  \mathrm{d}\xx / \mathrm{d}\sigma &~=~ 
    -\Big(\gweight(\sigma) D_\theta(\xx|\cc;\sigma) + \big( 1 - \gweight(\sigma) \big) D_\theta(\xx;\sigma) - \xx \Big) / \sigma
    \label{eq:ourodea} \text{,} \\
  \text{where} ~~~ \gweight(\sigma) &~=~
    \begin{cases}
      \gweight & \text{if } \sigma \in \left( \sigmalo, \sigmahi \right] \\
      1 & \text{otherwise.}
    \end{cases}
    \label{eq:ourodeb}
\end{align}
Here, $\sigmahi$ denotes the point in the sampling chain where we enable guidance and $\sigmalo$ is the point where we turn it off.
In our formulation, traditional CFG is recovered by setting $\sigmalo=0$ and $\sigmahi=\infty$.

Virtually all existing deterministic samplers can be seen as numerical Runge--Kutta solutions to the ODE of Equation~\ref{eq:cfgodeb}, obtained through a number of discrete steps.
While the correspondence might not be obvious in all cases, we can nevertheless characterize the steps with respect to $\sigma$ as detailed in Appendix~\ref{app:sigmas}.
For example, in the case of Stable Diffusion XL~\cite{Podell2024SDXL}, we have 32 steps corresponding to the transitions $\sigma_0 \rightarrow \sigma_1,~ \sigma_1 \rightarrow \sigma_2, \ldots,~ \sigma_{31} \rightarrow \sigma_{32}$, where $\sigma_0 = 14.61,~ \sigma_1 = 13.41,~ \sigma_2 = 12.28, \ldots,~ \sigma_{31} = 0.03,$ and $\sigma_{32} = 0$.

The underlying assumption common to all Runge--Kutta methods is that $\mathrm{d}\xx / \mathrm{d}\sigma$ should be sufficiently smooth within each step.
In Equation~\ref{eq:ourodeb}, however, we intentionally introduce discontinuities at $\sigmalo$ and $\sigmahi$.
In order to satisfy the smoothness requirement, we must thus ensure that both transitions happen {exactly} at step boundaries so that the value of $\gweight(\sigma)$ stays constant within each step.
In practice, we choose to do this by rounding $\sigmalo$ and $\sigmahi$ appropriately, i.e., by setting $\sigmahi = \sigma_i$ and $\sigmalo = \sigma_j$ for some $i < j$.
Note that this leads to a seemingly high numerical precision in the values of $\sigmalo$ and $\sigmahi$, which should not be taken as an indication of extremely precise tuning.

\section{Results}
\vspace{-1mm}
\label{sec:results}

We will first evaluate and ablate our method quantitatively using ImageNet~\cite{Deng2009imagenet}. 
Limiting the guidance interval leads to clearly identifiable qualitative changes in the images, which we subsequently demonstrate also in the large-scale context using Stable Diffusion XL~\cite{Podell2024SDXL}.
Please refer to Appendix~\ref{app:qualitative_results} for additional results.

\subsection{Main results}
\vspace{-1mm}
\label{sec:imagenet512_results}
\tabResultsFiveTwelve

We mainly evaluate our method on ImageNet at $512 \times 512$, using the current state-of-the-art approach EDM2~\cite{Karras2023edm2} as a baseline.\footnote{\url{https://github.com/NVlabs/edm2}}
We use the small (EDM2-S) and the largest (EDM2-XXL) models as-is with the default sampling parameters: 32 deterministic steps with a $2^{\text{nd}}$ order Heun sampler \cite{Karras2022elucidating}.

Table~\ref{tabResultsFiveTwelve} shows that our method improves FID~\cite{Heusel2017} and the more recently proposed \FDDINO{}~\cite{stein2023exposing} significantly. 
Using EDM2-S, FID improves from 2.23 to 1.68, already beating the state-of-the-art in this dataset.
With EDM2-XXL, the record further improves to 1.40 and \FDDINO{} also improves from 33.09 to 29.16. 

To find the optimal parameters for each case, we performed a full grid search over $\gweight$, $\sigmalo$, and $\sigmahi$.
In the case of EDM2-XXL, the best FID is achieved by applying guidance at 6 of the 32 steps, corresponding to noise levels $\sigma \in (0.19, 1.61]$, with weight $\gweight = 2.0$.
The best \FDDINO{} is obtained with slightly higher noise levels $\sigma \in (0.60, 5.00]$ and a slightly higher weight $\gweight = 2.9$.

For additional validation, we also tested our method on diffusion transformers~\cite{Peebles2022} using the DiT-XL/2 model\footnote{\url{https://github.com/facebookresearch/DiT}} with default sampling parameters: 250 step iDDPM \cite{Nichol2021a}. Limiting the guidance interval leads to significant improvements with this model as well. The best FID results were obtained by using guidance with $\gweight=2.5$ in 75 of the 250 sampling steps, corresponding to the interval $\sigma\in\left(0.34, 1.02\right]$. The best \FDDINO{} is again obtained with slightly higher noise levels $(0.45,1.23]$ and weight $\gweight = 4.0$.

\subsection{Ablations} 
\vspace{-1mm}
\figImageNetGuidanceFIDCurves
\figImageNetGuidancePRCurves
Figure~\ref{fig:imagenet512_guidance_vs_fid} shows that standard classifier-free guidance is quite sensitive to the guidance weight. When the weight is too high, the output image distribution is excessively truncated, and the harm caused outside the useful interval starts to outweigh the benefits obtained within. 
In contrast, limiting the guidance interval allows the use of much higher guidance weight, and \FID{} or \FDDINO{} are far less sensitive to the exact choice. 

Figure~\ref{fig:imagenet512_guidance_vs_pr} shows precision and recall~\cite{Kynkaanniemi2019PR} curves for CFG and our method, evaluated with varying guidance weights in DINOv2 feature space, as suggested by Stein at al.~\cite{stein2023exposing}. Compared to CFG, our method achieves better \FDDINO{} primarily by improving Recall without significantly affecting Precision. This is consistent with the qualitative observation that the results are more varied.%

\figAblationGraphs%
Figure~\ref{fig:ablation_graphs} probes the sensitivity of our results to the chosen guidance interval.
In this test, we sweep over $\sigmalo$ and $\sigmahi$, while keeping the other interval endpoint, $\sigmahi$ or $\sigmalo$, and the guidance weight $\gweight$  as the optimal choices as reported in Table~\ref{tabResultsFiveTwelve}.
The left side shows a sweep over $\sigmahi$, i.e., the highest noise level with guidance.
Including too high noise levels to the guidance interval leads to truncation of the image distribution, which can be seen as an increase in FID.
Furthermore, too narrow an interval (low $\sigmahi$) yields sub-optimal results. For both EDM2 models the optimal choice for $\sigmahi$ is located at the middle noise levels.
The right side shows a sweep over $\sigmalo$, i.e., the lowest noise level with guidance. Applying guidance at low noise levels does not bring additional benefits, compared to the middle levels. Thus, guidance can be disabled in most of the low noise levels to decrease sampling cost, an observation also made in \cite{Castillo2023AdaptiveGuidance}.

To estimate the optimal guidance interval in practice, the upper and lower guidance limits can be determined separately, without the need for a two-dimensional search. This happens by first establishing the optimal upper limit by keeping the lower limit at zero.
This can be done because the lower limit affects the result only weakly, and in a predictable way (Figure~\ref{fig:ablation_graphs}, right). 
Once the optimal upper limit is known, the lower limit is determined.
Optionally, a bisection method can be used for accelerating both search operations.
Finally, one can reduce the sample size of FID evaluation from 50k to, say, 5k, at least for an initial run, which accelerates the process by $10\times$.

We have found that the optimal choice of $\sigmalo$ and $\sigmahi$ is not overly sensitive to the other sampling parameters.
For example, if we halve or double the number of steps with EDM2-S, the optimal guidance interval remains unchanged.
With 16 steps, our method improves FID from 2.49 to 1.84, and with 64 steps, from 2.27 to 1.70.

In an additional test, we tried applying various smooth weighting functions to the guidance weight (less guidance towards the ends of the interval), but these tests did not improve the results over the simple binary inclusion.
We also tried estimating the importance of guidance at individual noise levels by enabling or disabling it at each sampling step at a time. However, these tests consistently underestimated the downsides of guidance, suggesting that they build up cumulatively over multiple consecutive steps.

\subsection{Qualitative analysis}
\vspace{-1mm}
\figQualitativeResultsA
\figQualitativeResultsB
\figQualitativeResultsC
\figQualitativeResultsD

With the rise of recent large-scale image generators, ImageNet can hardly be considered a meaningful benchmark for gauging perceptual image quality. Thus, we primarily focus on evaluating our method in the context of Stable Diffusion XL (SD-XL), but we also provide corresponding results for ImageNet using EDM2-XXL. For SD-XL, we use the official pre-trained checkpoint\footnote{\url{https://github.com/Stability-AI/generative-models}} with a standard 32-step deterministic Heun sampler, where the first step corresponds to $\sigma = 14.61$.

With SD-XL, we apply guidance at 50\% of the sampling steps, corresponding to noise levels $\sigma \in \left(0.28, 5.42\right]$, with weight $\gweight = 16$. These parameters were chosen by visual inspection. The beneficial interval is wider than in ImageNet, likely due to the more varied dataset used in the training of SD-XL. Consequently, our method leads to over $20\%$ speed-up due to a lower number of unconditional model evaluations \cite{BirchTwitter}.

Figure~\ref{fig:sdxl_imagenet_qualitative_results_A} shows a comparison between standard classifier-free guidance with low and high weights (left and middle columns) and our method with high guidance weight (right column).
When the guidance weight is increased in standard CFG (middle), the composition of the image tends to change drastically, towards some limited set of per-class ``templates''.
Furthermore, the colors saturate unnaturally as the guidance weight increases. When we limit the guidance interval (right), image diversity is preserved to a significant degree and the color saturation is also reduced, although excessively large guidance weights can still lead to over-saturation.

Figure~\ref{fig:sdxl_imagenet_qualitative_results_B} shows the effect of increasing the guidance weight with our method. With low weight, the images appear blurry, inconsistent, and lacking in detail. Increasing the weight improves the rendition of details while retaining the original image composition. 

As the task of selecting the best guidance interval $(\sigmalo, \sigmahi]$ with SD-XL is necessarily subjective, we provide a visual ablation of this choice in Figure~\ref{fig:sdxl_qualitative_results_C}. 
Modifying the upper limit $\sigmahi$, i.e., disabling guidance at high noise levels, has two distinct effects. First, it affects the overall image composition\,---\,higher values lead to more simplified image layouts whereas low values lead to unnecessary complexity. Second, high values lead to oversaturated colors whereas lower $\sigmahi$ leads to a blander color scheme.
Similar to EDM2 results, changing the lower limit $\sigmalo$ has only a modest effect\,---\,guidance can be disabled from most of the low noise levels with no noticeable impact while improving the inference speed. %

Lastly, Figure~\ref{fig:qualitative_results_D} compares the effects of increasing the guidance weight in standard CFG vs.~our method with EDM2-XXL.

\section{Conclusions}
\label{sec:conclusions}
\vspace{-1mm}

Classifier-free guidance is an indispensable tool for improving the results of practically all image-generating diffusion models. 
As our simple modification improves the results both numerically and visually, and also reduces sampling cost, we recommend exposing the guidance interval as an additional sampler parameter.

Future work could investigate whether the optimal guidance interval can be automatically derived from the ODE, and the role played by the non-idealities in the trained denoiser. 
A recent work by Biroli et al.~\cite{Biroli2024Regimes} predicts from a dataset the interval where the generated images specialize to a certain class. A follow-up study could examine whether their ``speciation'' interval overlaps with the interval that is beneficial for guidance.

\section*{Acknowledgements}

This work was partially supported by the European Research Council (ERC Consolidator Grant 866435), and made use of computational resources provided by the Aalto Science-IT project and the Finnish IT Center for Science (CSC).

{\small
	\bibliographystyle{ieee}
	\bibliography{main}
}

\FloatBarrier{}
\newpage{}
\appendix
\section{Characterizing sampling steps in noise levels}
\label{app:sigmas}

In the main paper, we reported the guidance interval measured in noise levels $\sigma$. Here, we show for each model how the indices of sampling steps are mapped to noise levels. For EDM2 models and SD-XL, we use the discretization from \cite{Karras2022elucidating}. The $i$th sampling step corresponds to noise level that is given by:
\begin{equation}
\sigma_i = \left( \sigmamax^{\frac{1}{\rho}} +\frac{i}{N-1}\left( \sigmamin^{\frac{1}{\rho}} - \sigmamax^{\frac{1}{\rho}}\right)\right)^\rho,
\label{eq:edm2_sdxl_timesteps_to_sigmas}
\end{equation}
where $N$ is the total number of sampling steps, $\sigmamin=0.002$, $\sigmamax=80$. With SD-XL, we use $\rho=3$, which is the default value in the official code, with EDM2 models we use $\rho=7$. With DiT, we use the iDDPM discretization from \cite{Karras2022elucidating} which maps the $i$th sampling step to the corresponding noise level in the following way:
\begin{equation}
\sigma_i = u_{\lfloor j_0 + \frac{M - 1 - j_0}{N - 1}i + \frac{1}{2} \rfloor},
\label{eq:dit_timesteps_to_sigmas}
\end{equation}
where $u_M=0$, $u_{j-1}=\sqrt{\frac{u_j^2 + 1}{\text{max}(\bar{\alpha}_{j-1}/\bar{\alpha}_j, C_1)}-1}$ and $\bar{\alpha}_j=\text{sin}^2\left( \frac{\pi}{2}\frac{j}{M(C_2 + 1)}\right)$. We use the default parameters $C_1=0.001$, $C_2=0.008$, $M=1000$ and $j_0=0$ from \cite{Karras2022elucidating}.

\section{Additional qualitative results}
\label{app:qualitative_results}

Figures~\ref{fig:sdxl_imagenet_qualitative_results_A_sdxl_supp} and \ref{fig:sdxl_imagenet_qualitative_results_A_imagenet_supp} show further comparisons between classifier-free guidance and our method. 
Figures~\ref{fig:qualitative_results_B_sdxl_supp} and \ref{fig:qualitative_results_B_imagenet_supp} show additional examples from our method where we increase the guidance weight. Figures~\ref{fig:qualitative_results_B_sdxl_supp_cfga}, \ref{fig:qualitative_results_B_sdxl_supp_cfgb} and \ref{fig:qualitative_results_B_imagenet_cfg_supp} compare classifier-free guidance to our method when the guidance weight is increased.

\figQualitativeResultsASDXLSupp
\figQualitativeResultsAImageNetSupp

\figQualitativeResultsBSDXLSupp
\figQualitativeResultsBImageNetSupp

\figQualitativeResultsBSDXLSuppCFGA
\figQualitativeResultsBSDXLSuppCFGB
\figQualitativeResultsBImageNetSuppCFG

\section{Broader impacts}
\label{app:broader_impacts}

Large-scale diffusion models, such as Stable Diffusion XL, might have various negative societal effects related to the spread of disinformation or amplifying harmful biases and stereotypes. Our method improves the result quality of these models which can potentially further magnify these issues. In the large-scale setting, our method decreases the cost of sampling, but diffusion models continue to require a lot of computing power, which may contribute to wider issues such as climate change.

\section{Licenses}
\label{app:licenses}

The pre-trained EDM2 \cite{Karras2023edm2} models are licensed under the 
CC BY-NC-SA 4.0
International License by NVIDIA corporation. The pre-trained SD-XL \cite{Podell2024SDXL} model is available under the CreativeML Open RAIL++-M License by Stability AI. ImageNet \cite{Deng2009imagenet} dataset uses a custom non-commercial license.

\end{document}